\documentclass[final,12pt]{msml2022} 

\usepackage{times}
\usepackage{soul}
\usepackage{url}
\usepackage{algorithm}
\usepackage{graphicx}
\usepackage{amsmath,amssymb} 
\usepackage{algorithmic}
\usepackage{booktabs}
\usepackage{multirow}
\usepackage{tikz}
\usetikzlibrary{plotmarks}
\usepackage[]{hyperref}
\usepackage{cancel}
 
\usepackage{color,xcolor}

\definecolor{blue}{rgb}{0.0,0.0,1.0}
\definecolor{red}{rgb}{1.0,0.0,0.0}
\definecolor{purple}{rgb}{0.75, 0.0, 1.0}

\newcommand\p{\partial}
\newcommand{\mbr}{\mathbb{R}}
\newcommand{\vxi}{{\boldsymbol \xi}}
\newcommand{\cL}{{\cal L}}
\newcommand{\nn}{{\vec{\bf n}}}

\def\cd{\cdots}

\def\vzero{\boldsymbol{0}}
\def\v1{\boldsymbol{1}}

\def\vI{{\boldsymbol{I}}}

\def\vW{{\boldsymbol{W}}}

\def\ds{\displaystyle}

\def\vb{{\boldsymbol{b}}}

\def\vv{{\boldsymbol{v}}}

\def\vx{{\boldsymbol{x}}}

\def\vxi{{\boldsymbol{\xi}}}

\def\vTheta{{\boldsymbol{\Theta}}}

\usepackage{changes} 

\title[GF-Net and Fast Numerical Solver]{Learning Green's Functions of Linear Reaction-Diffusion Equations with Application to Fast Numerical Solver}

\msmlauthor{%
 \Name{Yuankai Teng} \Email{yteng@email.sc.edu}\\
 \addr Department of Mathematics, University of South Carolina, Columbia, SC, USA
 \AND
 \Name{Xiaoping Zhang} \Email{xpzhang.math@whu.edu.cn}\\
 \addr School of Mathematics and Statistics, Wuhan University, Wuhan, China
 \AND
 \Name{Zhu Wang} \Email{wangzhu@math.sc.edu}\\
 \Name{Lili Ju}\footnote{Corresponding author} \Email{ju@math.sc.edu}\\
 \addr Department of Mathematics, University of South Carolina, Columbia, SC, USA
}

\makeatletter
 \let\Ginclude@graphics\@org@Ginclude@graphics
\makeatother

\begin{document}

\maketitle

\begin{abstract}%
Partial differential equations are commonly used to model various physical phenomena, such as heat diffusion, wave propagation, fluid dynamics, elasticity, electrodynamics and so on. Due to their tremendous applications in scientific and engineering research, many numerical methods have been developed in past decades for efficient and accurate solutions of these equations on modern computing systems. Inspired by the rapidly growing impact of deep learning techniques,  we propose in this paper a novel neural network method, ``GF-Net'', for learning the Green's functions of the classic linear reaction-diffusion equation with Dirichlet boundary condition in the unsupervised fashion. The proposed method overcomes the numerical challenges for finding the Green's functions of the equations on general domains by utilizing the physics-informed neural network and the domain decomposition approach. As a consequence, it also leads to a fast numerical solver for the target equation subject to arbitrarily given sources and boundary values without network retraining. We numerically demonstrate the effectiveness of the proposed method by extensive experiments with various domains and operator coefficients. 
\end{abstract}


\section{Introduction}

Rapid development and great success of deep learning for computer vision and natural language processing have significantly prompted its application to many other science and  engineering problems in recent years. Thanks to the integration of available big data, effective learning algorithms and unprecedented computing powers, the resulted  deep learning studies have  showed increasing impacts on various subjects including  partial differential equations (PDEs), dynamical system, reduced order modeling and so on. In particular, the synthesis of deep learning techniques and numerical solution of PDEs has become an emerging research topic in addition to conventional numerical methods such as finite difference, finite element and finite volume ones.

Some popular deep learning algorithms include  the physics-informed neural networks (PINNs) 
\citep{raissi2019physics},  the deep Ritz method (DRM)  \citep{e2018ritz},  the  deep Galerkin method (DGM) \citep{sirignano2018} and the PDE-Net \citep{long2018pde}. Note that the first three are meshfree and trained without any explicitly observed data and the last one uses instead rectangular meshes and ground-truth information for training. Deep learning based methods also have been applied to construct computational surrogates for PDE models in a series of research \citep{khoo2021solving,nagoor2017dr,nabian2019deep,lee2018model,san2019artificial,mucke2019reduced,zhu2019physics,sun2020surrogate}.
On the other hand, 
classic methods for solving PDEs also have been used to understand and further improve the network structure and training settings. For instance, the connection between multigrid methods and convolutional neural networks (CNNs) was discussed in \citep{He2019mgnet} and MGNet was then proposed to incorporate them. 

The goal of our work is to design a neural network based method for fast numerical solution of the classic linear reaction-diffusion equation on  arbitrary domains, that could yield an accurate response to various sources and Dirichlet boundary values without retraining. To achieve this, we propose a neural network, called ``GF-Net'', that computes the Green's functions associated with the target PDE under Dirichlet boundary conditions.
Note that the exact solution of the target equation can be explicitly expressed in terms of the Green's function, source term and boundary values via area and line integrals \citep{Evans1998}. 
After evaluating the Green's function at a set of sample points with the trained GF-Net, the target PDE problem can numerically be solved in an efficient manner. 
As the Green's function is the impulse response of the linear differential operator, which is well approximated by the GF-Net through a nonlinear mapping, a significant advantage of the proposed method compared to most existing deep learning methods for numerical PDEs is that it does not need network retraining when the PDE source and/or the Dirichlet boundary condition change. How to determine the Green's functions of a PDE is a classic problem. The analytic formulation of Green's functions are only known for a few operators on either open spaces or domains with simple geometry \citep{Evans1998}, such as the linear reaction-diffusion operator. On the other hand, finding their numerical approximations by traditional numerical methods often turns out to be too expensive in terms of computation and memory. In addition, the high-dimensional parameter space makes it almost impossible to use model reduction to find efficient  surrogates for the Green's functions. 

In this paper, we propose a neural network architecture GF-Net that can provide a new way to tackle this classic problem for the linear reaction-diffusion equation with Dirichlet boundary condition through deep learning to overcome some limitations of traditional methods. In particular, our GF-Net is physics-informed: a forward neural network is trained by minimizing the loss function measuring the pointwise residuals, discrepancy in boundary values, and an additional term for penalizing the asymmetry of the output due to the underlying property of symmetry possessed by Green's functions. Meanwhile, to accelerate the training process, we also design a sampling strategy based on the position of the point source, and further put forth a domain decomposition approach to train multiple GF-Nets in parallel on many blocks. Note that each GF-Net is assigned and associated with a specific subdomain block. Finally, the application of the produced GF-Nets to fast numerical solution of the target equation with different sources and boundary values is carefully tested and demonstrated through experiments with various domain and operator coefficients. 

\section{Related work}

Using neural networks to solve differential equations has been investigated in several early works, e.g.,  \citep{dissanayake1994neural,lagaris1998artificial}, and recent advances in deep learning techniques have further stimulated new exploration towards this direction.

The physics-informed neural network (PINN) \citep{raissi2019physics} represents the mapping from spatial and/or temporal variables to the state of the system by deep neural networks, which is then trained by minimizing the weighted sum of the residuals of PDEs at randomly selected interior points and the errors at initial/boundary points.  This approach later has been extended to solve inverse problems \citep{raissi2020hidden}, fractional differential equations \citep{pang2019fpinns}, stochastic differential equations and uncertainty quantification \citep{nabian2019deep,yang2020physics,zhang2020learning,zhang2019quantifying}. 
Improved sampling and training strategies have been considered in \citep{lu2021deepxde,anitescu2019artificial,zhao2020solving,krishnapriyan2021characterizing}. In order to solve topology optimization problems for inverse design, PINNs with hard constraints were also investigated in \citep{lu2021physics}.
The deep Ritz method (DRM) \citep{e2018ritz} considers the variational form of PDEs, which combines the mini-batch stochastic gradient descent algorithms with numerical integration to optimize the network. Note that later the variational formulation was also considered in weak adversarial networks \citep{zang2020weak}. 
The deep Galerkin method (DGM) \citep{sirignano2018} merges the classic Galerkin method and machine learning, that is specially designed for solving a class of high-dimensional free boundary PDEs. The above three learning methods use no meshes as opposed to traditional numerical methods and are trained in the unsupervised fashion (i.e., without ground-truth data).
The PDE-Net \citep{long2018pde} proposes a stack of networks ($\delta t$-blocks) to advance the PDE solutions over a multiple of time steps. It recognizes the equivalence between convolutional filters and differentiation operators in rectangular meshes under the supervised training with ground-truth data. This approach was further combined with a symbolic multilayer neural network for recovering PDE models in \citep{long2019pde}.

Learning the operators can provide better capability and efficiency by solving a whole family of PDEs instead of a single fixed equation. The Fourier neural operator (FNO)  \citep{li2020fourier} aims to parameterize the integral kernel in Fourier space and is able to generalize trained models to different spatial and time resolutions. The DeepONet \citep{lu2021learning} extends the universal approximation theorem for operators in \citep{chen1995universal} to deep neural networks. It contains two subnetworks to encoder the input functions and its transformed location variable respectively and then uses the extended theorem to generate the target output. DeepONet has quite powerful generalization ability to handle diverse linear/nonlinear explicit and implicit operators. Error estimation for DeepONet was recently investigated in \citep{lanthaler2022error}. Physics-informed DeepONets proposed in \citep{wang2021learning} further reduces the requirement for data and achieves up to three order of magnitude faster inference time than convectional method. Learning operators by using graph neural networks (GNN) was first considered in \citep{anandkumar2020neural} and later improved in \citep{li2020multipole}. In \citep{li2021physics}, the physics-informed neural operator (PINO), which combines the operator learning and function approximation frameworks to achieve higher accuracy, was proposed. In addition, attention mechanism was further introduced in \citep{kissas2022learning} to solve the climate prediction problem.

There also exist a few works which compute or use Green's functions to solve PDEs.
The data-driven method recently proposed in \citep{boulle2022data} applies rational neural network structure to train networks with generated excitation for approximating Green's functions and the homogeneous solution separately. The PINN was recently used to solving some PDEs with point source in \citep{Dong2021}.
For handling nonlinear boundary value problems, the DeepGreen in \citep{gin2021deepgreen} first linearizes the nonlinear problems using a dual autoencoder architecture, then evaluates the Green's function of the linear operator, and finally inversely transforms the linear solution to solve the nonlinear problem.

\section{Linear Reaction-diffusion equations and Green's functions}

Let $\Omega\subset \mathbb{R}^d$ be a bounded domain, we consider the linear reaction-diffusion operator of the following form:
\begin{equation}
\begin{array}{rrl}
\cL (u)(\vx)&:=&-\nabla\cdot (a(\vx) \nabla u(\vx))+r(\vx)u(\vx), \quad \vx\in \Omega,
\end{array}
\label{operator}
\end{equation}
where  $a(\vx)>0$  is the diffusion coefficient and  $r(\vx)\geq 0$ is the reaction coefficient. 
The corresponding linear reaction-diffusion problem with  Dirichlet boundary condition then reads: 

\begin{equation}\left\{
\begin{array}{rl}
\cL(u)(\vx) = f(\vx),& \quad \vx\in \Omega,\\
u(\vx) = g(\vx),& \quad \vx\in  \p \Omega,
\end{array}\right.
\label{eq:dr}
\end{equation}
where $f(\vx)$ is the given source term and $g(\vx)$ the boundary value.
The Green's function $G(\vx, \vxi)$ represents the impulse response of the PDE subject to homogeneous Dirichlet boundary condition, that is, for any impulse source point $\vxi \in \Omega$,
\begin{equation}\left\{
\begin{array}{rl}
\cL(G)(\vx,\vxi) = \delta(\vx-\vxi), &\quad \vx\in \Omega,\\
G(\vx,\vxi) = 0, &\quad \vx\in  \p \Omega,
\end{array}\right.
\label{eq:Green}
\end{equation}
where $\delta(\vx)$ denotes the Dirac delta source function satisfying $\delta(\vx)=0$ if $\vx \ne {\bf 0}$ and $\int_{\mbr^d}\delta(\vx)\,d\vx=1$. Note that the Green's function $G$ is symmetric, i.e., $G(\vx,\vxi) = G(\vxi,\vx)$.
If $G(\vx,\vxi)$ is known, the solution of the problem \eqref{eq:dr} can be readily expressed by the following formula: 
 \begin{equation}
   u(\vx) =   \int_\Omega f(\vxi)  G(\vx, \vxi) \,d\vxi
   - \int_{\p \Omega} g(\vxi)a(\vxi)({\nabla_{\vxi} G}(\vx,\vxi)\cdot {\vec{\bf n}}_{\vxi})  \,dS(\vxi),\quad \forall\,\vx\in \Omega,
 \label{eq:rd-soln}
 \end{equation}
 where ${\vec{\bf n}}_{\vxi}$ denotes the unit outer normal vector  on $\p \Omega$. However, the above Green's function (i.e., the solution of \eqref{eq:Green}) on a general domain usually does not have analytic form, and consequently we need to numerically approximate the Green's function.

\section{GF-Net: Learning Green's functions} 

We will construct a {\em deep feedforward network}, {\bf GF-Net}, to learn the Green's function associated with the operator \eqref{operator}, and then use it to fast solve  the problem \eqref{eq:dr} based on the formula \eqref{eq:rd-soln}. In order to represent the Green's function obeying \eqref{eq:Green}, we adopt the framework of PINNs \citep{raissi2019physics}, a {\em fully connected neural network}, with slight modifications and accommodate the symmetric characteristic of Green's function into the network structure. In this work, we take the two-dimensional problem for illustration and testing of the GF-Net considering computing and memory budgets, but the proposed method can be naturally generalized to higher dimensions.

\subsection{Network architecture}

\begin{figure}[!ht]
  \centering
  \includegraphics[width=0.85\textwidth]{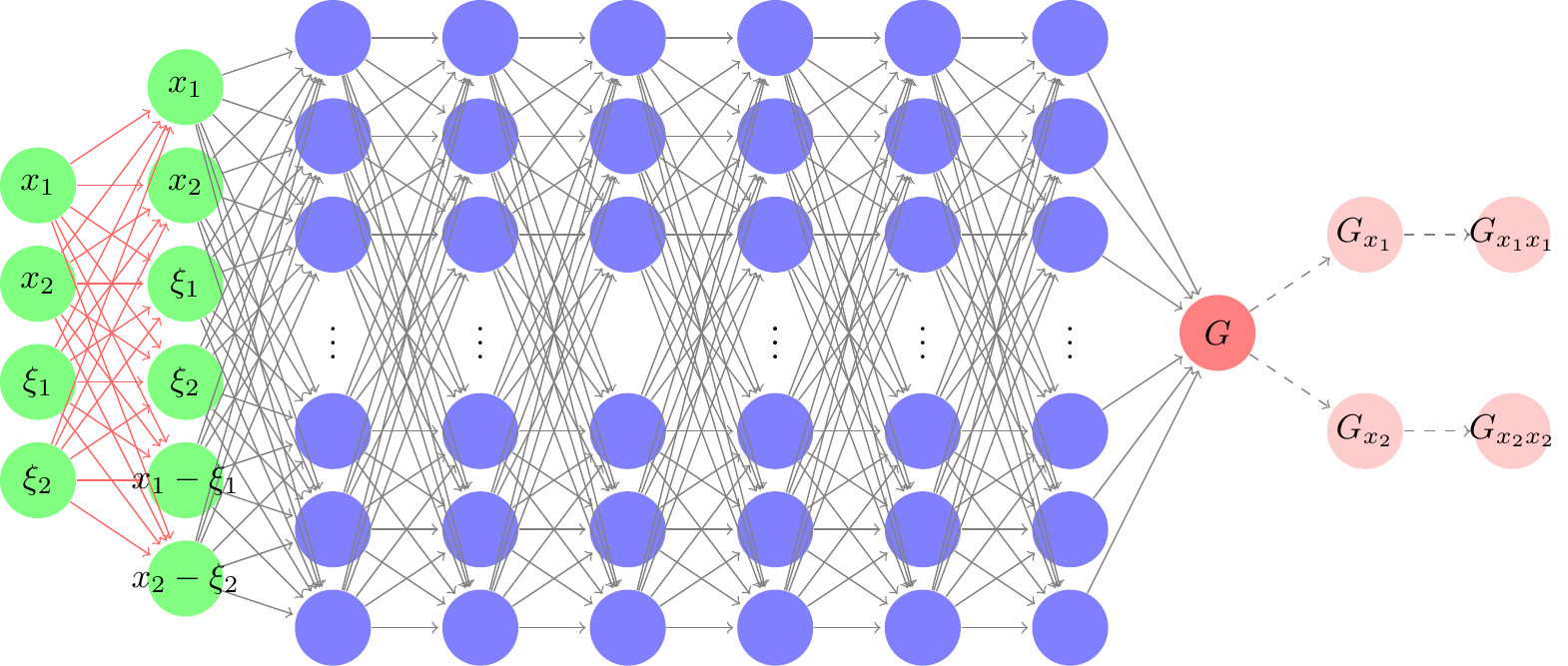}
  \caption{An illustration of the GF-Net architecture with 1 auxiliary layer (green) and 6 hidden layers (blue) for a 2D problem. The derivatives, such as $G_{x_1}, G_{x_1x_1}, \cdots$, can be naturally obtained by applying derivatives on $G$ with automatic differentiation.}
  \label{architecture}\vspace{-0.5cm}
\end{figure}

The architecture of {GF-Net} for a 2D problem is shown in Figure \ref{architecture}. 

The vector $\vv = [\vx, \vxi]^\intercal$ is fed as input to the network, followed by an auxiliary layer without bias and activation:
$$
\ell_0(\vx, \vxi) = \vW^0 \left[\begin{array}{c}\vx \\ \vxi\end{array}\right], ~ \quad
\vW^0 = \left[
  \begin{array}{rr}
    \vI & \vzero\\
    \vzero & \vI\\
    \vI & -\vI
  \end{array}
\right].
$$
This is a preprocessing or auxiliary  layer inspired by the formulation of the Greens' function containing the variables $\vx, \vxi$ and $\vx-\vxi$.  
The layer is then connected to $D-1$ hidden layers and an output layer, which form a fully-connected neural network of depth $D$. Letting $\ell_k$ be the $k$-th layer after the auxiliary layer, then this layer receives an input $\vv^{k-1}$ from the previous layer output and transforms it by an affine mapping to 
\begin{equation}
  \label{eq:affine}
\ell_k(\vv^{k-1}) = \vW^{k} \vv^{k-1} + \vb^k,
\end{equation}
where $\vW^{k}$ is called the connection weight and $\vb^k$ the bias. The nonlinear activation function $\sigma(\cdot)$ is applied to each component of the transformed vector before sending it to the next layer, except the last hidden one. The network thus is a composite of a sequence of nonlinear functions:
\begin{equation}\hspace{-0.0cm}
  \label{eq:nn}
  G(\vx, \vxi;\vTheta) = \left(\ell_D \circ \sigma \circ \ell_{D-1} \circ \cd \circ \sigma \circ \ell_1 \circ \ell_0 \right)(\vx, \vxi),
\end{equation}
where the operator ``$\circ$'' denotes the composition and $\vTheta = \{\vW^k, \vb^k\}_{k=1}^D$ represents the trainable parameters in the network. It should be noted that the weight $\vW^0$ is frozen that needs not to be updated during the training process. 

\subsection{Approximation of the Dirac delta function}
In the network setting, we seek for a classic (smooth) solution satisfying the strong form of the PDE \eqref{eq:Green}. However, $G(\vx,\vxi) $ is not  differentiable everywhere as it is a response to the impulse source defined by the Dirac delta function. Indeed, it can only be well defined in the sense of distribution. 
In practice, we  approximate the Dirac delta function by a multidimensional Gaussian density function:
\begin{equation}\label{eq:GDF}
\rho(\vx-\vxi) = {\frac{1}{(\sqrt{2\pi} s)^2}} e^{-\frac{|\vx-\vxi|^2}{2s^2}},
\end{equation}
where the parameter $s>0$ denotes the standard deviation of  the distribution.
As $s\rightarrow 0$, the function \eqref{eq:GDF} converges to the Dirac delta function pointwisely except at the point $\vx=\vxi$.

\subsection{Sampling strategy for the variable \texorpdfstring{$\vx$}{vx}}\label{sec:x_sample}
Since the GF-Net takes both $\vx$ and $\vxi$ as input, how to sample them in a reasonable manner could be crucial to the training process. In particular, the distribution of $\vx$-samples with respect to different $\vxi$ should vary following the behaviour of $\rho(\vx-\vxi)$: most samples need to be placed within $3$ standard deviations away from the mean of a Gaussian distribution based on the empirical rule. 

\begin{figure}[!ht]  
  \begin{minipage}{0.3\linewidth}
    \centering
    \includegraphics[width=1\textwidth]{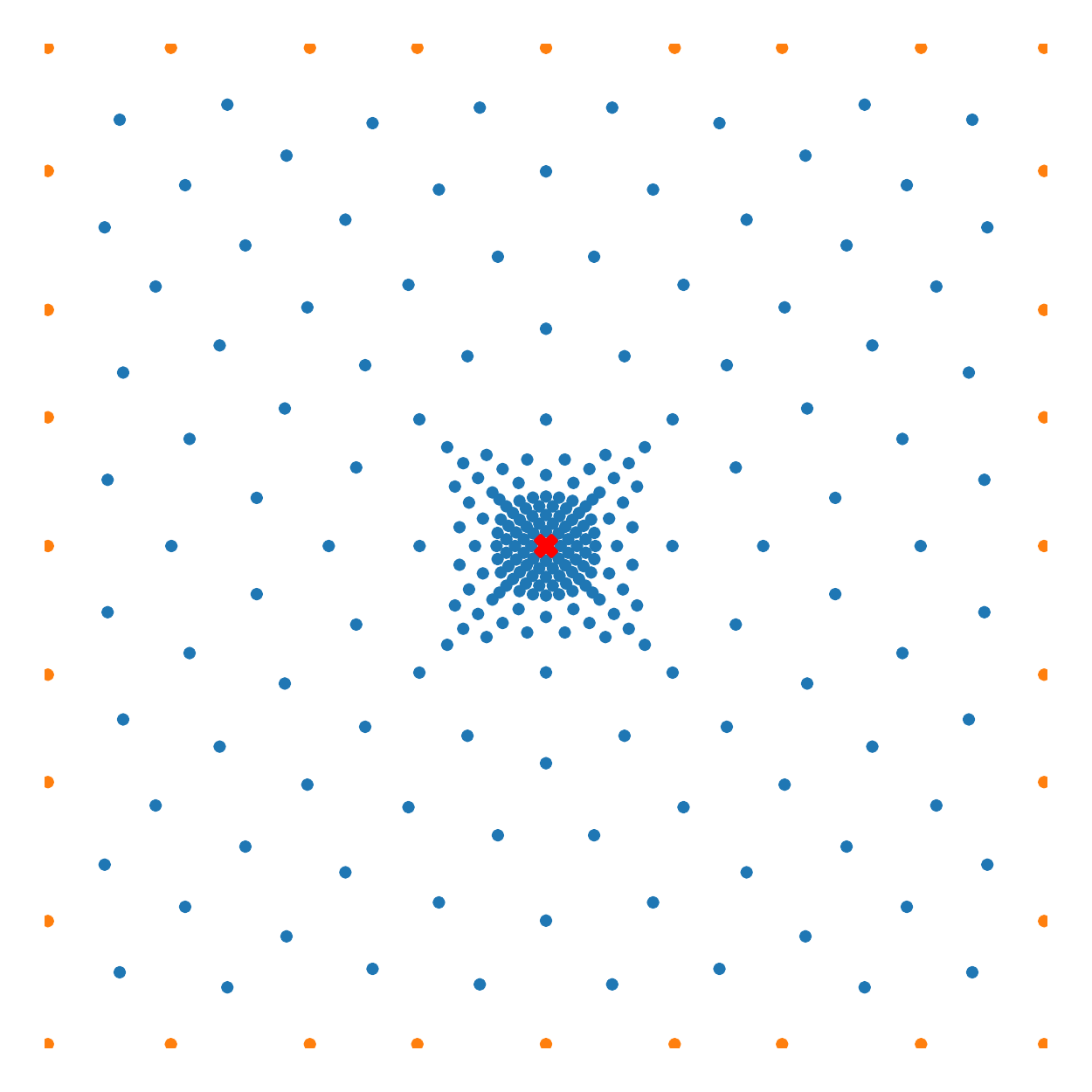}
  \end{minipage}\hfill
  \begin{minipage}{0.3\linewidth}
    \centering
    \includegraphics[width=1\textwidth]{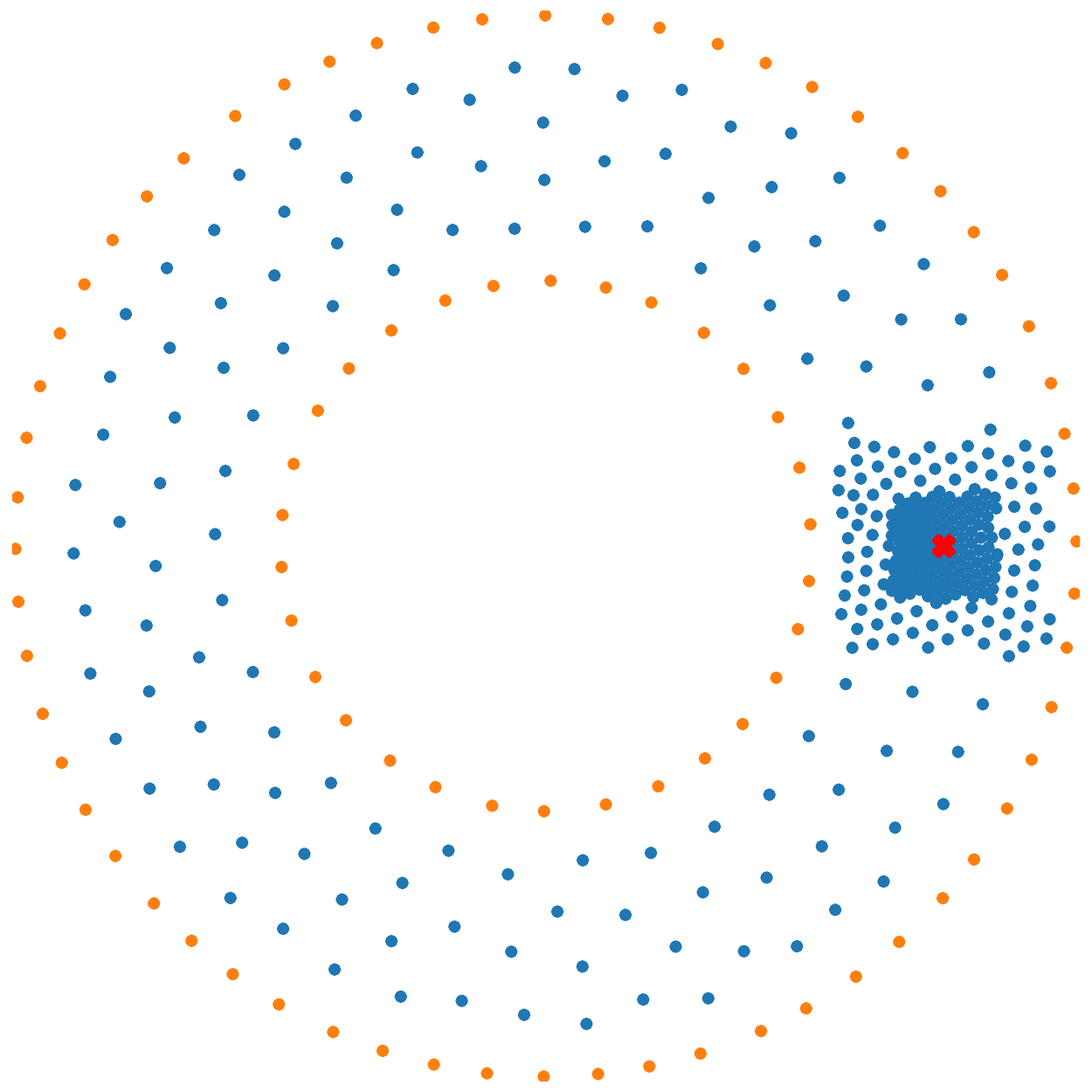}
  \end{minipage}\hfill
  \begin{minipage}{0.3\linewidth}
    \centering
    \includegraphics[width=1\textwidth]{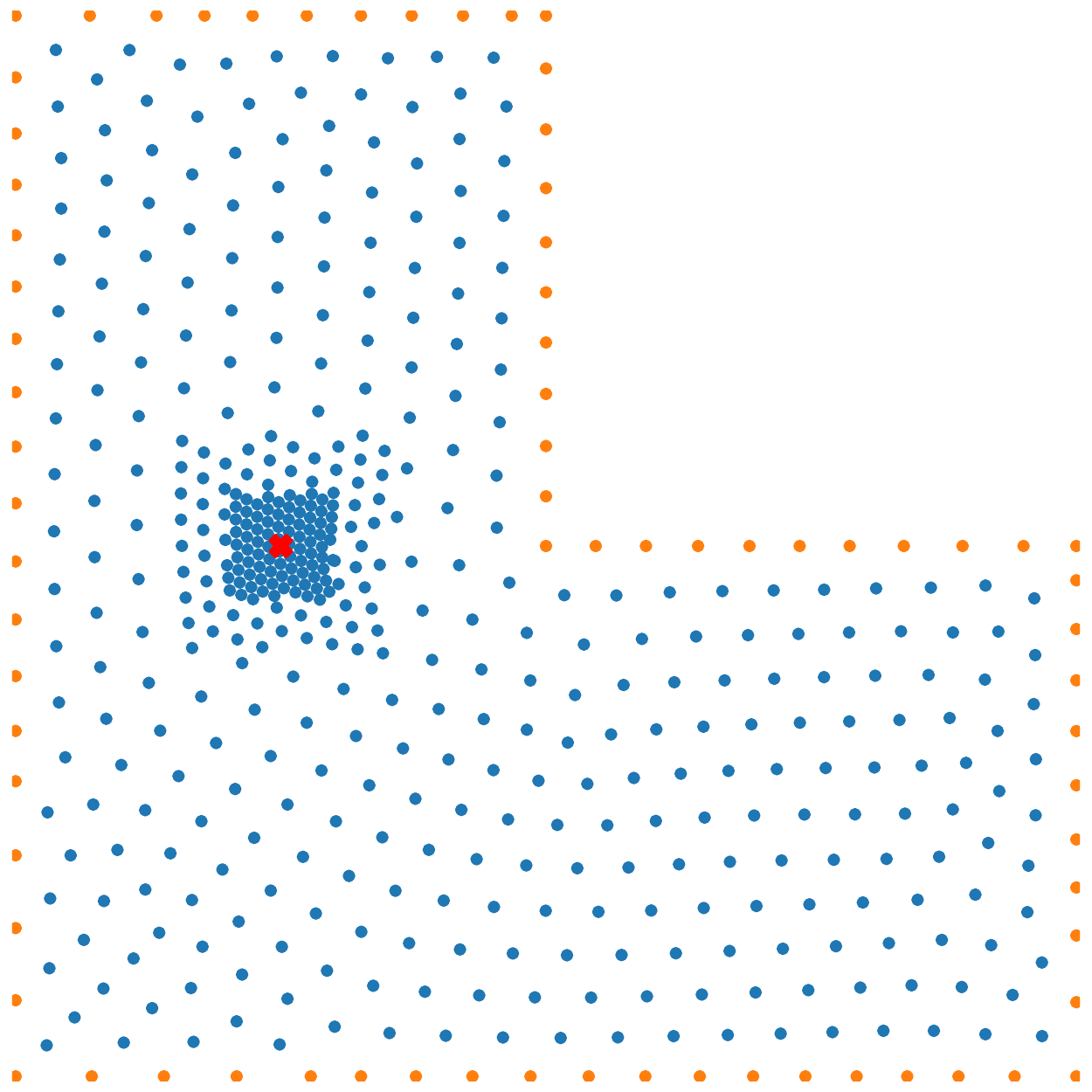}
  \end{minipage}\vspace{0.2cm}
  \caption{Locally refined $\vx$-samples for a given $\vxi$ (marked by the large red cross). Left: $\vxi=(0,0)$ in a square; middle: $\vxi=(\frac{3}{4}, 0)$ in an annulus; right: $\vxi=(-\frac{1}{2}, 0)$ in a L-shaped domain. 
  } \vspace{-0.75cm}
  \label{xSampling}
\end{figure}

To make the sampling  effective, we put forth the following strategy: Since the spatial domain $\Omega$ may have complex geometrical shape, we adopt a mesh generator to first partition $\Omega$ into a triangular mesh $\mathcal T_\vxi = (\mathcal V_\vxi, \mathcal E_\vxi)$ where $\mathcal V_\vxi$ denotes the vertex set and $\mathcal E_\vxi$ denotes the edge set, and then collect 
{\em $\vxi$-samples} from the interior vertices to form $
\mathcal S_\vxi = \{\vxi \in \mathcal V_{\vxi}: \vxi \notin \partial \Omega\}
$. 
For each fixed $\vxi$, we select {\em $\vx$-samples} that concentrate around it because the Gaussian density function centers at this $\vxi$.
Hence, we generate another three meshes $\{\mathcal T_\vx^i = (\mathcal V_\vx^i, \mathcal E_\vx^i)\}_{i=1}^3$ of  resolutions from high to low, and collect $\vx$-samples to form the set 
$
\mathcal S_{\vx,\vxi} = \mathcal S_{\vx, \vxi}^1 \cup \mathcal S_{\vx, \vxi}^2 \cup \mathcal S_{\vx, \vxi}^3,
$
in which
\begin{align*}
\mathcal S_{\vx, \vxi}^1 &= \{\vx \in \mathcal V_\vx^1: \|\vx-\vxi\|_\infty \le c_1 s\}, \\
\mathcal S_{\vx, \vxi}^2 &= \{\vx \in \mathcal V_\vx^2: c_1 s < \|\vx-\vxi\|_\infty < c_2 s\}, \\
\mathcal S_{\vx, \vxi}^3 &= \{\vx \in \mathcal V_\vx^3: \|\vx-\vxi\|_\infty \ge c_2 s\},
\end{align*}
and $c_1, c_2$ are two hyperparameters.
Finally, the overall dataset is selected as
$
\mathcal S = \{(\vx, \vxi): \vxi \in \mathcal S_\vxi, \vx \in \mathcal S_{\vx, \vxi}\}.$
To highlight this sampling strategy, we plot in Figure \ref{xSampling} the $\vx$-samples associated to given $\vxi$ in three types of domains (square, annulus  and L-shaped) considered in numerical tests.

In all experiments, we adopt the mesh generator in \citep{MESHGEN} to generate the sampling points since it is  easily applicable to plenty of commonly used complex domains. Random or pseudo-random sampling methods may have some inconveniences and need extra process in dealing with irregular regions; for instance, the popular Latin hypercube sampling \citep{LHS1979} method plus the rejection procedure \citep{Ross1976} also can be used here.

\subsection{Partitioning strategy for the point source location \texorpdfstring{$\vxi$}{vxi}} \label{sec:xi_sample}

\begin{figure}[!ht]  
  \begin{minipage}{0.3\linewidth}
    \centering
    \includegraphics[width=1\textwidth]{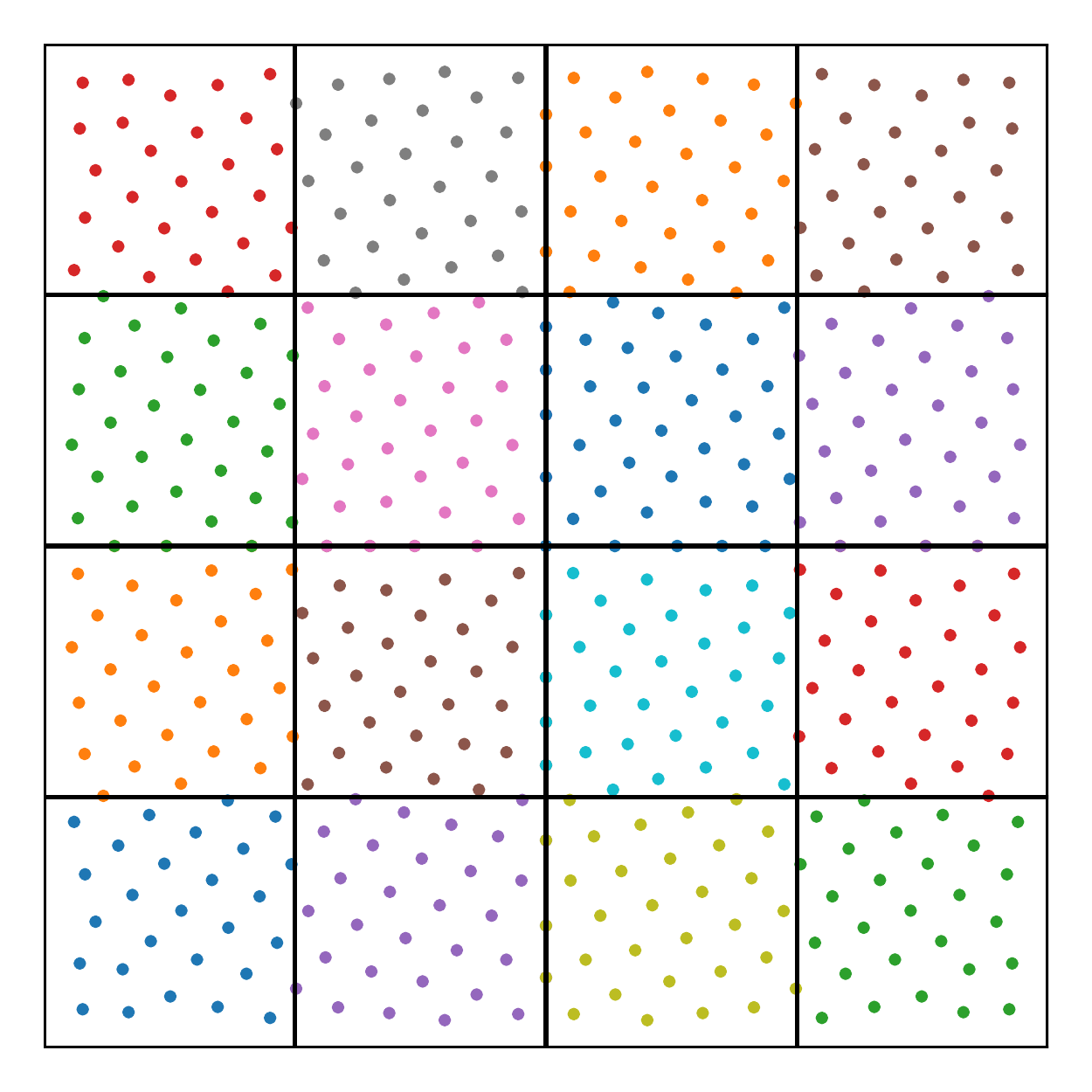}
  \end{minipage}\hfill
  \begin{minipage}{0.3\linewidth}
    \centering
    \includegraphics[width=1\textwidth]{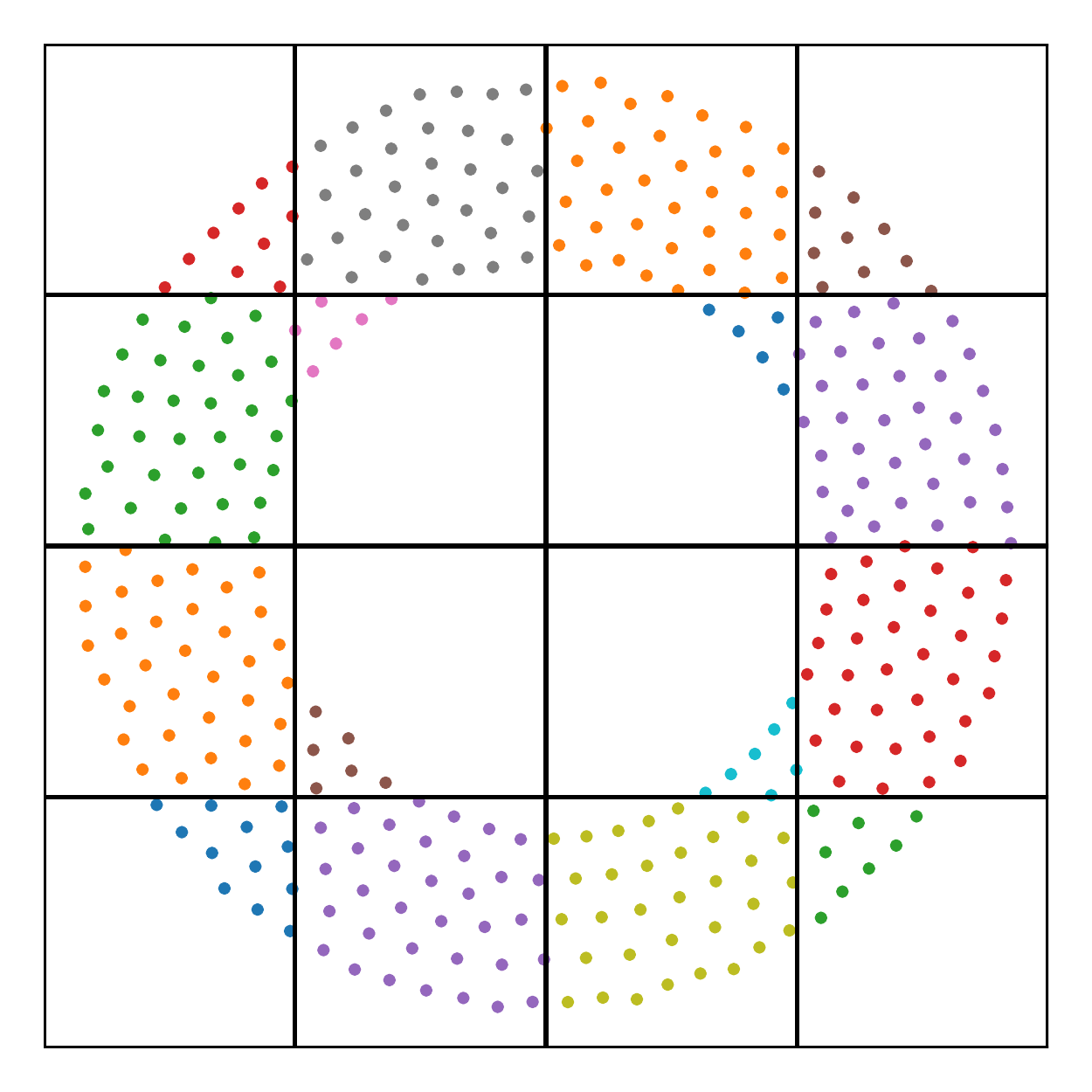}
  \end{minipage}\hfill
  \begin{minipage}{0.3\linewidth}
    \centering
    \includegraphics[width=1\textwidth]{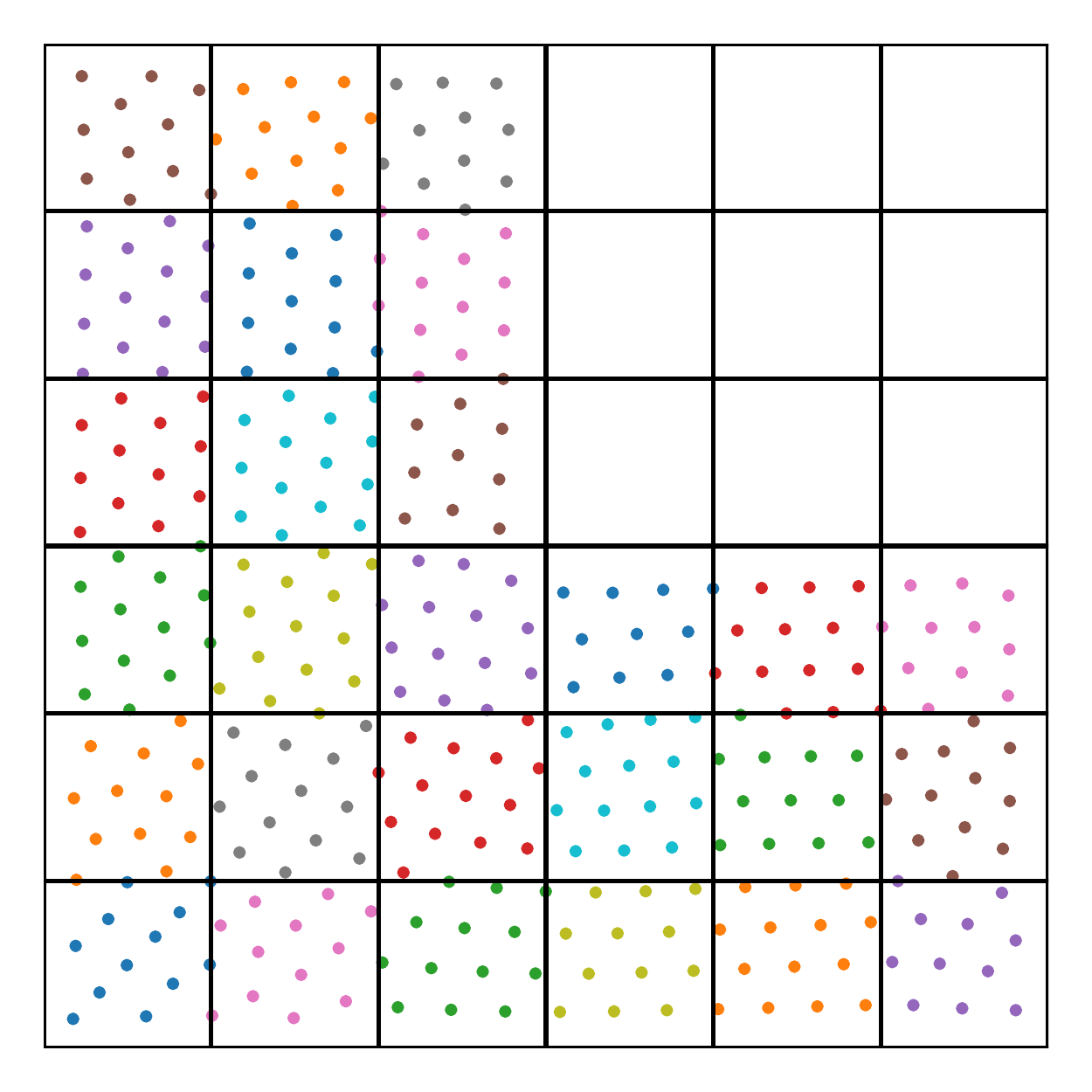}
  \end{minipage}
  \caption{Illustration of uniform sampling on three different domains (square, annulus and L-shaped) and domain partitioning for the point source  location $\vxi$. Left: $4 \times 4$ blocks; middle: $4 \times 4$ blocks;  right: $6 \times 6$ blocks.
  }\vspace{-0.75cm}
  \label{zSampling}
\end{figure}

Ideally, we wish to use one single GF-Net to model the Green's function associated to any $\vxi$ in the domain $\Omega$, but such a network may easily become unmanageable due to a large amount of data in $\mathcal S$, or be very difficult to train as different $\vxi$ may yield totally distinct behaviors of the Green's function. On the other hand,
since the Green's function corresponding to different $\vxi$ can be solved individually, it is feasible to train a GF-Net for each sample $\vxi$, which however would cause the loss of efficiency and result in large storage issues. Therefore, we propose an domain decomposition strategy for $\vxi$ and train a set of GF-Nets on  $\vxi$-blocks. 
Given any target 2D domain $\Omega$, we first identify an circumscribed rectangle of $\Omega$, then divide it into $m\times n$ blocks uniformly. 
Suppose there are $K$ $(\le m\times n)$ blocks containing samples of $\vxi$, we denote the $\vxi$-sample set in the $k$-th block by $\mathcal S_{\vxi}^k$, for $k=1, \ldots, K$. Figure \ref{zSampling} shows the $\vxi$-blocks associated to the three different domains (square, annulus and L-shaped). 
Consequently, we define the sample set associated to the $k$-th $\vxi$-block by 
$\mathcal S^k = \{(\vx, \vxi): \vxi \in \mathcal S_\vxi^k, \vx \in \mathcal S_{\vx, \vxi}\}$.
Based on the new partitioned samples, a set of $K$ GF-Nets will be independently trained. 
The approach has at least two advantages: first, the training tasks are divided into many small subtasks, which are naturally parallelizable and can be distributed to multiple GPUs for efficient implementations; second, locally trained models tend to obtain better accuracy and stronger generalization ability {(for unseen/predicted $\vxi$ located in corresponding blocks)} than the global single model. 

\subsection{Loss function}
Define the training data for the set of GF-Nets: 
$
\mathcal S^k = \mathcal S_c^k \cup \mathcal S_b^k,
$
where
$\mathcal S_c^k = \{(\vx, \vxi) \in \mathcal S^k: \vx \notin \partial \Omega\}$ and 
$\mathcal S_b^k = \{(\vx, \vxi) \in \mathcal S^k: \vx \in \partial \Omega\}$, for $k=1, \ldots, K$. 
The $k$-th GF-Net is trained by minimizing the following total loss: 
\begin{equation}
  \label{eq:loss}
  L(\vTheta_k) = L_{res}(\vTheta_k) + \lambda_{b} L_{bdry} (\vTheta_k) + \lambda_{s} L_{sym}(\vTheta_k),
\end{equation}
where
$$  
  \begin{aligned}
  L_{res}(\vTheta_k) &= \ds \frac1{|\mathcal S_{c}^k|} \sum_{(\vx,\vxi)\in\mathcal S_c^k}  \left[\mathcal L G(\vx, \vxi; \vTheta_k) - \rho(\vx, \vxi)\right]^2, \\
  L_{bdry}(\vTheta_k) &= \ds \frac1{|\mathcal S_{b}^k|} \sum_{(\vx,\vxi)\in\mathcal S_b^k}  \left[G(\vx, \vxi; \vTheta_k) \right]^2,\\
    L_{sym}(\vTheta_k) &= \ds \frac1{|\mathcal S_{c}^k|} \sum_{(\vx,\vxi)\in\mathcal S_c^k} \left[G(\vx, \vxi; \vTheta_k) - G(\vxi, \vx; \vTheta_k) \right]^2. 
\end{aligned}
$$
Here, $L_{res}(\vTheta_k)$ represents the pointwise PDE residual defined for each $(\vx,\vxi)$ pair, $L_{bdry}$ measures the errors on boundary, $L_{sym}(\vTheta_k)$ is introduced to enforce the intrinsic symmetry property of the Green's function, and $\lambda_b$ and $\lambda_s$ are two hyperparameters for balancing the three terms. Note that the loss function \eqref{eq:loss} 
does not use any ground-truth data (i.e., the exact or certain approximate solution of the problem \eqref{eq:Green}).

\section{A Fast Numerical Solver using GF-Nets}
\begin{figure}[!ht]
\centering
\begin{minipage}{.37\linewidth}
\includegraphics[width=0.9\textwidth]{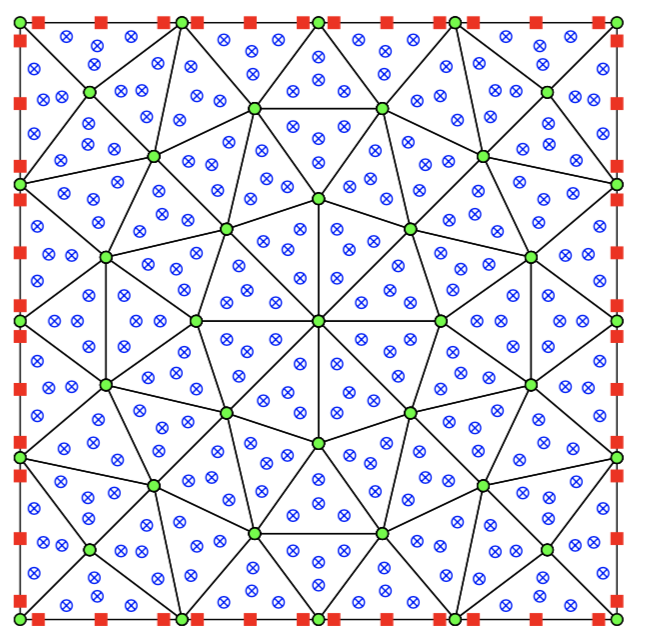}
\end{minipage}
\hspace{.3cm}
\begin{minipage}{.5\linewidth}
\includegraphics[width=0.85\textwidth]{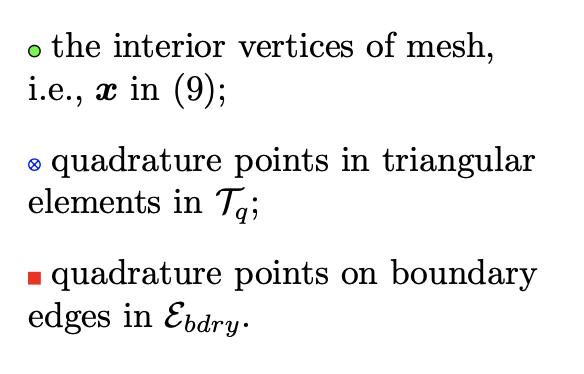}
\end{minipage}
\caption{An illustration of the triangular mesh $\mathcal T_q$ and the quadrature points.}
\label{fig:Quadrature}\vspace{-0.4cm}
\end{figure}

After training the set of GF-Nets, numerical solutions of the linear reaction-diffusion problem \eqref{eq:dr} can be directly computed based on the formula \eqref{eq:rd-soln} using GF-Nets. In order to evaluate the integrals in \eqref{eq:rd-soln} accurately, we apply numerical quadrature on triangular meshes. To this end, we generate a triangulation for the domain $\Omega$ consisting of triangles $\mathcal T_q=\{T_l\}$. Denote the intersection of the triangle edges with the domain boundary by ${\cal E}_q^{bdry} = \{E_m\}$. For any $\vx\in \Omega$, we have
  \begin{equation}
  u(\vx) \approx \sum_{T_l\in {\cal T}_q} I^{T_l}_{\vxi,h}\big[f(\vxi) G(\vxi, \vx)\big] - \sum_{E_m\in {\cal E}_q^{bdry}} I^{E_m}_{\vxi,h}\big[g(\vxi) a(\vxi)({\nabla_{\vxi} G}(\vxi,\vx)\cdot \nn_{\vxi})\big],
 \label{eq:rd-numsoln}
 \end{equation}
where $I^{T_l}_{\vxi,h}[\cdot]$ denotes the numerical quadrature for evaluating $\int_{T_l} f(\vxi)  G(\vxi, \vx) \,d\vxi$ and 
 $I^{E_m}_{\vxi,h}[\cdot]$ the quadrature for evaluating $\int_{E_m} g(\vxi)a(\vxi)({\nabla_{\vxi} G}(\vxi,\vx)\cdot \nn_{\vxi})  \,dS(\vxi)$, respectively. 
 For clarity, in Figure \ref{fig:Quadrature}, we plot the quadrature points used in evaluating $I^{T_l}_{\vxi,h}[\cdot]$ by a $4$-point Gaussian quadrature rule for triangular elements and $I^{E_m}_{\vxi,h}[\cdot]$ by a $3$-point Gaussian quadrature rule on boundary segments. Due to the symmetry of $G(\vxi, \vx)$, the formula \eqref{eq:rd-numsoln} can also be rewritten as an integration with respect to $\vx$ instead of $\vxi$. The algorithm for solving the PDE problem \eqref{eq:dr} with GF-Nets is summarized in Algorithm \ref{alg:GF-Nets}.

\begin{algorithm}[htbp]
  \caption{Solving the model problem \eqref{eq:dr} by GF-Nets}\label{alg:GF-Nets}
  \KwIn{ $\cL(\cdot)$, $f(\vx)$ and $g(\vx)$, the mesh $\mathcal T_q$, and an interior vertex $\widehat{\vx}\in\Omega$}  
  \KwOut{The PDE solution at $\widehat{\vx}$: $u(\widehat{\vx})$}
  \begin{algorithmic}[1]
   \STATE Generate sampling points $ \mathcal S_{\vxi}$ and $\mathcal S_{\vx,\vxi}$.
   \STATE Apply the domain partition to divide $\Omega$ into $K$ blocks.
  \FOR{$k=1$ {\bfseries to} $K$}
  \STATE Collect the dataset in the $k$-th $\vxi$-block $\mathcal S^k=\{(\vx, \vxi): \vxi \in \mathcal S_\vxi^k, \vx \in \mathcal S_{\vx, \vxi}\}$.
  \STATE Train the GF-Nets $G(\vx, \vxi;\vTheta_k)$ by feeding all $(\vx, \vxi) \in \mathcal S^k$.\
  \ENDFOR
  \STATE Check index $k$ of the $\vxi$-block which $\widehat{\vx}$ locates in, $u(\widehat{\vx})$ is computed from \eqref{eq:rd-numsoln} using the $G(\vxi, \vx;\vTheta_k)$.
  \end{algorithmic}

\end{algorithm}

\section{Experimental results}
In this section, we will investigate the performance of the proposed GF-Nets for approximating the Green's functions \eqref{eq:Green} and its application for fast solution of the linear reaction-diffusion problem \eqref{eq:dr} by Algorithm \ref{alg:GF-Nets}.

\subsection{Model parameters setting}
Each GF-Net (associated with one block) in the experiments has $1$ auxiliary layer and $6$ hidden layers with $50$ neurons per layer, $\sigma(\vx) = \sin(\vx)$ is used as the activation function, $\lambda_b = 400$ and $\lambda_s = 1$ are taken in the loss function if not otherwisely specified, and $s=0.02$ is used in the Gaussian density function for approximating the Dirac delta function. For generating the training sample sets, we choose $c_1=5$ and $c_2=10$.
Both the Adam and LBFGS optimizers are used in the training process. 
The purpose of the former is to provide a good initial guess to the latter. 
The Adam is run for up to $2 \times 10^4$ steps with the training loss tolerance $\epsilon_1=0.5$ for possible early stopping, which is then followed by the LBFGS optimization for at most $1\times 10^4$ steps with the loss tolerance $\epsilon_2 =1\times 10^{-4}$. 
The same settings are used in training all GF-Nets for ensuring them to possess the same level of accuracy. 

To test the ability of the GF-Net for approximating the Green's functions, we consider both the Poisson's (pure diffusion) equations and a reaction-diffusion equation in the square $\Omega_1=[-1,1]^2$, the annulus $\Omega_2= B_1(\vzero)\backslash B_{1/2}(\vzero)$ with $B_r(\vzero)$ denoting a circle centered at the origin with radius $r$, and the L-shaped domain $\Omega_3= [-1, 1]^2\backslash [0,1]^2$, as shown in Figure \ref{zSampling}. 
The numerical solutions of the PDEs at all the interior vertices $\mathcal V_q$ of the triangulation $\mathcal T_q$ are used for quantifying the performance, which are
evaluated by the relative error in the $l_2$ norm:
$$
\begin{aligned}
Error &= \frac{\big(\sum_{\vx \in \mathcal V_q} \left|u_{e}(\vx) - u_p(\vx)\right|^2 A_\vx\big)^{1/2}}{\big(\sum_{\vx \in \mathcal V_q} \left|u_{e}(\vx)\right|^2 A_\vx\big)^{1/2}},
\end{aligned}
$$
where $A_\vx$ is the area of the dual cell related to the vertex $\vx$, $u_e$ and $u_p$ denote the exact solution and  approximate solution, respectively. 
All the experiments reported in this work are performed on an Ubuntu 18.04.3 LTS desktop with a 3.6GHz 
Intel Core i9-9900K CPU, 64GB DDR4 memory and Dual NVIDIA RTX 2080 GPUs. 

\subsection{Ablation study based on the Green's function with a fixed point source}\label{sec:unitdisk}

In this subsection, we choose the classic Poisson's equation in a unit disk with a fixed point source $\vxi$ located at its center (i.e., the origin) to conduct ablation study of the proposed GF-Net, including the learning accuracy and the effect of using the symmetric loss term $L_{sym}$ in \eqref{eq:loss}. Furthermore, detailed discussions about the choice of activation function and the effect of the  auxiliary layer are presented in  Appendix \ref{apdx:UnitDisk}. The analytic form of this specific Green's function is given as below:
$$
G_{\mathrm{e}}(\vx, \vzero) = -\frac{1}{4\pi} \ln{( x_1^2+x_2^2 )}, \quad \vx \in B_1(\vzero). 
$$
Since the point source $\vxi$ is fixed at the origin, only $\vx$ needs to be sampled for training and we still adopt sampling strategy in Subsection \ref{sec:x_sample} to obtain $\mathcal S_{\vx,\vzero}$, where three different levels of meshes $\{\mathcal T_{\vx}\}_{i=1}^3$ with $\#\mathcal V_\vx^1=31213, \#\mathcal V_\vx^2=7842, \#\mathcal V_\vx^3=1982$ are used. No domain partition is used since there is only one source point in this case.

\begin{figure}[!ht]
  \centering
\subfigure[\texttt{$G_{\mathrm{e}} (\vx, \vzero)$}]{\hspace{-0.4cm}
\begin{minipage}[t]{0.3\linewidth}
\centering
\includegraphics[width=\textwidth]{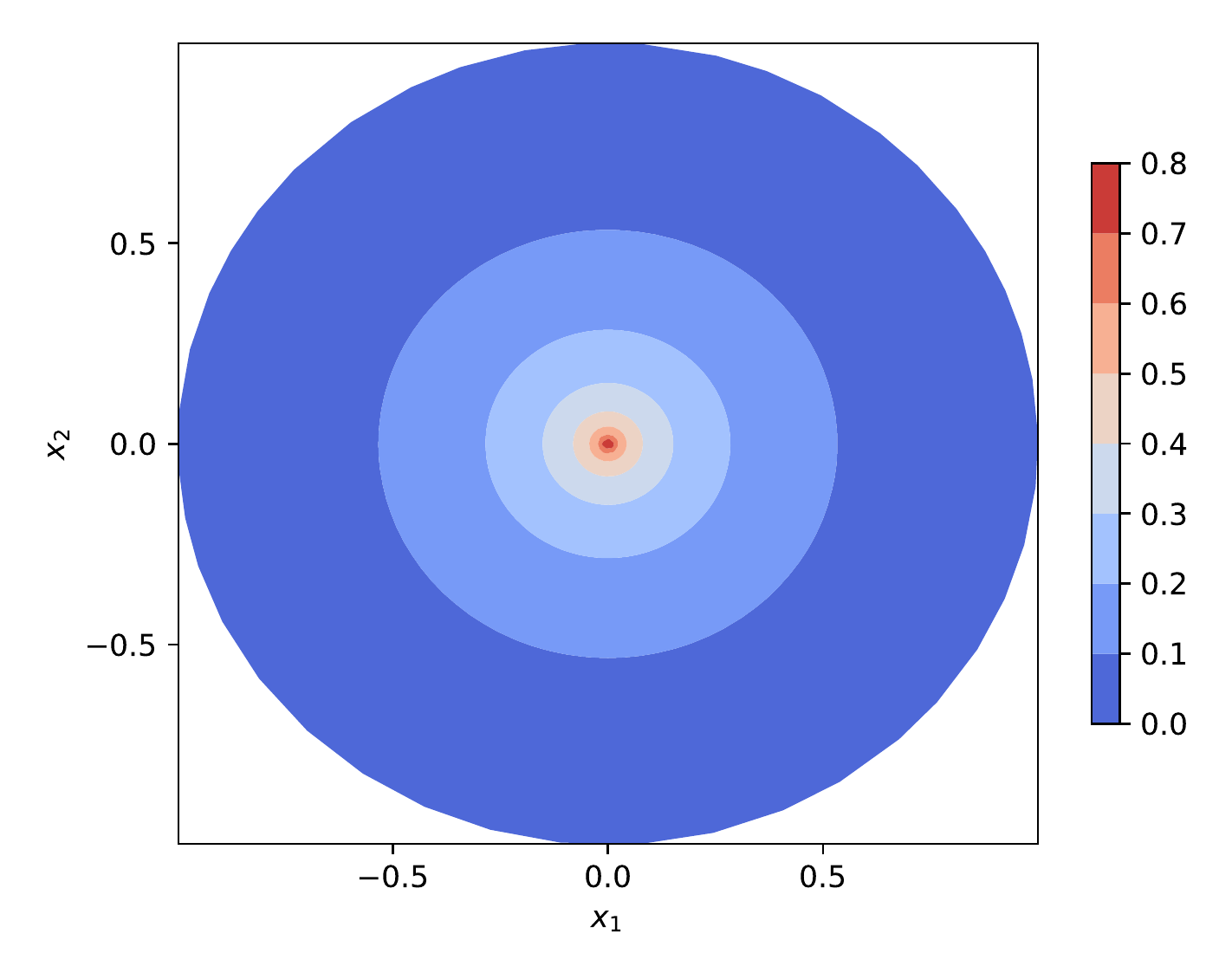}
 \end{minipage}
}
\subfigure[\texttt{$G_{\mathrm{a}} (\vx, \vzero)$}]{
\begin{minipage}[t]{0.3\linewidth}
\centering
\includegraphics[width=\textwidth]{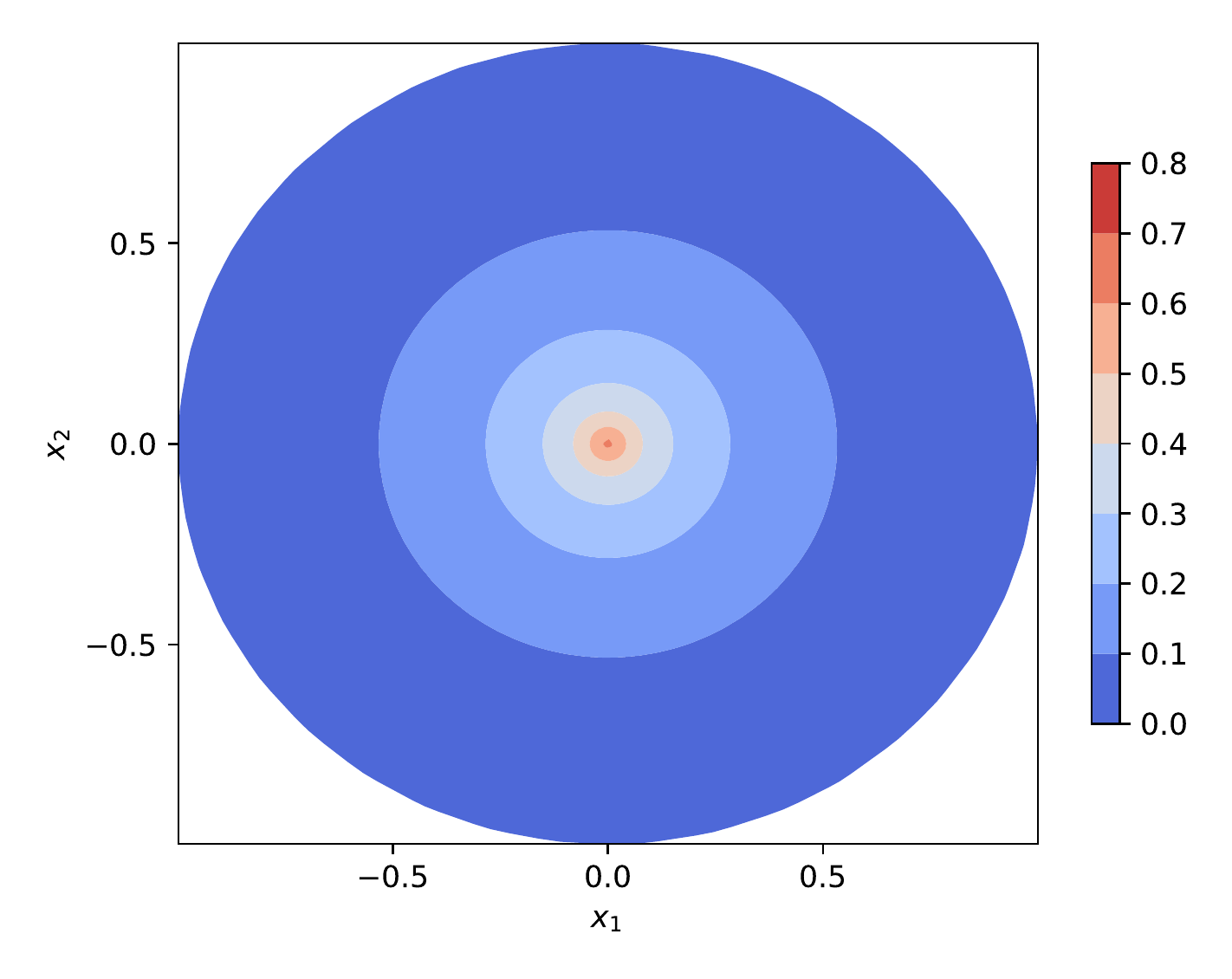}
  \end{minipage}
}
\subfigure[\texttt{$|G_{\mathrm{a}}(\vx, \vzero)-G_{\mathrm{e}} (\vx, \vzero)|$}]{
\begin{minipage}[t]{0.3\linewidth}
\centering
\includegraphics[width=\textwidth]{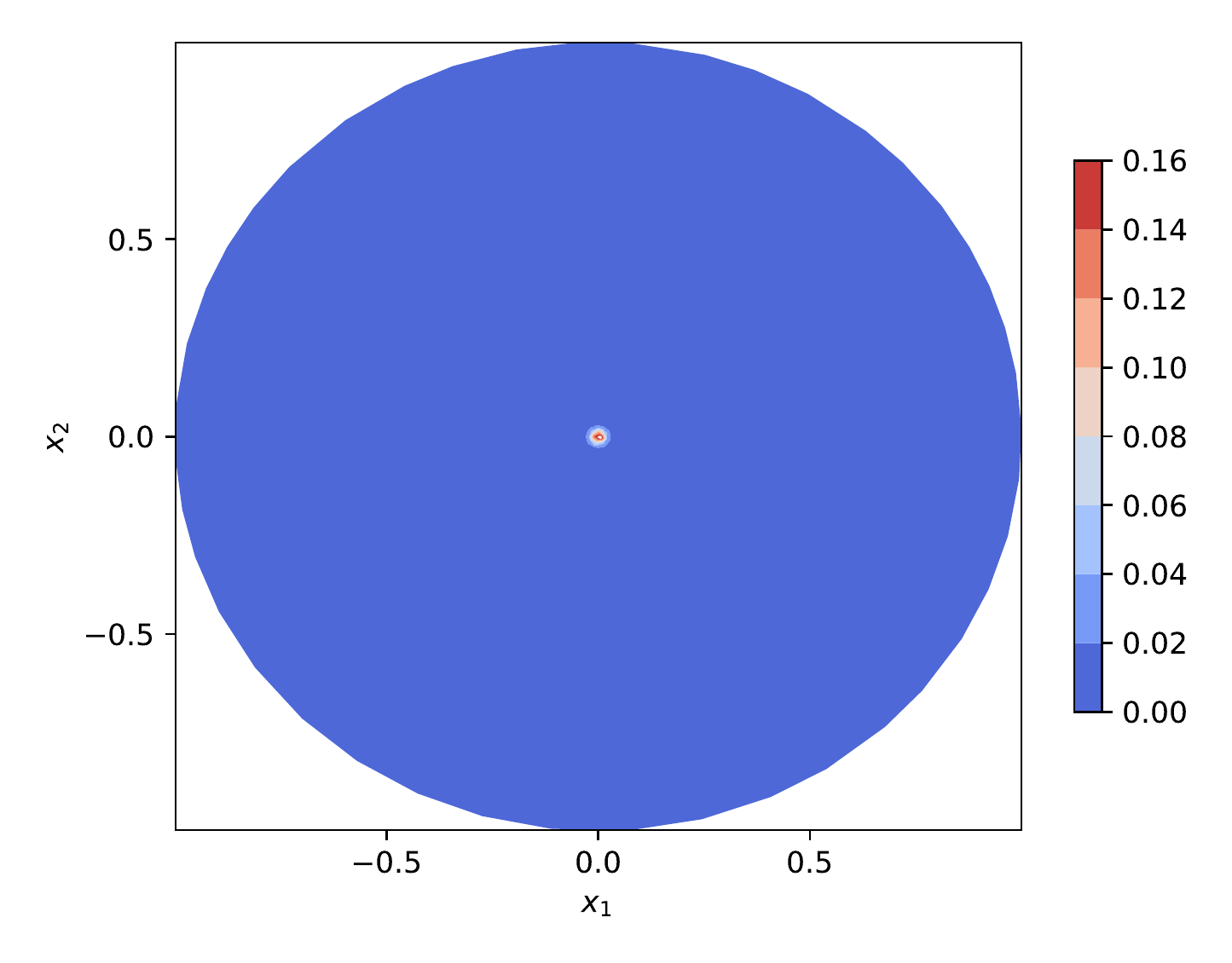}
  \end{minipage}
}
  \caption{Numerical results for the Green's function in the unit disk with the point source at the origin: the exact solution  (left), the predicted solution by GF-Net (middle) and the numerical  error (right).}
  \label{fig:GFPoissonSingle}\vspace{-0.4cm}
\end{figure}

\paragraph{Accuracy of learned Green's function}
To measure the accuracy of the predicted Green's function produced by GF-Net, we calculate its relative $l_2$ error in $B_{\backslash\epsilon}(\vzero) = B_1(\vzero) \backslash B_\epsilon(\vzero)$ with $\epsilon=3s$ to avoid the singularity at the origin. 
Quantitatively, we find the relative $l_2$ error is only about $1.29 \times 10^{-3}$, which demonstrate very good performance of the proposed GF-Net for predicting the Green's function. Figure \ref{fig:GFPoissonSingle} plots qualitative comparison of the Green’s function and its prediction by GF-Net, from which we see that the error mainly concentrates at the origin (the point source location) and decays rapidly to the boundary.

\begin{figure}[!ht]
  \centering
\subfigure[\texttt{$G_{\mathrm{e}}(\vzero,\vx)$}]{\hspace{-0.4cm}
\begin{minipage}[t]{0.3\linewidth}
\centering
\includegraphics[width=\textwidth]{./image/Poisson_circle_SingleSource_Gexact_hm.pdf}
 \end{minipage}
}
\subfigure[\texttt{$G_{\mathrm{a}}(\vzero,\vx)$ with $\lambda_s=1$}]{
\begin{minipage}[t]{0.3\linewidth}
\centering
\includegraphics[width=\textwidth]{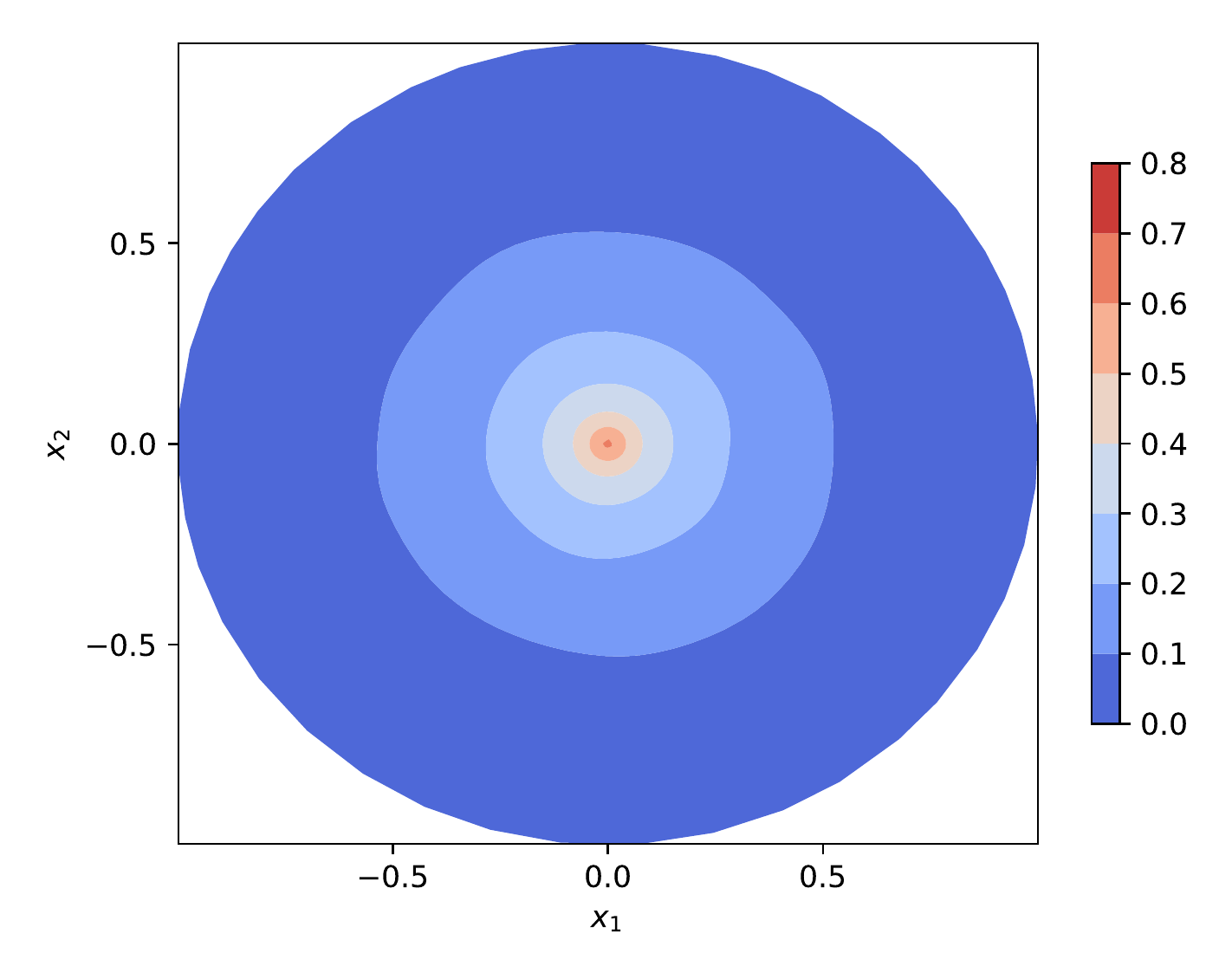}
  \end{minipage}
}
\subfigure[\texttt{$G_{\mathrm{a}}(\vzero,\vx)$ with $\lambda_s=0$}]{
\begin{minipage}[t]{0.3\linewidth}
\centering
\includegraphics[width=\textwidth]{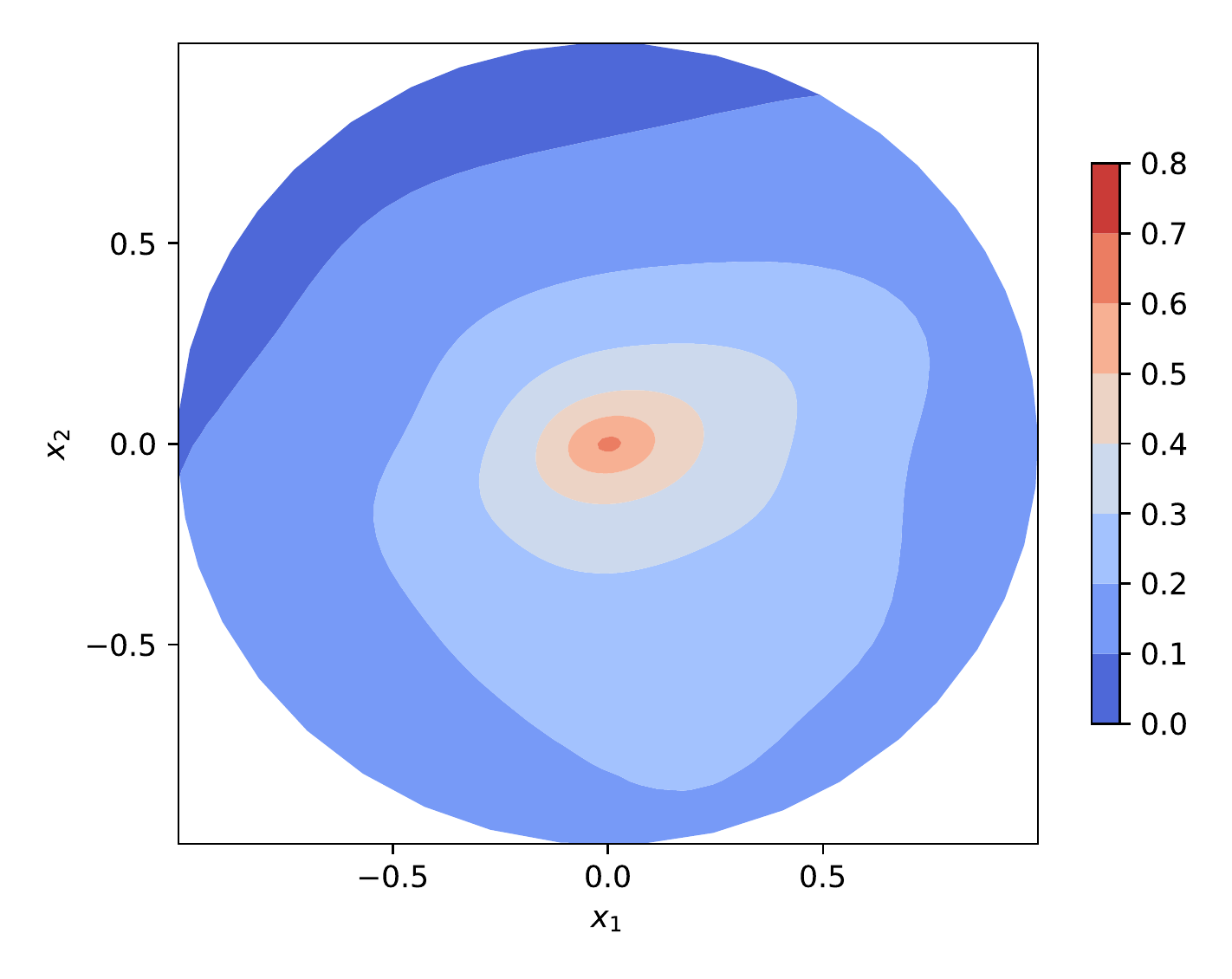}
  \end{minipage}
}
\caption{Numerical results for the {symmetric property of} Green's function in the unit disk with the point source at the origin: the exact solution  (left) and the predicted solutions  by GF-Net with (middle) and without (right) the symmetric loss, respectively.}
  \label{fig:GFPoissonSingle_symm}\vspace{-0.5cm}
\end{figure}

\paragraph{Effect of the symmetric loss}
Since the Green's function is symmetric about $\vx$ and $\vxi$, i.e., $G(\vx,\vxi)=G(\vxi,\vx)$ and such symmetric property is crucial to the quadrature formula \eqref{eq:rd-numsoln}, we introduce a symmetric error term into the total loss \eqref{eq:loss} so that the predicted Green's function by GF-Net could preserve this property. 
To test the effect of this term to the learning outcome, we train the model in two ways: one includes the  symmetric loss with $\lambda_s=1$, and the other excludes the symmetric loss from \eqref{eq:loss}. 
The test results are  shown in Figure \ref{fig:GFPoissonSingle_symm}, from which we see that adding the symmetric loss clearly improves the accuracy of predicted Green's function.

\subsection{Ablation study based on Poisson's equation with homogeneous and inhomogeneous boundary conditions}

In this subsection, we investigate the effect of  the domain partitioning strategies and the choice of the Gaussian parameter $s$ on the  performance of the proposed GF-Net. For testing purpose, we consider the Poisson's equation (i.e., the pure diffusion case with $a(\vx)=1$ and $r(\vx)=0$ in \eqref{operator}) in the square  $\Omega_1$. Both homogeneous and inhomogeneous Dirichlet boundary conditions are considered as  below:
\begin{align*}
    \textit{Case I (homogeneous BC)}: \quad &u(x_1, x_2) =\sin(2\pi x_1)\sin(2\pi x_2),\\
    \textit{Case II (inhomogeneous BC)}: \quad &u(x_1, x_2) =\cos(\pi x_1)\cos(\pi x_2).
\end{align*}
The source term $f(x_1, x_2)$ and the boundary values $g(x_1, x_2)$ are accordingly imposed to match the exact solution for interior and boundary points. $s=0.02$ and $s=0.015$ are used to examine how approxiamtion of Dirac delta function would affect the GF-Net.
To train our GF-Net, we choose the mesh $\mathcal T_\vxi$ with $\#\mathcal V_\vxi=545$ for $\vxi$-samples and the related $\vx$-samples are selected from three meshes $\{\mathcal T_{\vx}^i\}_{i=1}^3$ with $\#\mathcal V_\vx^1=32753, \#\mathcal V_\vx^2=8265, \#\mathcal V_\vx^3=2105$ to generate the sampling point set $\mathcal S$. 

\begin{figure}[!ht]
\centerline{\hspace{-0.3cm}
\subfigure[Exact solution]{
\begin{minipage}[t]{0.31\linewidth}
\centering 
\includegraphics[width=\textwidth]{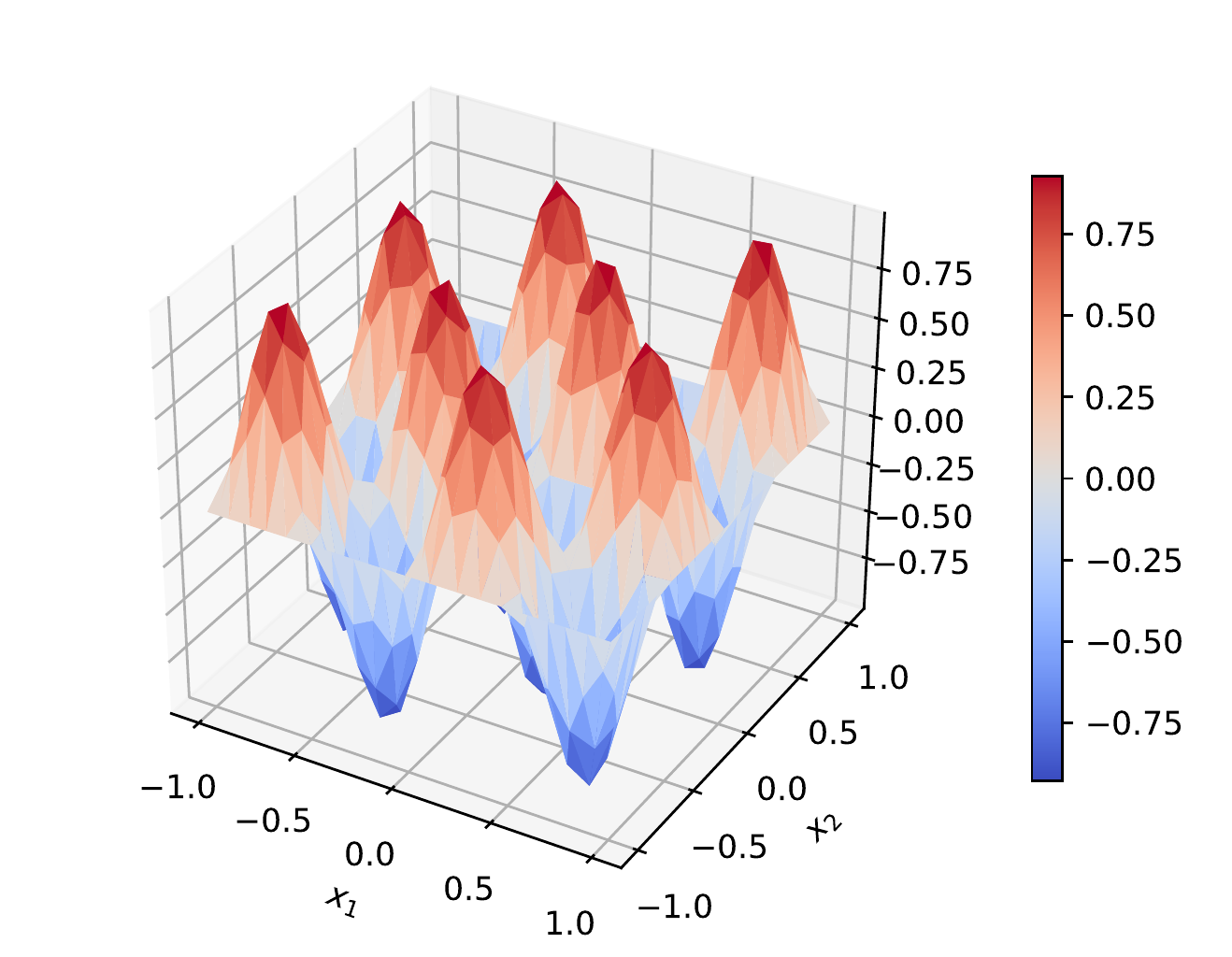}\\
 \includegraphics[width=\textwidth]{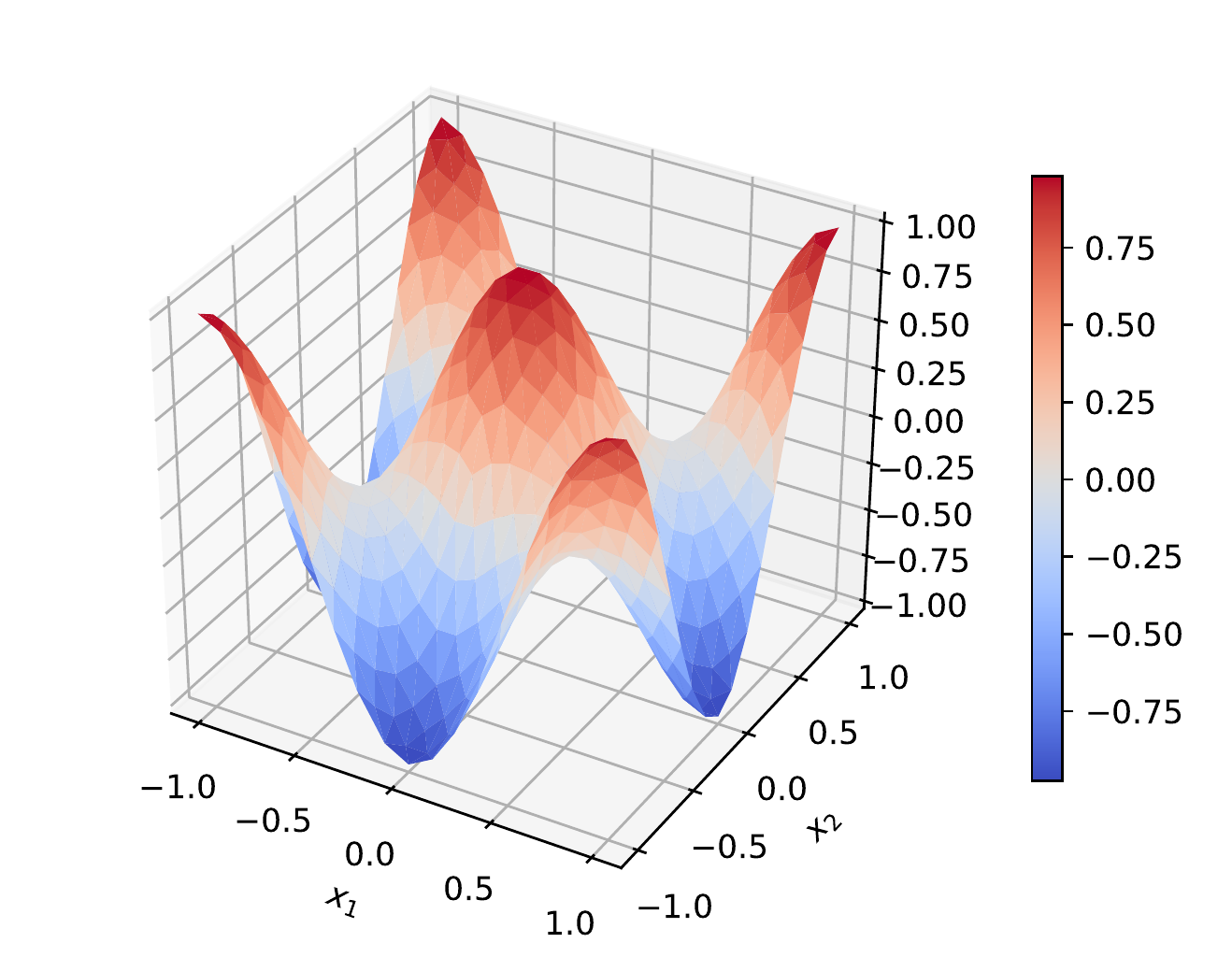}
 \end{minipage}
}
\subfigure[Predicted solution]{
\begin{minipage}[t]{0.31\linewidth}
\centering 
\includegraphics[width=\textwidth]{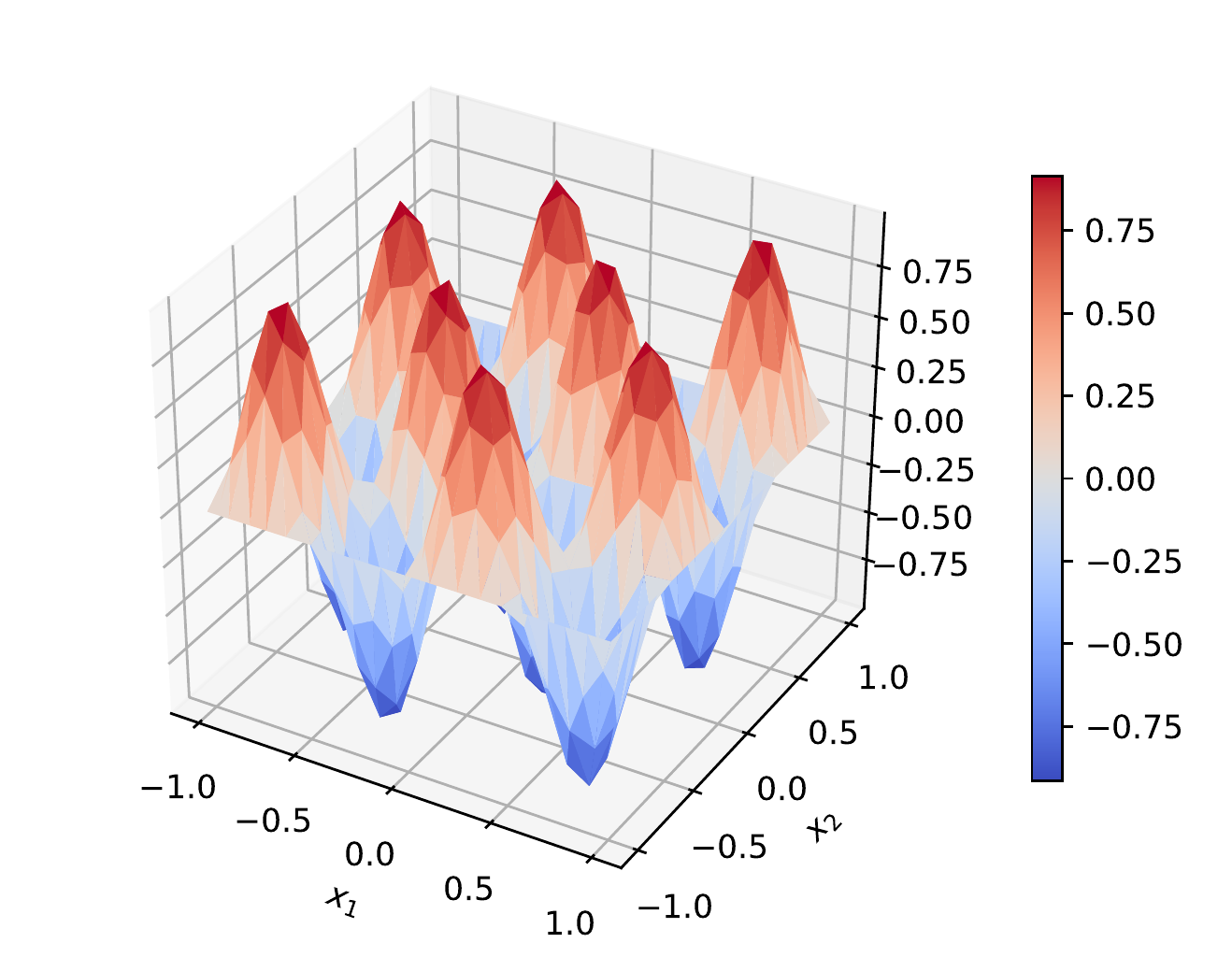}\\
\includegraphics[width=\textwidth]{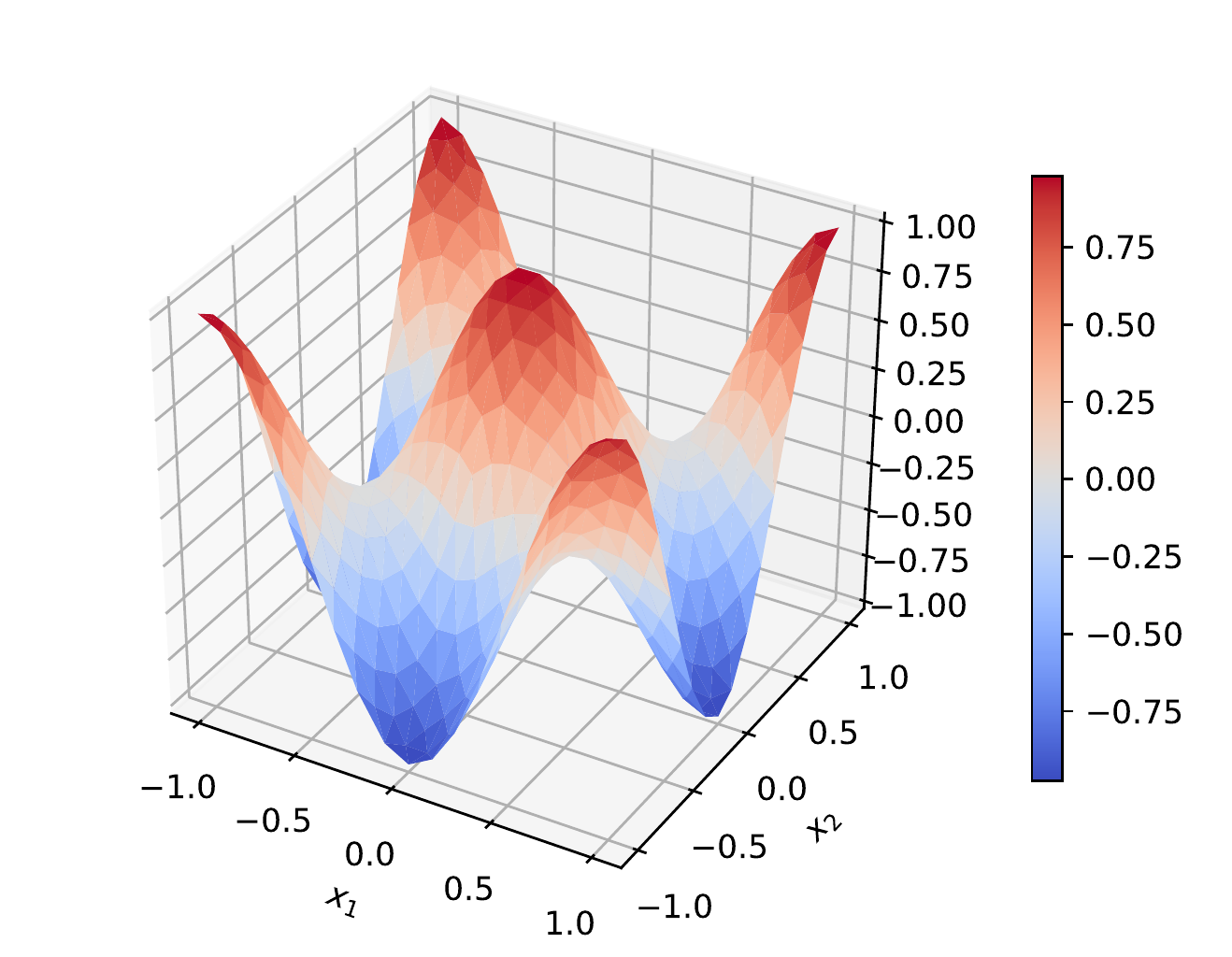}
\end{minipage}
}
\subfigure[Error]{
\begin{minipage}[t]{0.3\linewidth}
\centering   
\includegraphics[width=\textwidth]{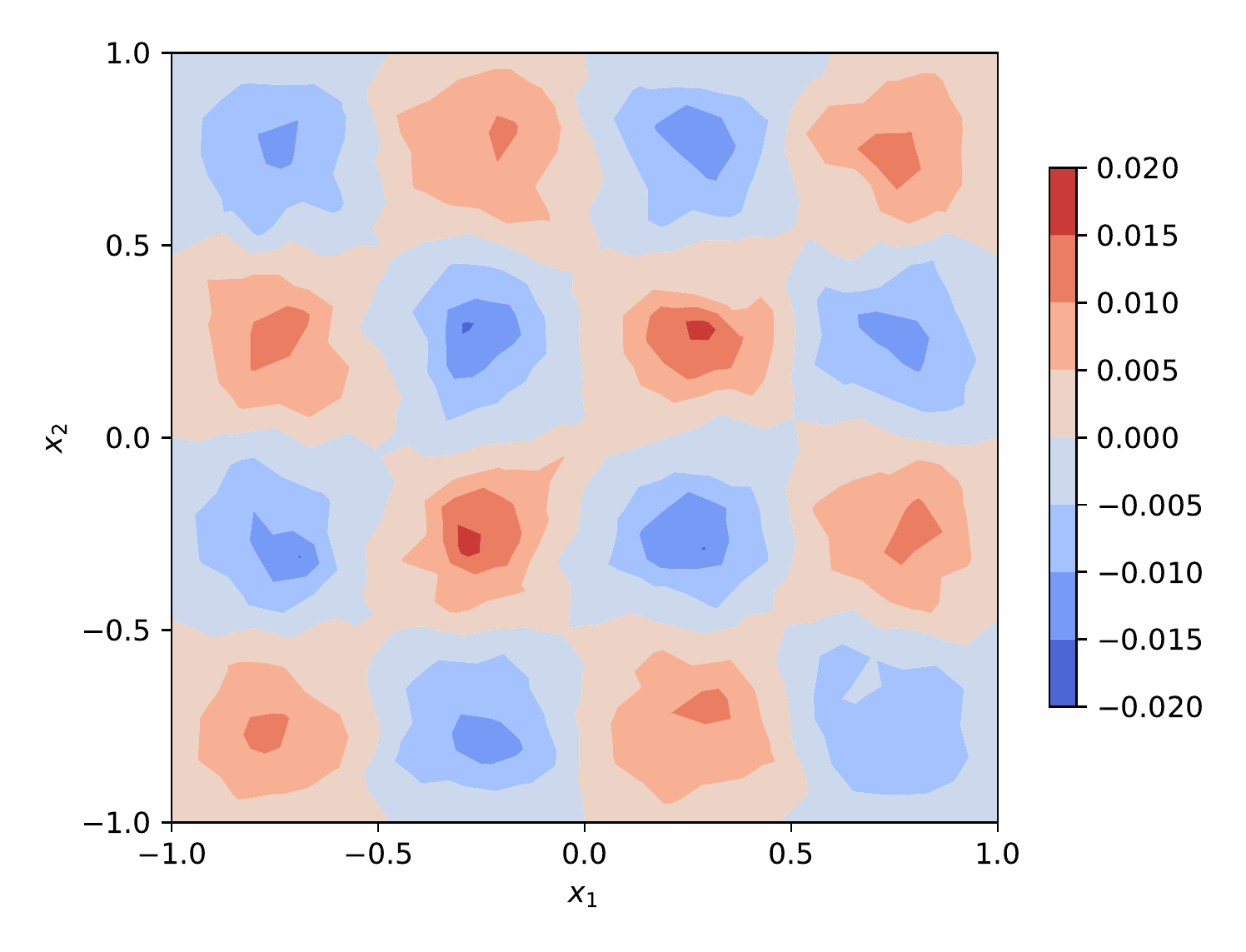}\\ 
\includegraphics[width=\textwidth]{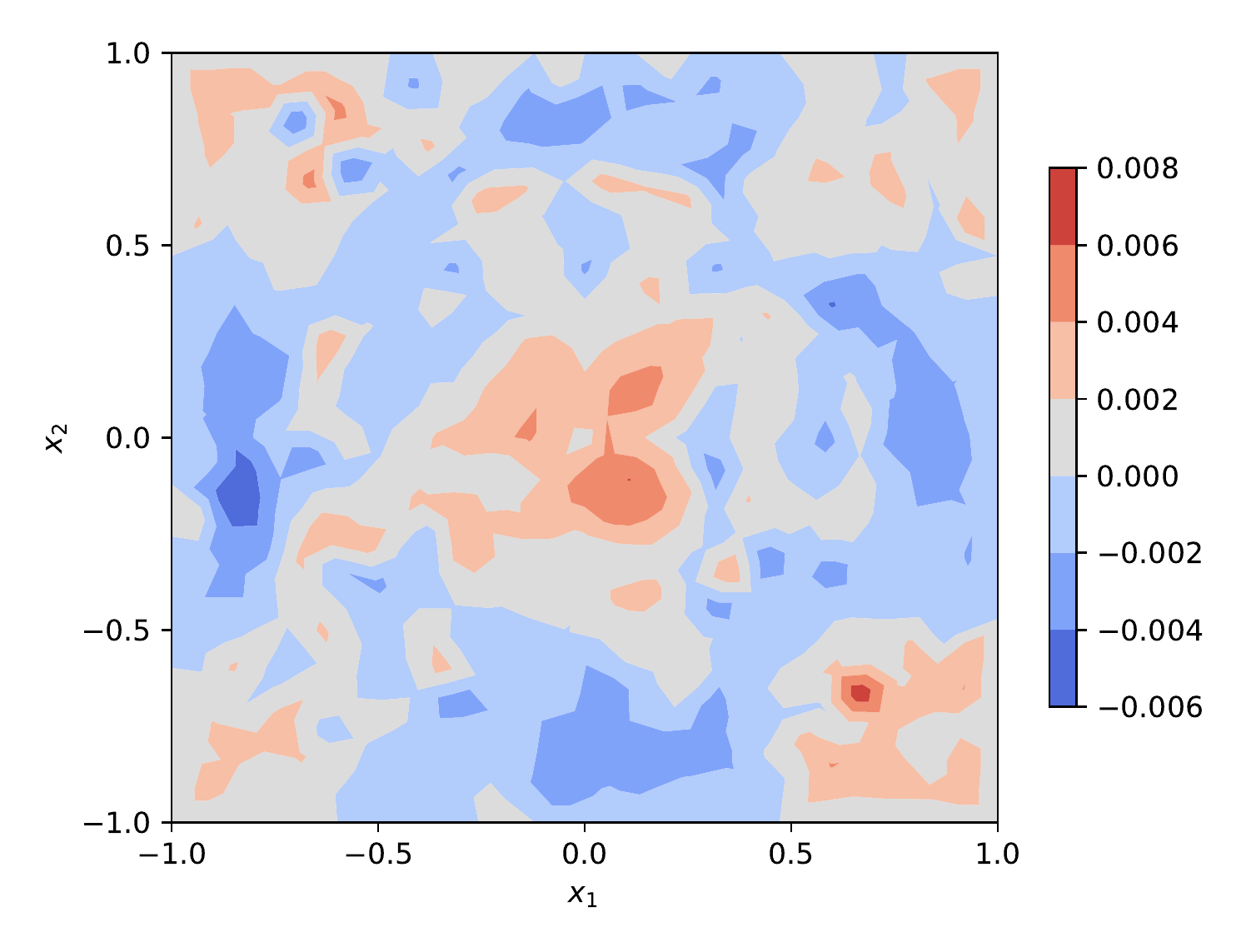}
 \end{minipage}
}}
  \caption{Numerical results for the solution of Poisson's equation in $\Omega_1$ under Cases I (top row) and II (bottom row) obtained by using the trained GF-Nets with $6\times 6$ subdomain blocks. Left: the exact solutions; middle: the predicted solutions; right: the numerical errors.}\vspace{-0.4cm}
  \label{fig:PoissonSquare}
\end{figure}

\paragraph{Effect of the domain partitioning strategy}
We test the impact of the domain partitioning strategy on GF-Nets by considering $4\times 4, 5\times 5$ and $6\times 6$ blocks. 
For the case of $4\times 4$ blocks, the predicted Green's function with the source point $\vxi=(-0.8,0.8)$ is shown in Figure \ref{fig:Poisson_GF} (left). Moreover, the time costs of the training process under different domain partition settings are reported in Appendix \ref{apdx:partitiontime}.
The trained GF-Net is then applied for solving the Poisson's equation. Three sets of quadrature points ($\#\mathcal V_q=145, 289, 545$) for numerical integration are considered and the resulted solution errors are reported in Table \ref{tab:Partitioning}.
It is easy to see that although different domain partitions are used, the numerical accuracy remains almost  at the same level, with only slight improvements for larger partitions and more quadrature points in both cases. The predicted results by GF-Nets with $6\times 6$ subdomain blocks are presented in Figure \ref{fig:PoissonSquare} for visual illustration.

\begin{table}[!ht]\small\renewcommand{\arraystretch}{}
  \centering
  \begin{tabular}{|c|ccc|ccc|}\hline
   \multirow{2}*{$\#\mathcal V_q$}  
   & \multicolumn{3}{c|}{Case I} & \multicolumn{3}{c|}{Case II}   \\\cline{2-7}
      & $4\times4$ & $5\times5$ & $6\times6$  & $4\times4$ & $5\times5$ & $6\times6$  \\\hline
    145 &  9.97e-3 & 9.63e-3& 8.67e-3 & 6.00e-3 & 6.17e-3 & 5.78e-3 \\
    289 &  9.42e-3 & 8.91e-3& 8.54e-3 & 4.46e-3 & 4.67e-3 & 5.32e-3 \\
    545 &  1.26e-2 & 1.19e-2& 1.18e-2 & 4.31e-3 & 4.79e-3 & 4.23e-3 \\
    \hline
  \end{tabular}
  \caption{Numerical  errors of the predicted solutions to the Poisson's equation in $\Omega_1$ obtained by using GF-Nets when three different domain partitions are used.}
  \label{tab:Partitioning}
\end{table}

\begin{figure}[!ht]
\centerline{\hspace{-0.6cm}
\includegraphics[width=1.9in]{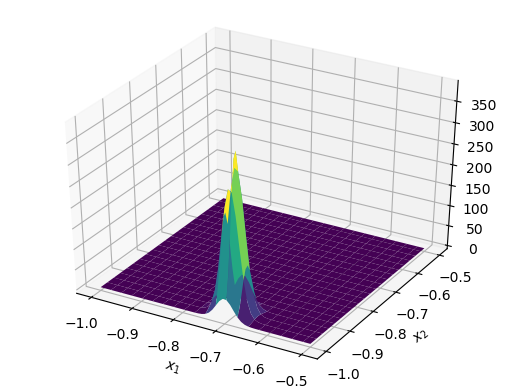}\hspace{-0.2cm}
\includegraphics[width=1.9in]{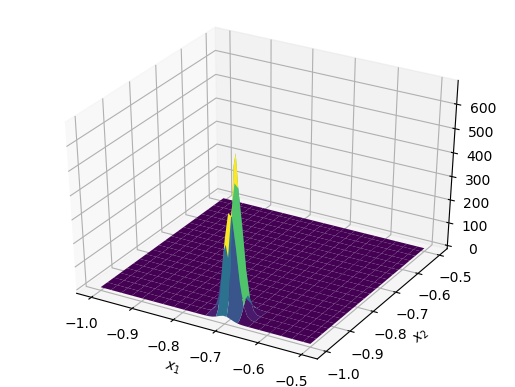}\hspace{-0.2cm}
\includegraphics[width=1.9in]{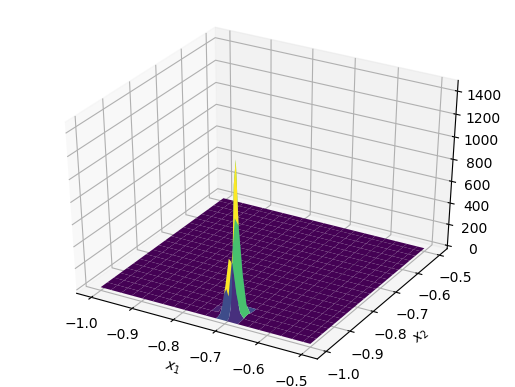}}
\caption{The approximations of the Dirac delta function with the source point at $(-0.7, -0.96)$ by the  Gaussian density functions with $s=0.02$ (left), $s=0.015$ (middle) and $s=0.01$ (right), respectively.}\vspace{-0.3cm}
\label{fig:GDFwithdifferents}
\end{figure}

\vspace{-0.3cm}
\paragraph{Effect of the Gaussian parameter \texorpdfstring{$s$}{s}} 
The value of the Gaussian parameter $s$ plays the most important role in accurately approximating the Dirac delta function. When the impulse source point $\vxi$ is positioned near the boundary, the Gaussian density function could not quickly decay to zero on the boundary if $s$ is not sufficiently small, which then causes large approximation errors due to the sudden truncation on the boundary (see the corresponding Gaussian density functions illustrated in Figure \ref{fig:GDFwithdifferents}). To find how such truncation error would affect the accuracy of GF-Nets and  corresponding fast solver for the Poisson's equation, we repeatedly fine-tuned the obtained GF-Net from $s=0.02$ to $s=0.015$ based on two experimental observations: 1) Directly training GF-Nets with $s=0.015$ or even smaller could be unstable because training samples are insufficient to represent a sharp distribution change around the impulse source; 2) Even by applying the fine-tuning strategy, the training time for a smaller $s$ is much higher. 
The resulting numerical errors of the predicted solutions to the Poisson’s equation are compared in Table \ref{tab:Poisson_sigma}, where 545 training samples for $\vxi$ and $6\times 6$ subdomain block are used. It is observed that a smaller $s$ leads to more accurate results when the integral quadrature is accurate enough, but of course at the cost of longer training times and larger memory usages. Considering that the choice of $s=0.02$ already yields good approximations, we will stick with it in the subsequent numerical tests.

\begin{table}[!ht]\small
  \centering
  \begin{tabular}{|c|cc|cc|}\hline
    \multirow{2}*{$\#\mathcal V_q$}  & \multicolumn{2}{c|}{Case I} & \multicolumn{2}{c|}{Case II} \\\cline{2-5}
     & $s=0.02$  & $s=0.015$ & $s=0.02$  & $s=0.015$   \\\hline
    145  & 8.67e-3  & 8.84e-3 & 5.78e-3 & 6.15e-3\\
    289  & 8.54e-3  & 6.80e-3 & 5.32e-3 & 5.27e-3\\
    545  & 1.18e-2  & 5.71e-3 & 4.23e-3 & 3.95e-3\\\hline
  \end{tabular}
  \caption{Numerical errors of the predicted solutions to the Poisson's equation on $\Omega_1$ obtained by using  GF-Nets with two different Gaussian parameter $s$.}
  \label{tab:Poisson_sigma}\vspace{-0.4cm}
\end{table}

\subsection{More tests on Poisson's equation}

\begin{figure}[!ht]
\centerline{\hspace{-0.6cm}
\begin{minipage}{0.3\linewidth}
\includegraphics[width=\textwidth]{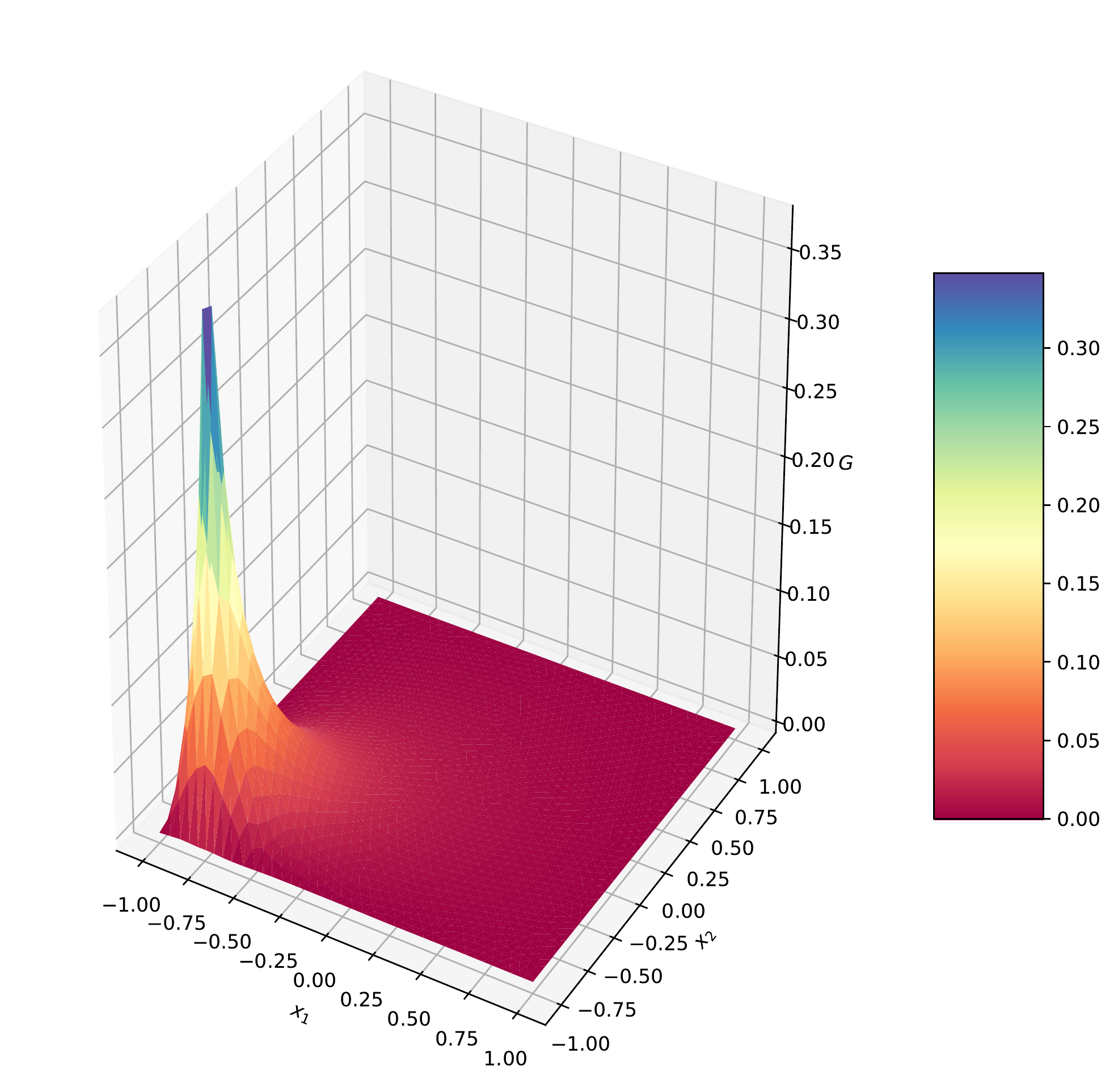}
\end{minipage}
\begin{minipage}{0.3\linewidth}
\includegraphics[width=\textwidth]{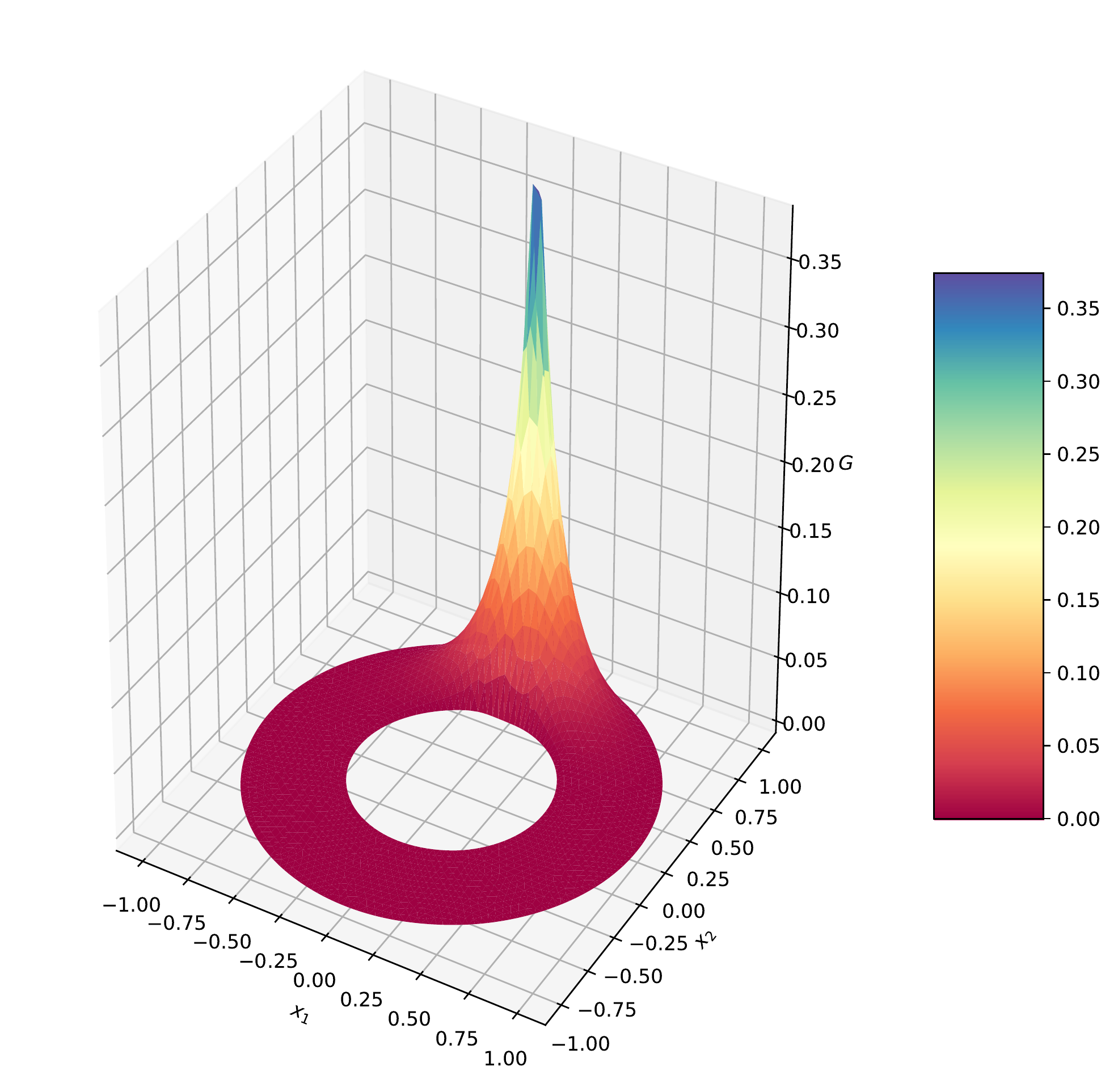}
\end{minipage}
\begin{minipage}{0.3\linewidth}
\includegraphics[width=\textwidth]{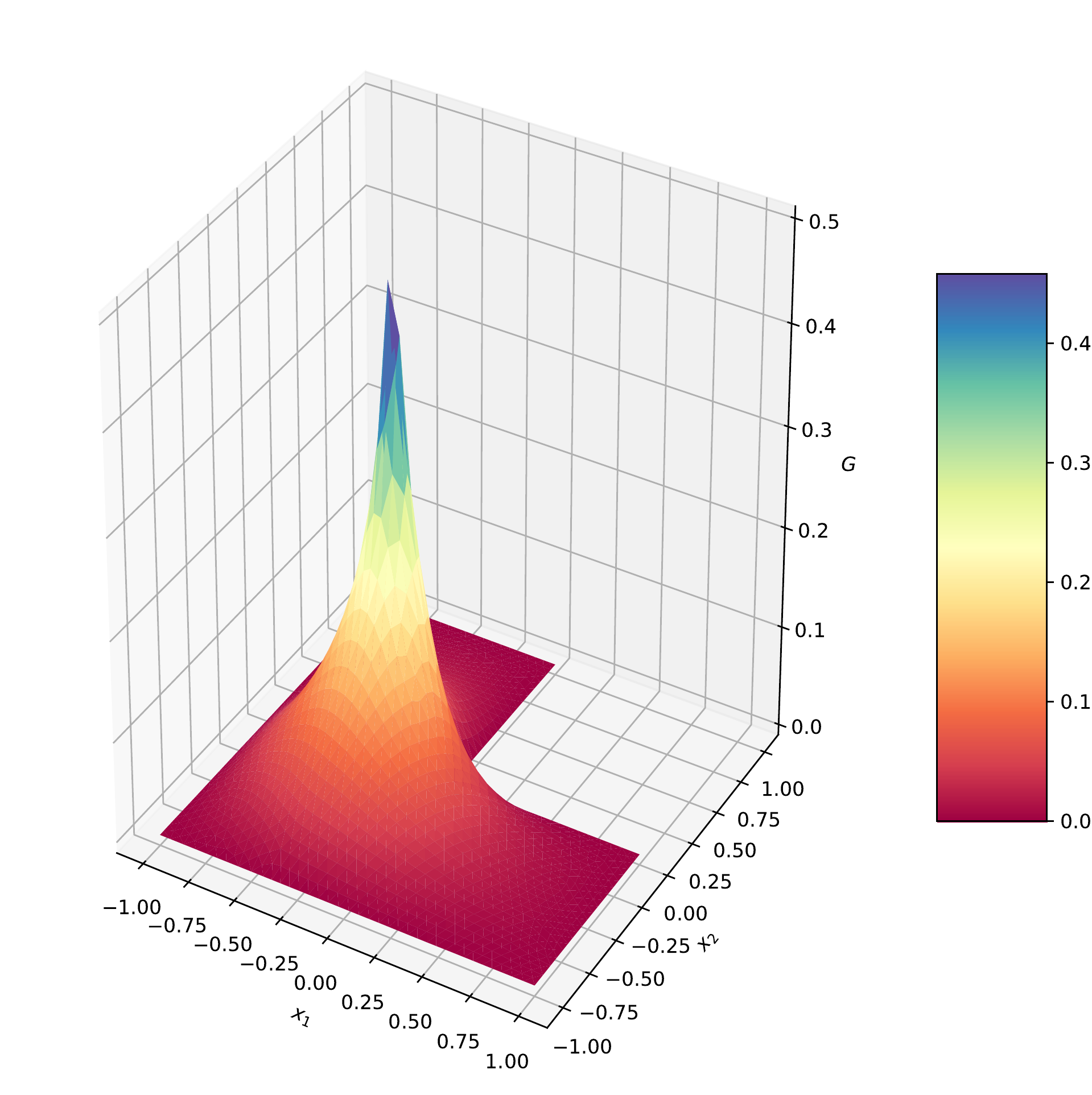}
\end{minipage}}
\caption{Predicted Green's functions $G(\vx, \vxi)$ of the Poisson's equation by GF-Nets. Left: $\Omega_1$ with $\vxi=(-0.8,-0.8)$; middle: $\Omega_2$ with $\vxi=(0,0.8)$; right: $\Omega_3$ with $\vxi=(-0.2,-0.2)$.
}\vspace{-0.4cm}
\label{fig:Poisson_GF}
\end{figure}

To further investigate the performance of the proposed GF-Net on non-convex domains,  we also test the proposed GF-Nets and corresponding fast  solver in the annulus $\Omega_2$ and the L-shaped domain $\Omega_3$. The same exact solutions as those (Cases I and II) in the previous subsection are considered. Some model parameters are listed in Table \ref{tab:dataset}. Examples of the predicted Green's functions $G(\vx, \vxi)$ in $\Omega_2$ with the source point $\vxi=(0, 0.8)$ and in $\Omega_3$ with  $\vxi=(-0.2,-0.2)$ are shown in Figure \ref{fig:Poisson_GF} (middle and right). As an example, numerical results for the solution of Poisson’s equation under Case II in $\Omega_2$ and $\Omega_3$ obtained using the trained GF-Nets are also plotted in Figure \ref{fig:Poisson_Washer_Lshape} for visual illustration. More related test results  are provided in Appendix \ref{apdx:ExtraResults}.

\begin{table}[!ht]\small\renewcommand{\arraystretch}{}
\centering
\begin{tabular}{|c|cccccccc|}\hline
Domain & $\#\mathcal V_\vxi$ & $\#\mathcal V_\vx^1$ & $\#\mathcal V_\vx^2$ & $\#\mathcal V_\vx^3$ & $m$ & $n$ & $c_1$ & $c_2$ \\\hline
$\Omega_1$ & 545 & 32753 & 8265 & 2105 & 4 & 4 & 5 & 10 \\
$\Omega_2$ & 493 & 27352 & 6981 & 1819 & 4 & 4 & 5 & 10 \\
$\Omega_3$ & 411 & 49663 & 6102 & 1565 & 6 & 6 & 5 & 10 \\\hline
\end{tabular}
\caption{Parameter settings for the GF-Nets and the corresponding fast solver.}
\label{tab:dataset}\vspace{-0.3cm}
\end{table}

\begin{figure}[!ht]
\centerline{\hspace{-0.3cm}
\subfigure[Exact solution]{
\begin{minipage}[t]{0.31\linewidth}
\centering
\includegraphics[width=\textwidth]{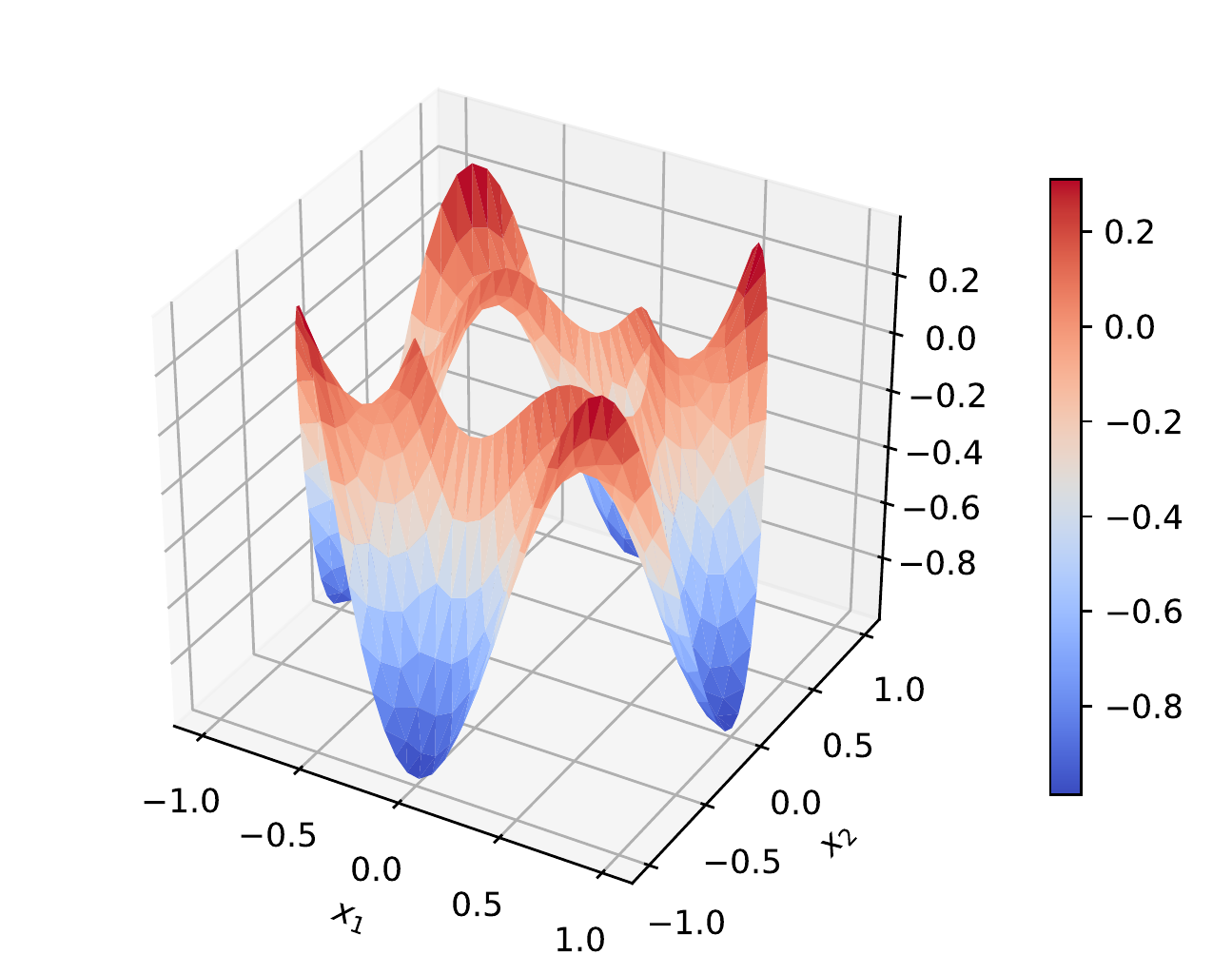}\\
 \includegraphics[width=\textwidth]{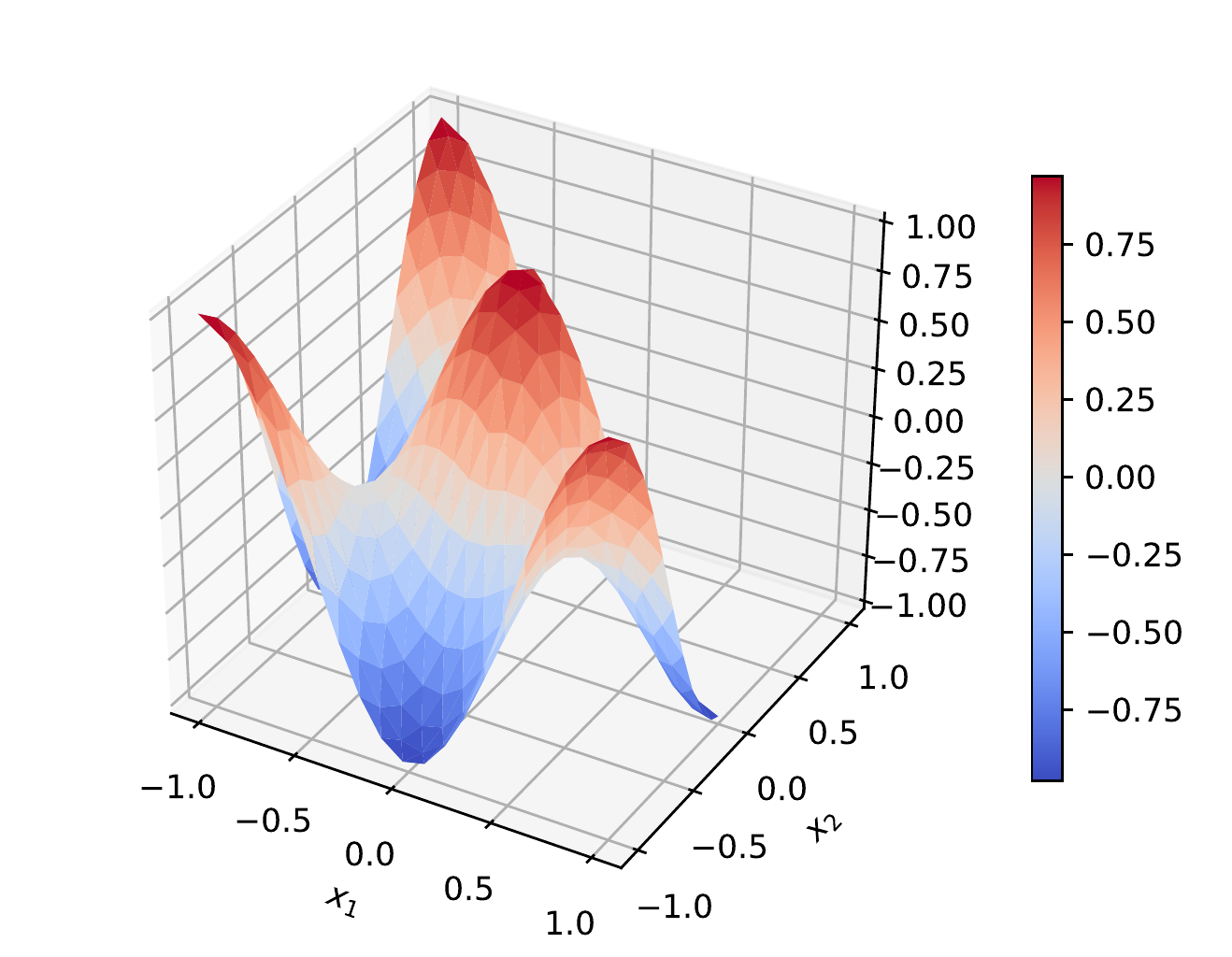}
 \end{minipage} 
}
\subfigure[Predicted solution]{
\begin{minipage}[t]{0.31\linewidth}
\centering 
\includegraphics[width=\textwidth]{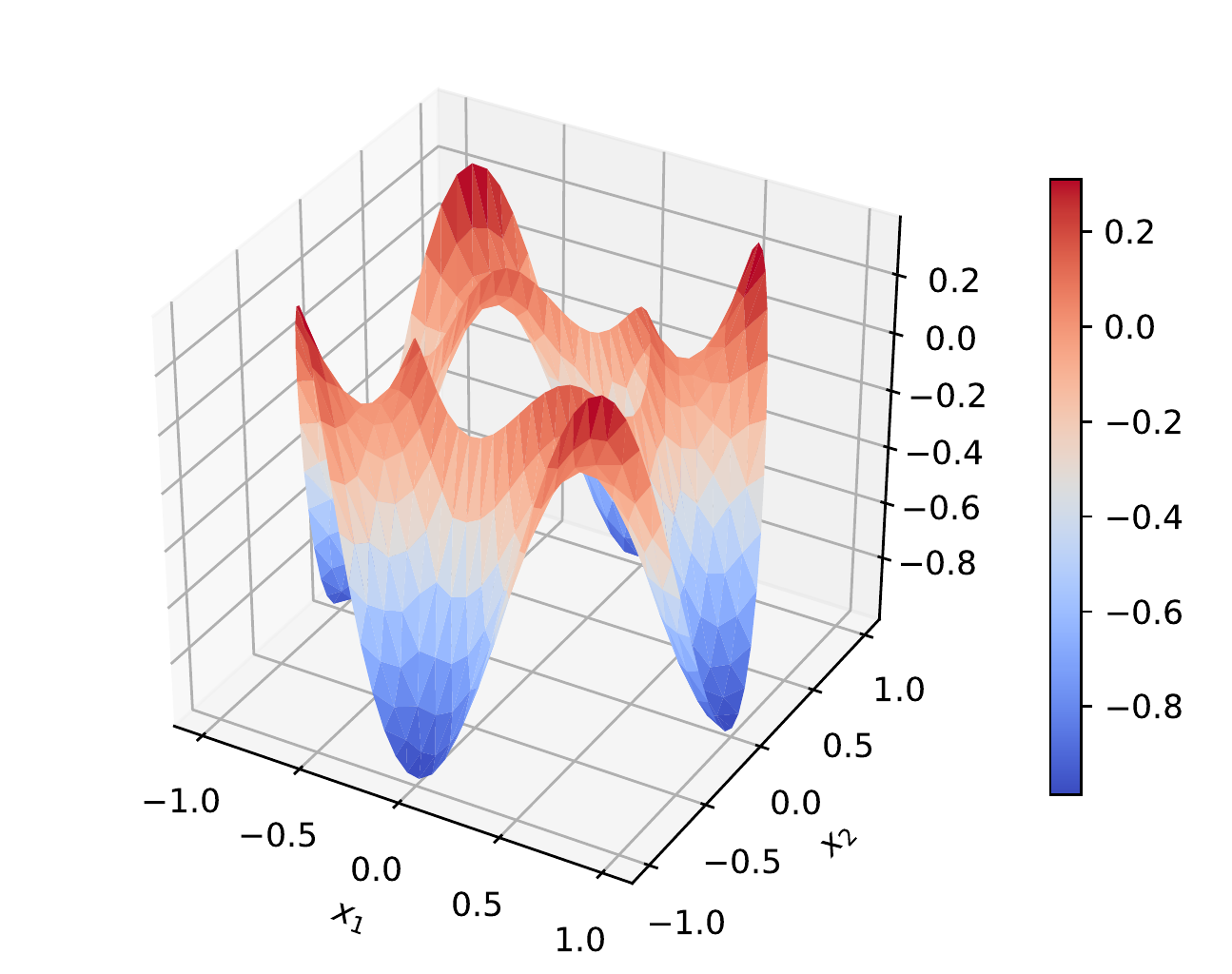}\\
 \includegraphics[width=\textwidth]{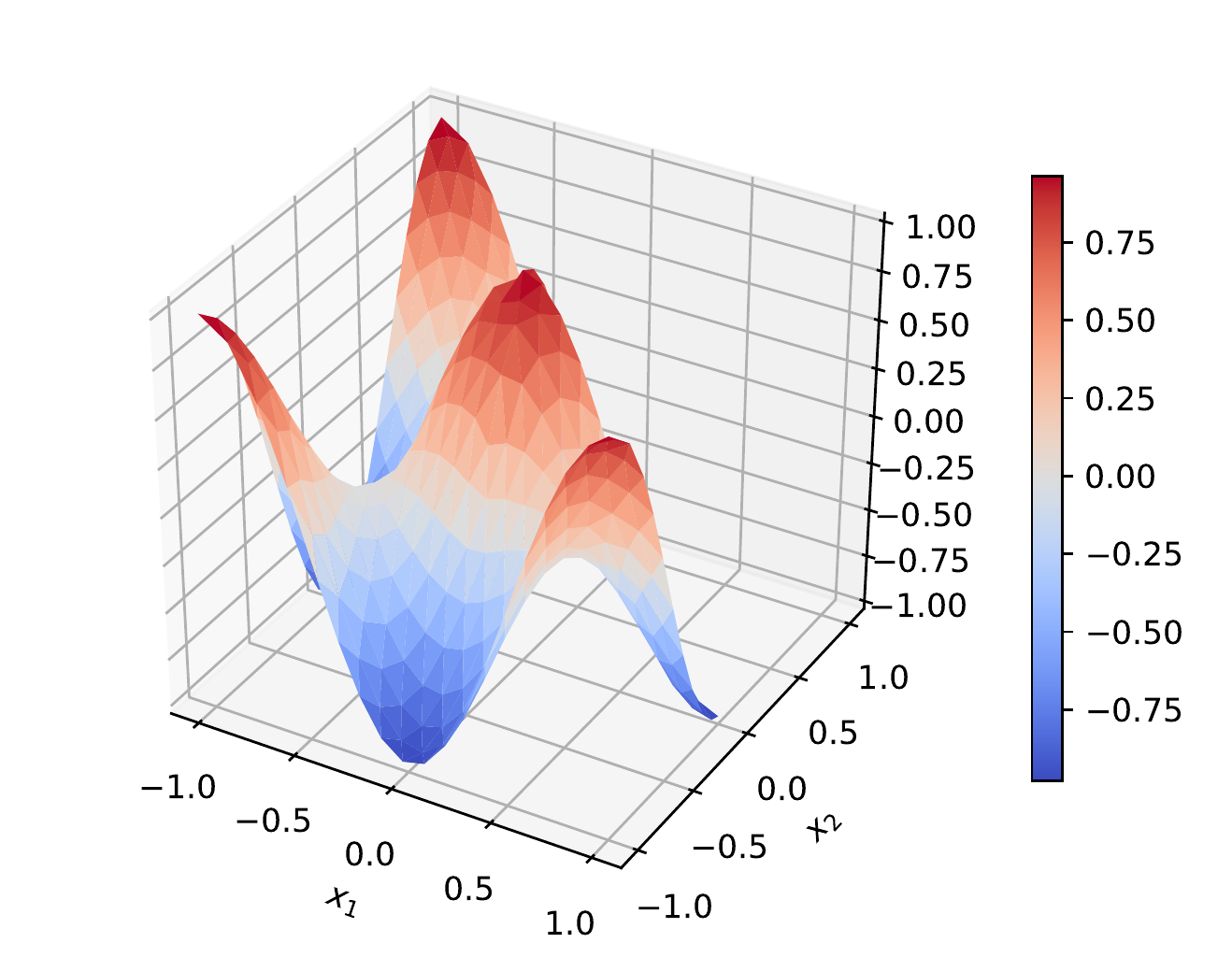}
  \end{minipage}
}
\subfigure[Error]{
\begin{minipage}[t]{0.3\linewidth}
\centering  
\includegraphics[width=\textwidth]{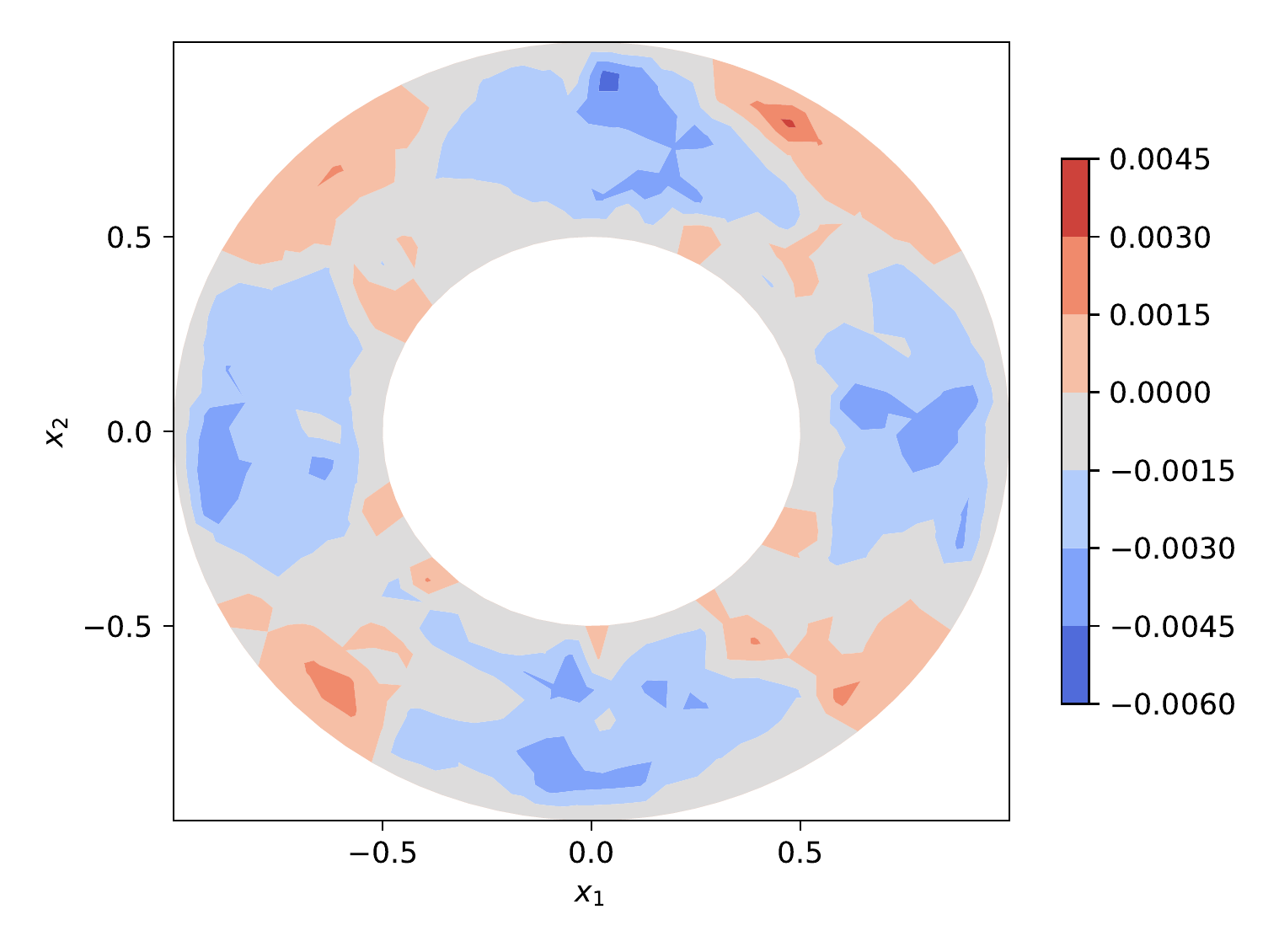}\\ 
 \includegraphics[width=\textwidth]{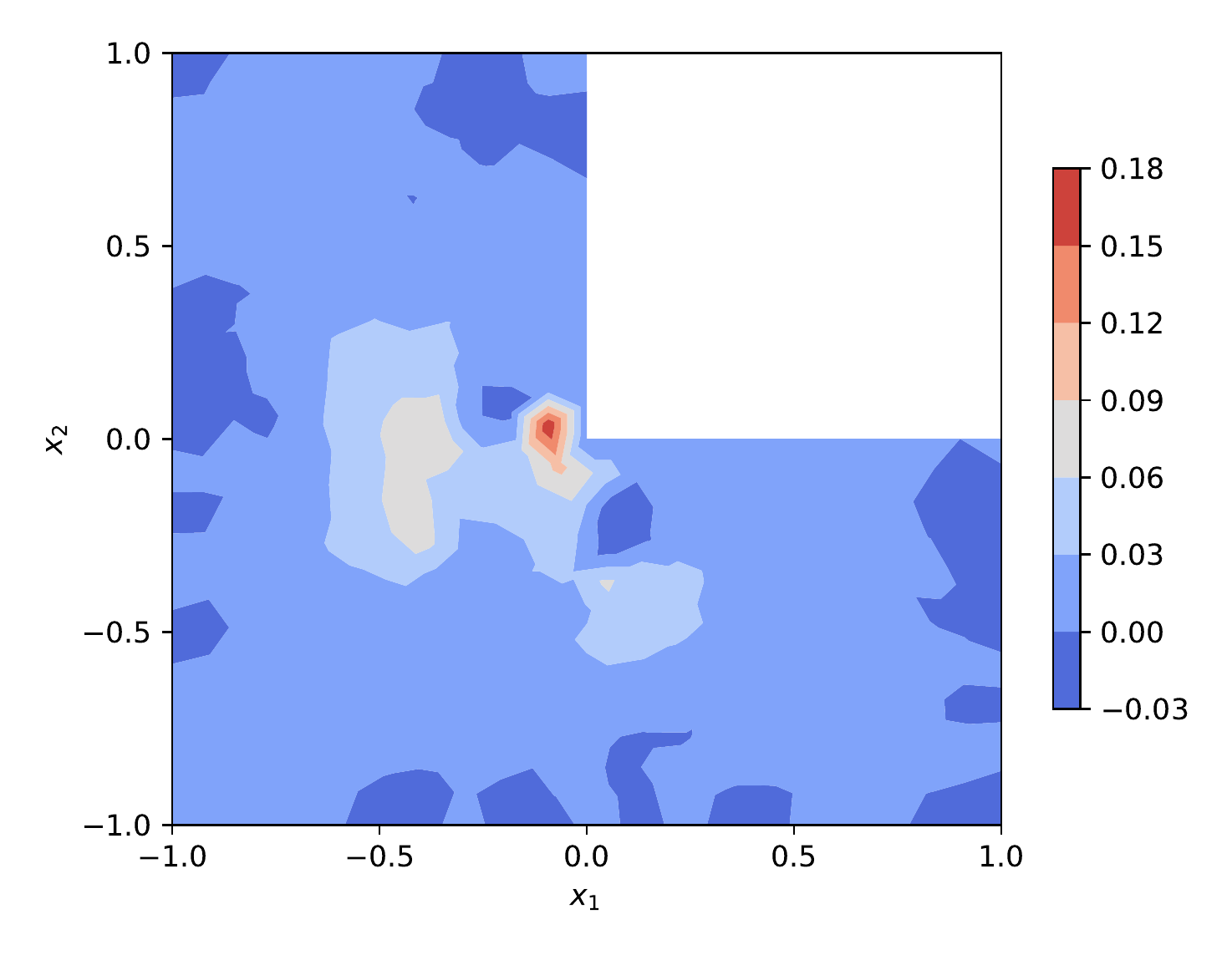}
  \end{minipage}
}}
  \caption{
  Numerical results for the solution of Poisson's equation (Case
II) in $\Omega_2$ (top row) and $\Omega_3$ (bottom row) obtained by using  the trained GF-Nets. Left: the exact solutions; middle: the predicted solutions; right: the numerical errors.}
  \label{fig:Poisson_Washer_Lshape}\vspace{-0.5cm}
\end{figure}

To demonstrate the accuracy and efficiency of the proposed method as a numerical solver of the target PDE, 
we also compare the numerical solutions of the Poisson's equation obtained by the trained GF-Nets with those of the classic finite element method (FEM) (implemented by FEniCS \citep{AlnaesEtal2015}) in the three domains $\Omega_1,\Omega_2$ and $\Omega_3$. For a fair comparison, the FEM solutions are computed on the same meshes as those used for evaluating \eqref{eq:rd-numsoln} with GF-Net. Numerical results for the Poisson's equation, including  solution errors and  computation times (in seconds per GPU card), are reported in Table \ref{tab:PoissonWasher}. We observe: 1) GF-Net is able to predict Green's functions as well as FEM on all the three domains; 2) 
Evaluating the formula \eqref{eq:rd-numsoln} on a finer quadrature mesh doesn't improve the accuracy significantly, which indicates the numerical error is dominated by Green's function approximation error in these cases; 3) the prediction accuracy of  GF-Net can be better than that of FEM, at least on the relatively coarse grid; 4) the time costs of GF-Net are comparable to that of FEM, and furthermore, due to the superior parallelism for multiple GF-Nets, the computation time by GF-Nets can be be significantly reduced when multiple GPU cards are available.

\vspace{0.3cm}
\begin{table}[!ht]\small
\renewcommand{\arraystretch}{}
  \centering
\begin{tabular}{|ccccccccc|}
\hline
\multicolumn{1}{|c|}{\multirow{2}{*}{$\Omega_1$: $\#\mathcal V_q$}} & \multicolumn{4}{c|}{Case I}                          & \multicolumn{4}{c|}{Case II}    \\ \cline{2-9} 
\multicolumn{1}{|c|}{}                                  & GF-Net  & Time & FEM     & \multicolumn{1}{c|}{Time} & GF-Net  & Time & FEM     & Time \\ \hline
\multicolumn{1}{|c|}{145}                               & 9.97e-3 & 0.11 & 2.36e-1 & \multicolumn{1}{c|}{0.15} & 6.00e-3 & 0.11 & 6.35e-2 & 0.16 \\
\multicolumn{1}{|c|}{289}                               & 9.42e-3 & 0.18 & 1.15e-1 & \multicolumn{1}{c|}{0.17} & 4.46e-3 & 0.18 & 2.67e-2 & 0.16 \\
\multicolumn{1}{|c|}{545}                               & 1.26e-2 & 0.40 & 5.19e-2 & \multicolumn{1}{c|}{0.24} & 4.31e-3 & 0.34 & 1.46e-2 & 0.23 \\ \hline
\multicolumn{1}{|c|}{\multirow{2}{*}{$\Omega_2$: $\#\mathcal V_q$}} & \multicolumn{4}{c|}{Case I}                          & \multicolumn{4}{c|}{Case II}    \\ \cline{2-9} 
\multicolumn{1}{|c|}{}                                  & GF-Net  & Time & FEM     & \multicolumn{1}{c|}{Time} & GF-Net  & Time & FEM     & Time \\ \hline
\multicolumn{1}{|c|}{143}                               & 1.27e-2 & 0.11 & 6.11e-1 & \multicolumn{1}{c|}{0.16} & 8.14e-3 & 0.11 & 4.33e-2 & 0.15 \\
\multicolumn{1}{|c|}{224}                               & 1.42e-2 & 0.14 & 5.51e-1 & \multicolumn{1}{c|}{0.17} & 8.27e-3 & 0.13 & 2.45e-2 & 0.16 \\
\multicolumn{1}{|c|}{493}                               & 1.36e-2 & 0.30 & 5.20e-1 & \multicolumn{1}{c|}{0.22} & 4.53e-3 & 0.28 & 9.95e-3 & 0.30 \\ \hline
\multicolumn{1}{|c|}{\multirow{2}{*}{$\Omega_3$: $\#\mathcal V_q$}} & \multicolumn{4}{c|}{Case I}                          & \multicolumn{4}{c|}{Case II}    \\ \cline{2-9} 
\multicolumn{1}{|c|}{}                                  & GF-Net  & Time & FEM     & \multicolumn{1}{c|}{Time} & GF-Net  & Time & FEM     & Time \\ \hline
\multicolumn{1}{|c|}{113}                               & 1.08e-2 & 0.17 & 2.48e-1 & \multicolumn{1}{c|}{0.14} & 5.09e-2 & 0.17 & 6.11e-2 & 0.15 \\
\multicolumn{1}{|c|}{225}                               & 8.79e-3 & 0.24 & 1.13e-1 & \multicolumn{1}{c|}{0.15} & 5.77e-2 & 0.23 & 2.65e-2 & 0.16 \\
\multicolumn{1}{|c|}{411}                               & 1.18e-2 & 0.38 & 5.37e-2 & \multicolumn{1}{c|}{0.22} & 4.89e-2 & 0.37 & 1.46e-2 & 0.23 \\ \hline
\end{tabular}
  \caption{Quantitative comparisons of solution errors and computation times (seconds) used by the  GF-Net and the FEM for solving the Poisson's equation.}
  \label{tab:PoissonWasher}\vspace{-0.2cm}
\end{table}

\subsection{Tests on the reaction-diffusion equation}

\begin{figure}[!ht]
\centerline{\hspace{-0.6cm}
\begin{minipage}{0.3\linewidth}
\includegraphics[width=\textwidth]{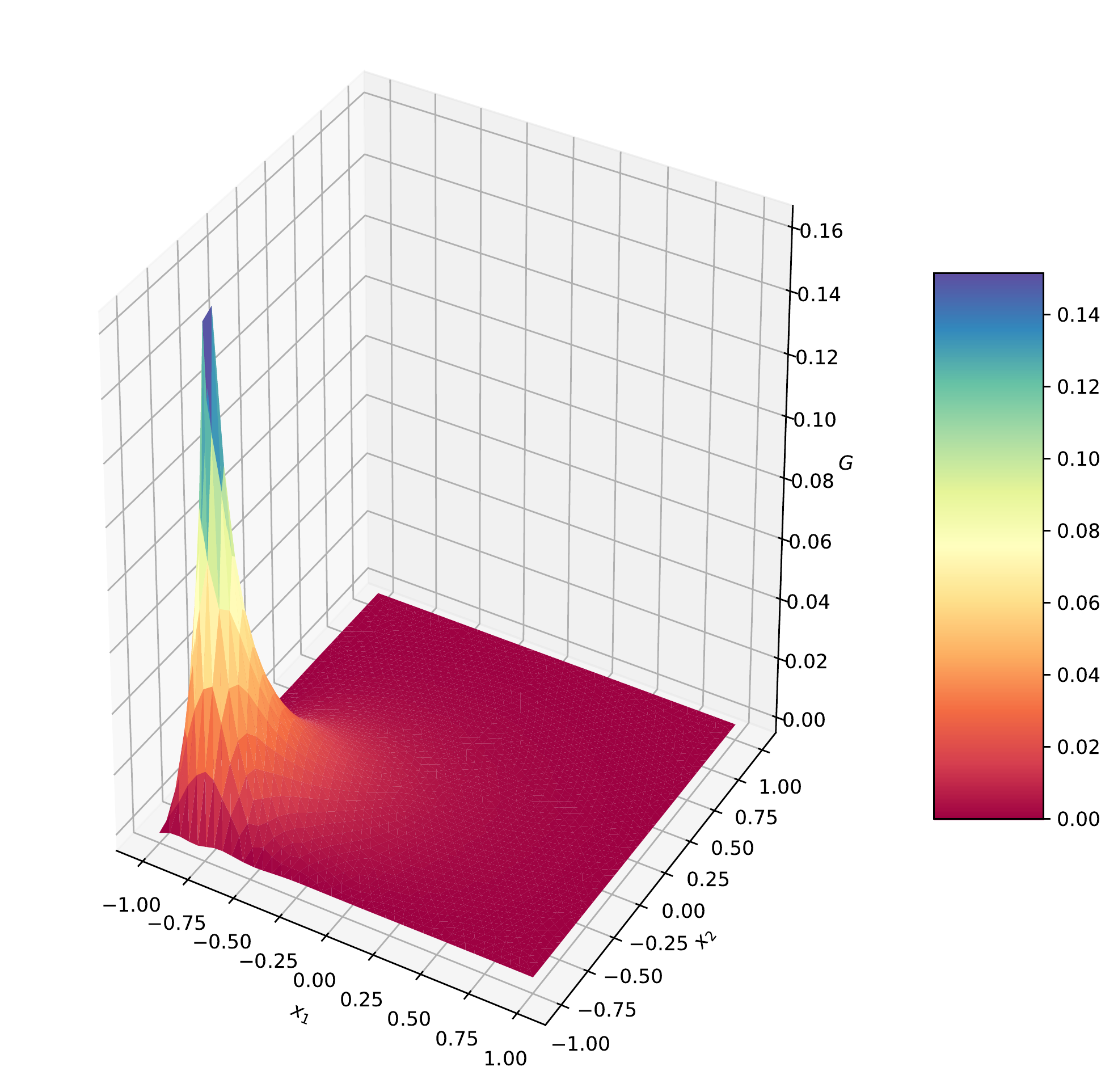}
\end{minipage}
\begin{minipage}{0.3\linewidth}
\includegraphics[width=\textwidth]{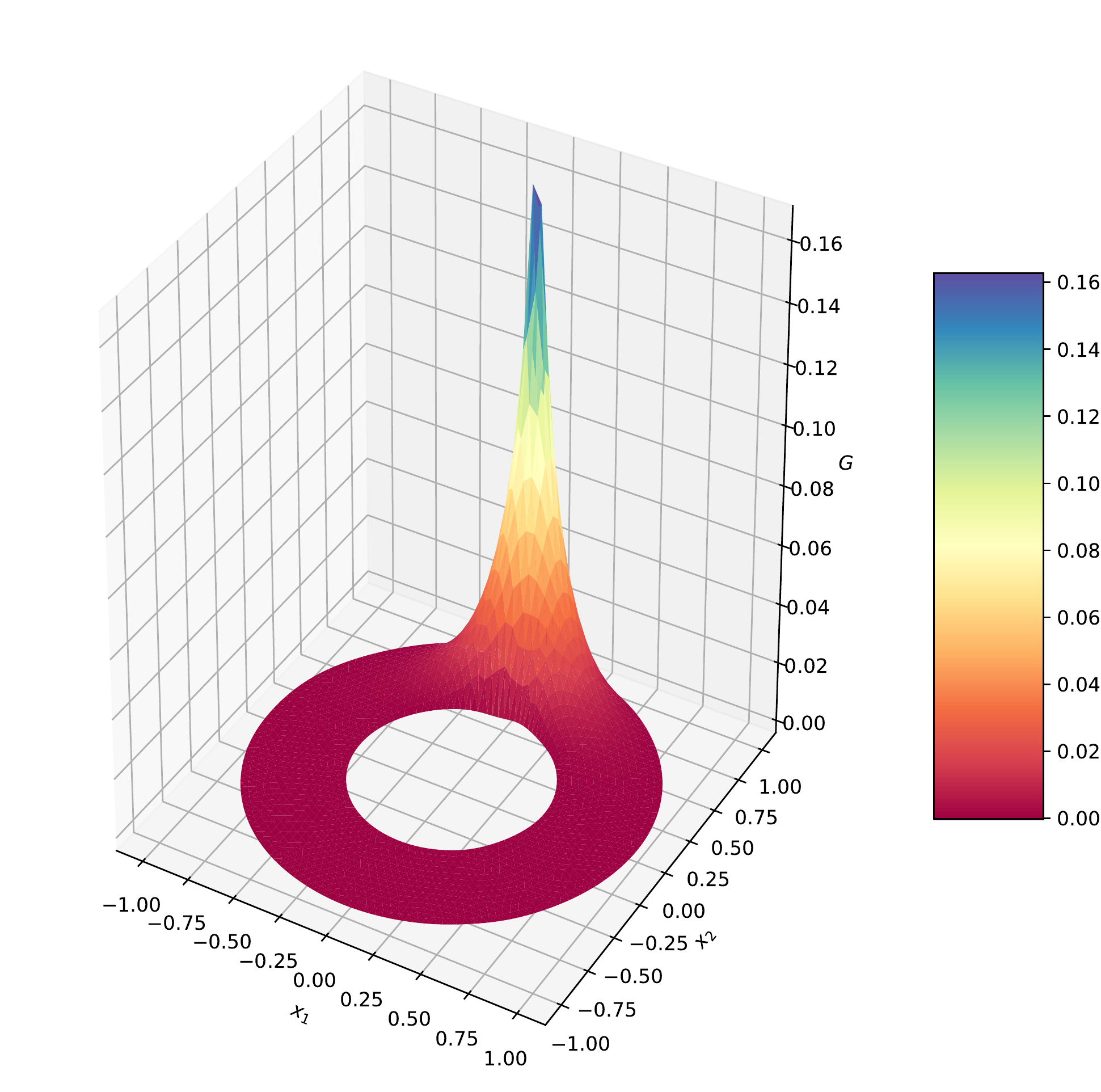}
\end{minipage}
\begin{minipage}{0.3\linewidth}
\includegraphics[width=\textwidth]{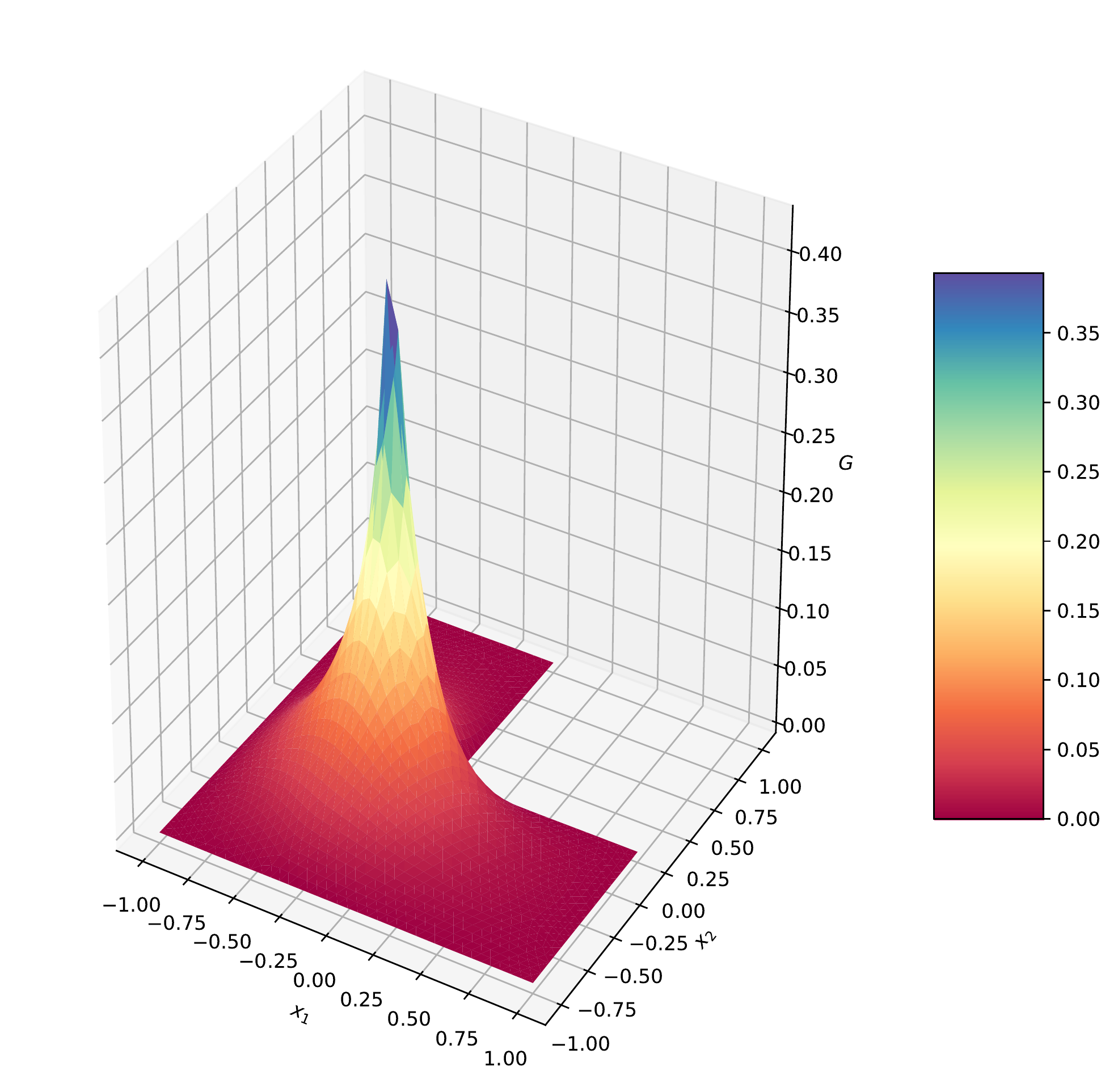}
\end{minipage}}
\caption{
Predicted Green's functions $G(\vx, \vxi)$ of the reaction-diffusion equation \eqref{readiff} by GF-Nets. Left: $\Omega_1$ with $\vxi=(-0.8,-0.8)$; middle: $\Omega_2$ with $\vxi=(0,0.8)$; right: $\Omega_3$ with $\vxi=(-0.2,-0.2)$.}
\label{fig:DiffusionReaction_GF}\vspace{-0.3cm}
\end{figure}

\begin{figure}[!t]
\centerline{\hspace{-0.3cm}
\subfigure[Exact solution]{\hspace{-0.5cm}
\begin{minipage}[t]{0.31\linewidth}
\centering 
\includegraphics[width=\textwidth]{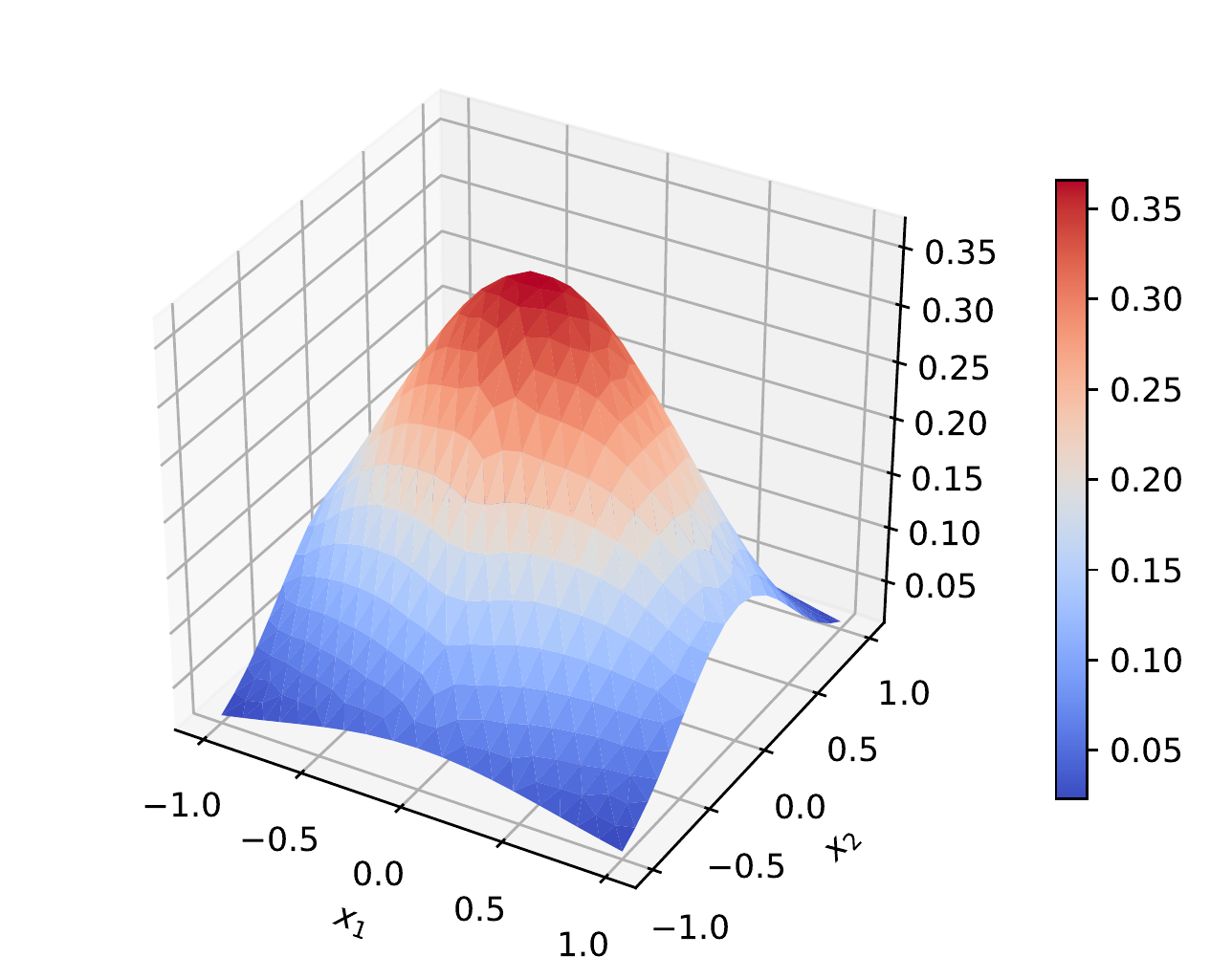}\\
 \includegraphics[width=\textwidth]{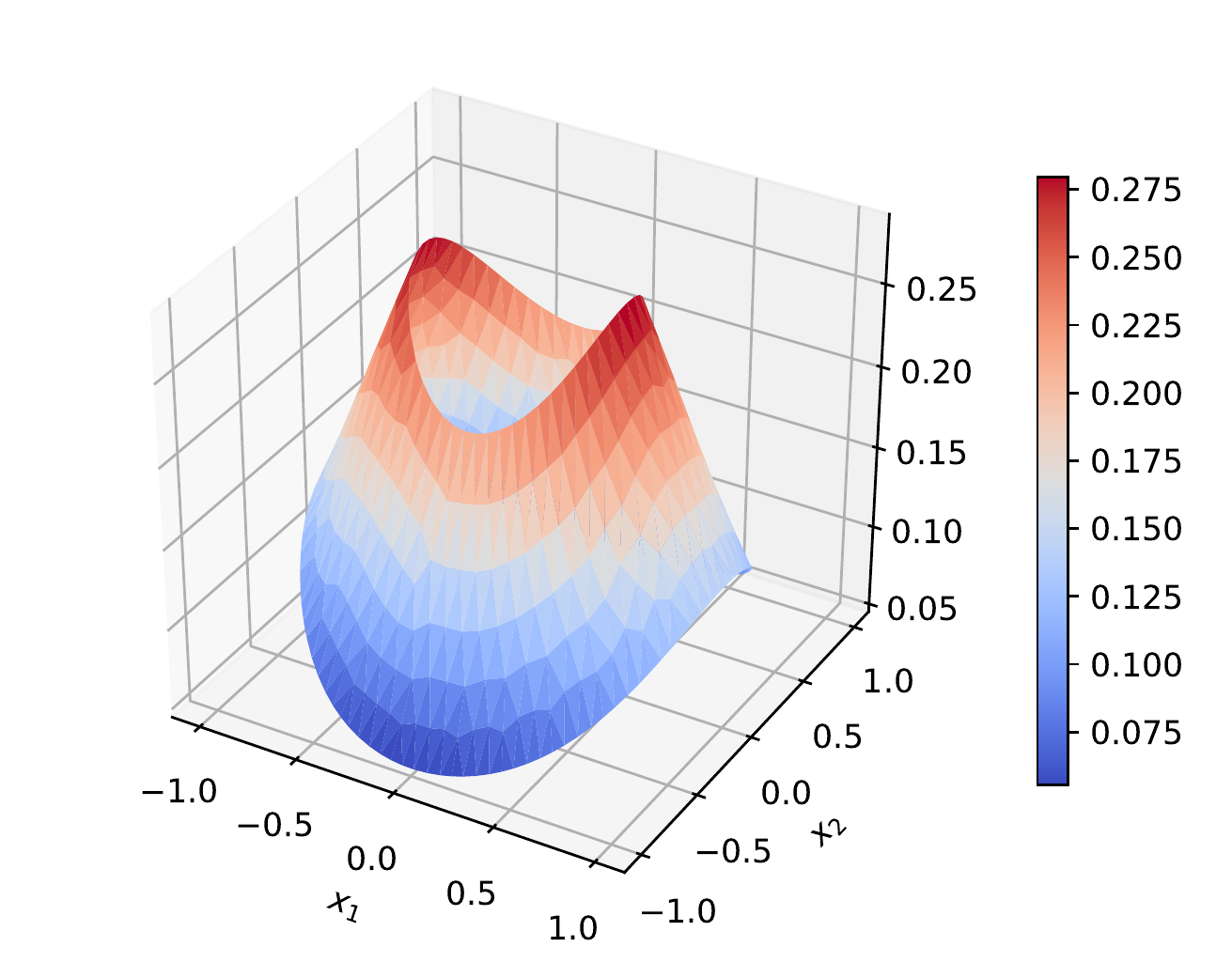}\\
 \includegraphics[width=\textwidth]{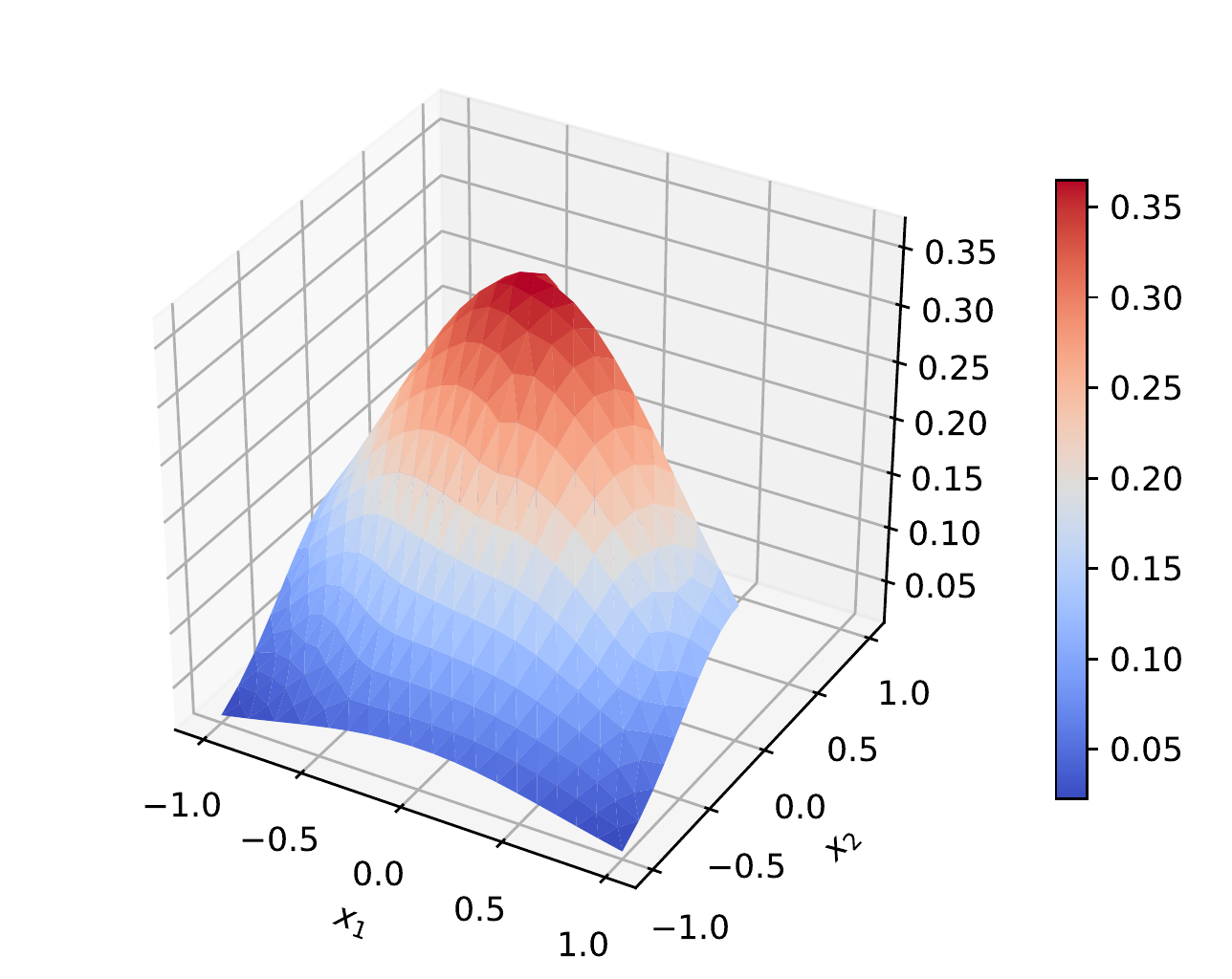}
 \end{minipage}
}
\subfigure[Predicted solution]{
\begin{minipage}[t]{0.31\linewidth}
\centering 
\includegraphics[width=\textwidth]{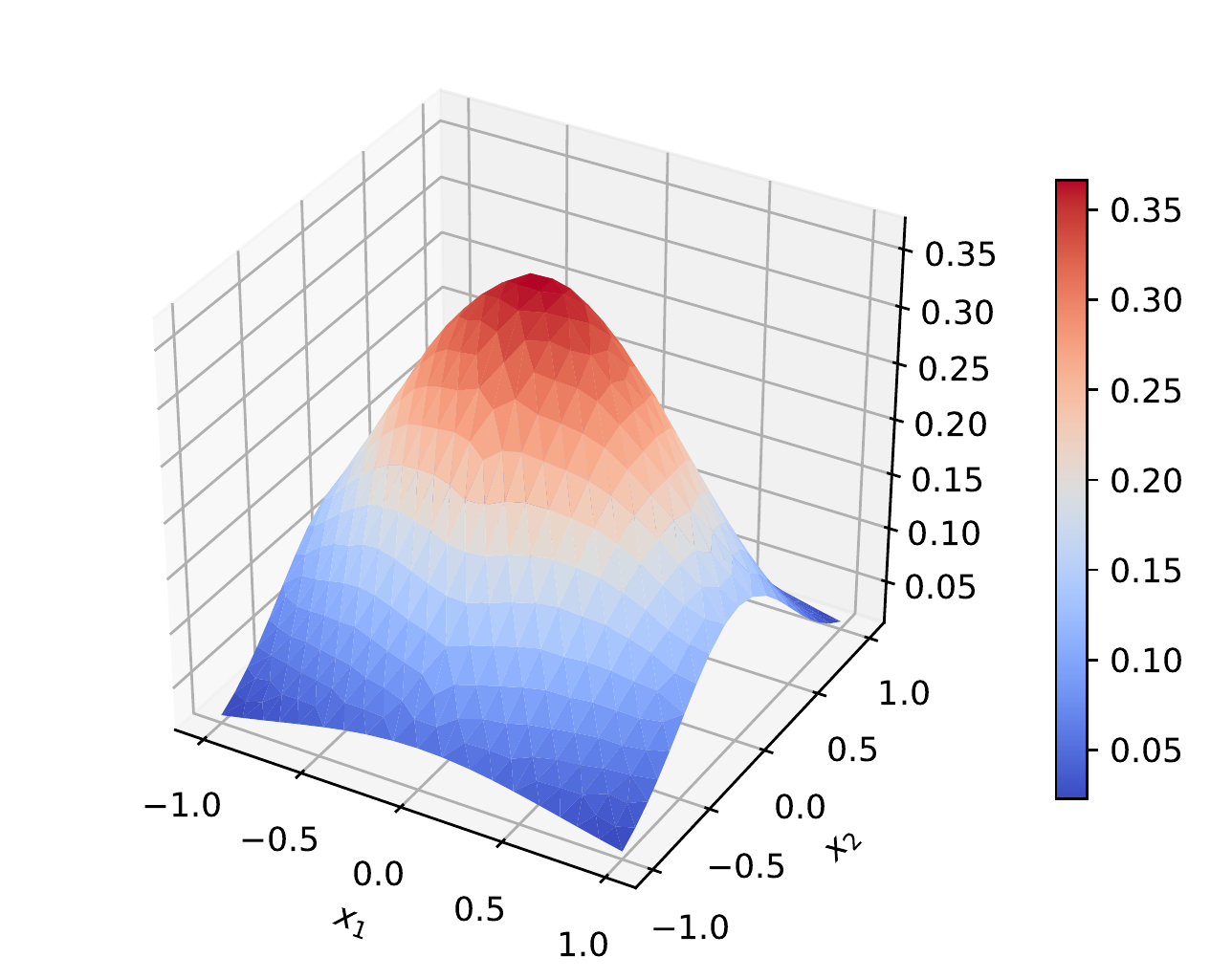}\\
 \includegraphics[width=\textwidth]{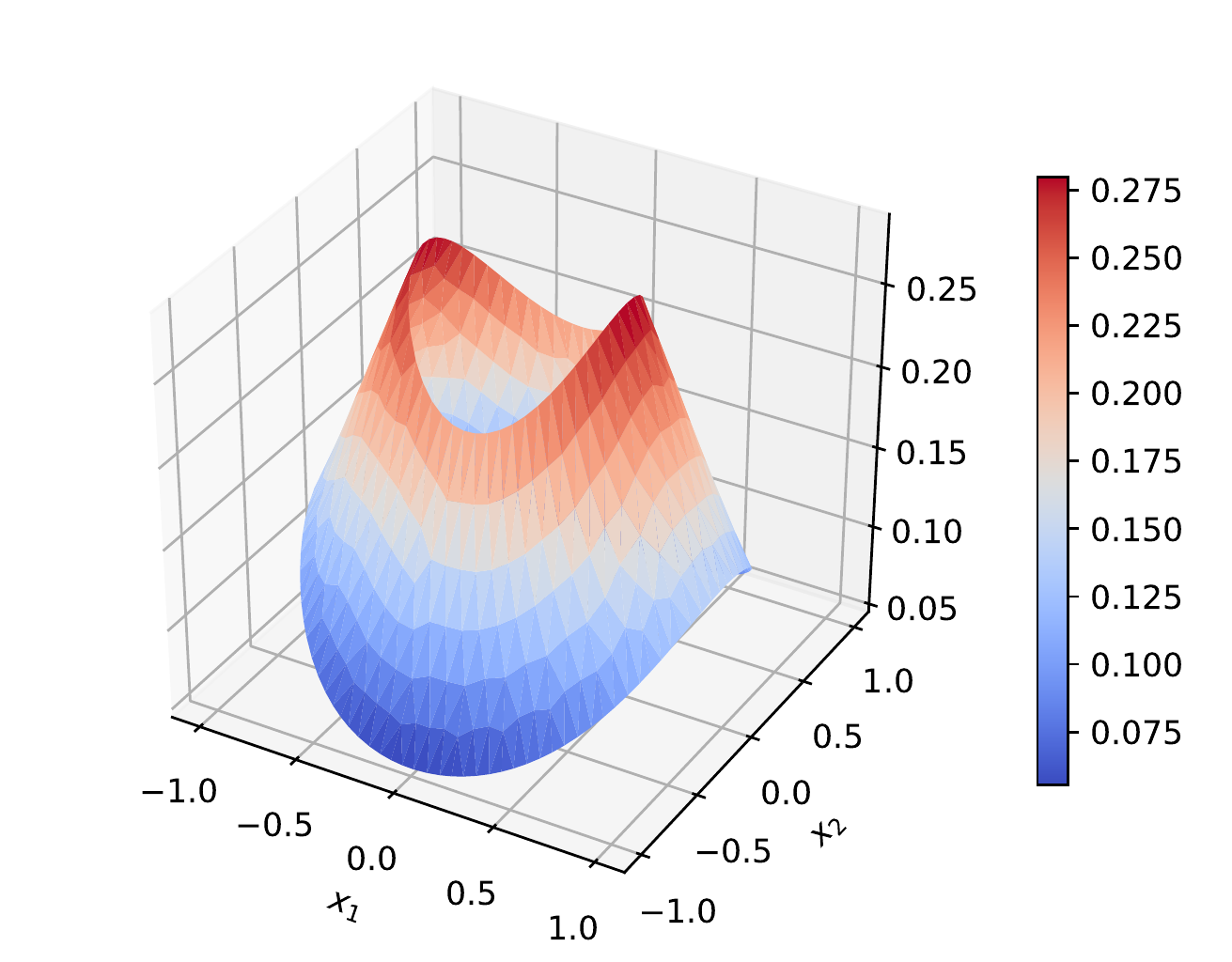}\\
 \includegraphics[width=\textwidth]{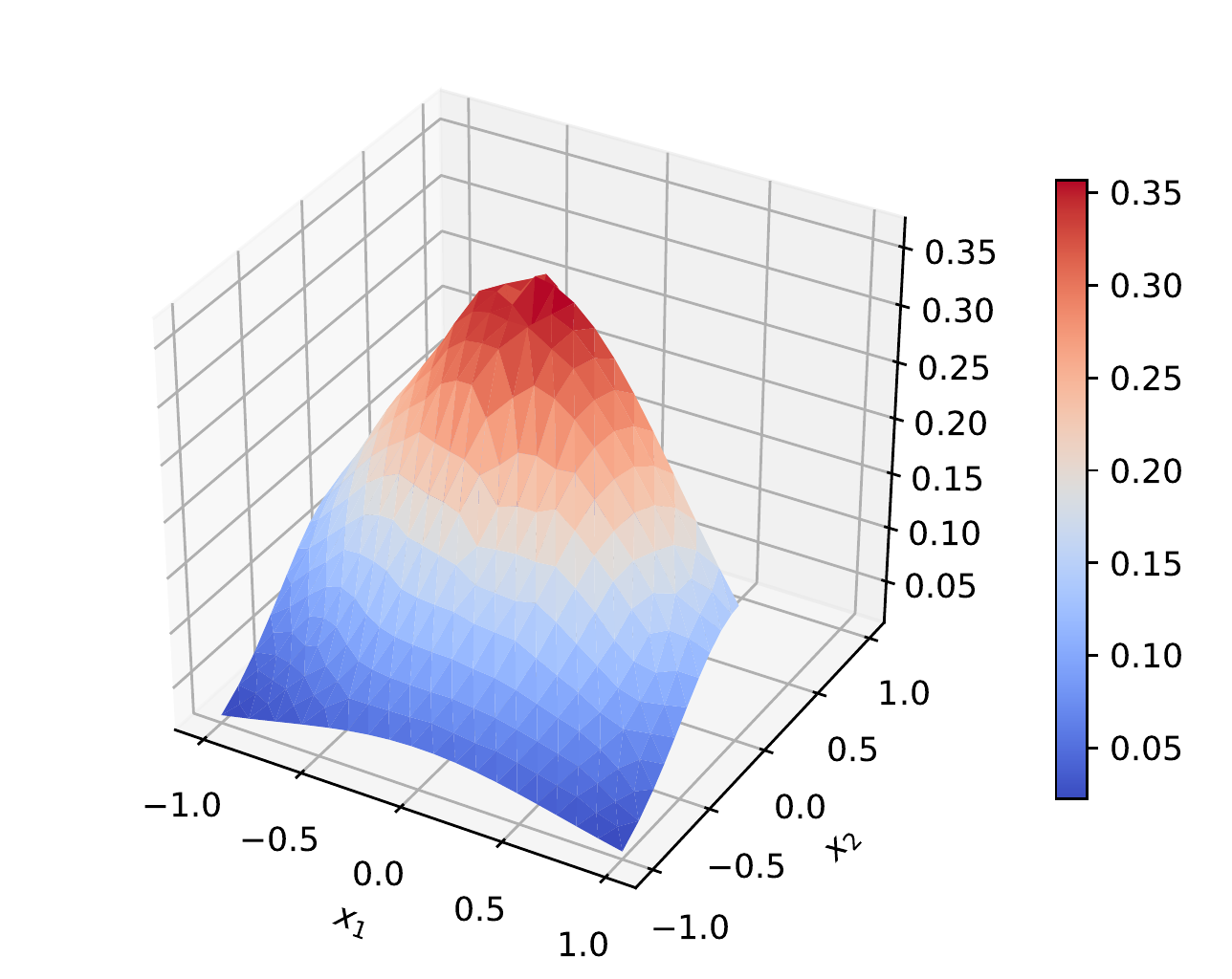}
  \end{minipage}
}
\subfigure[Error]{
\begin{minipage}[t]{0.31\linewidth}
\centering 
\vspace{.5cm}
\includegraphics[width=\textwidth]{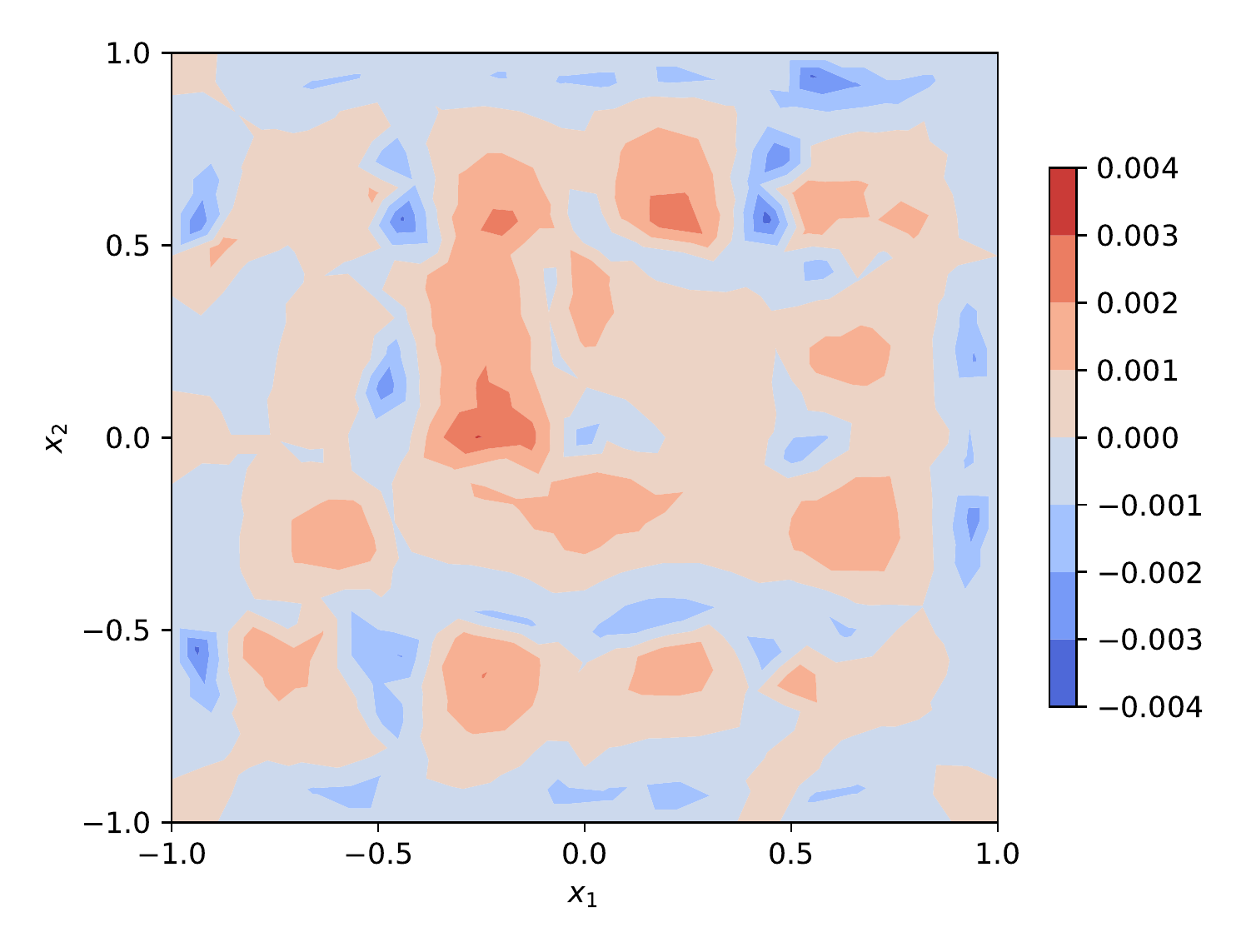}\\ 
 \includegraphics[width=\textwidth]{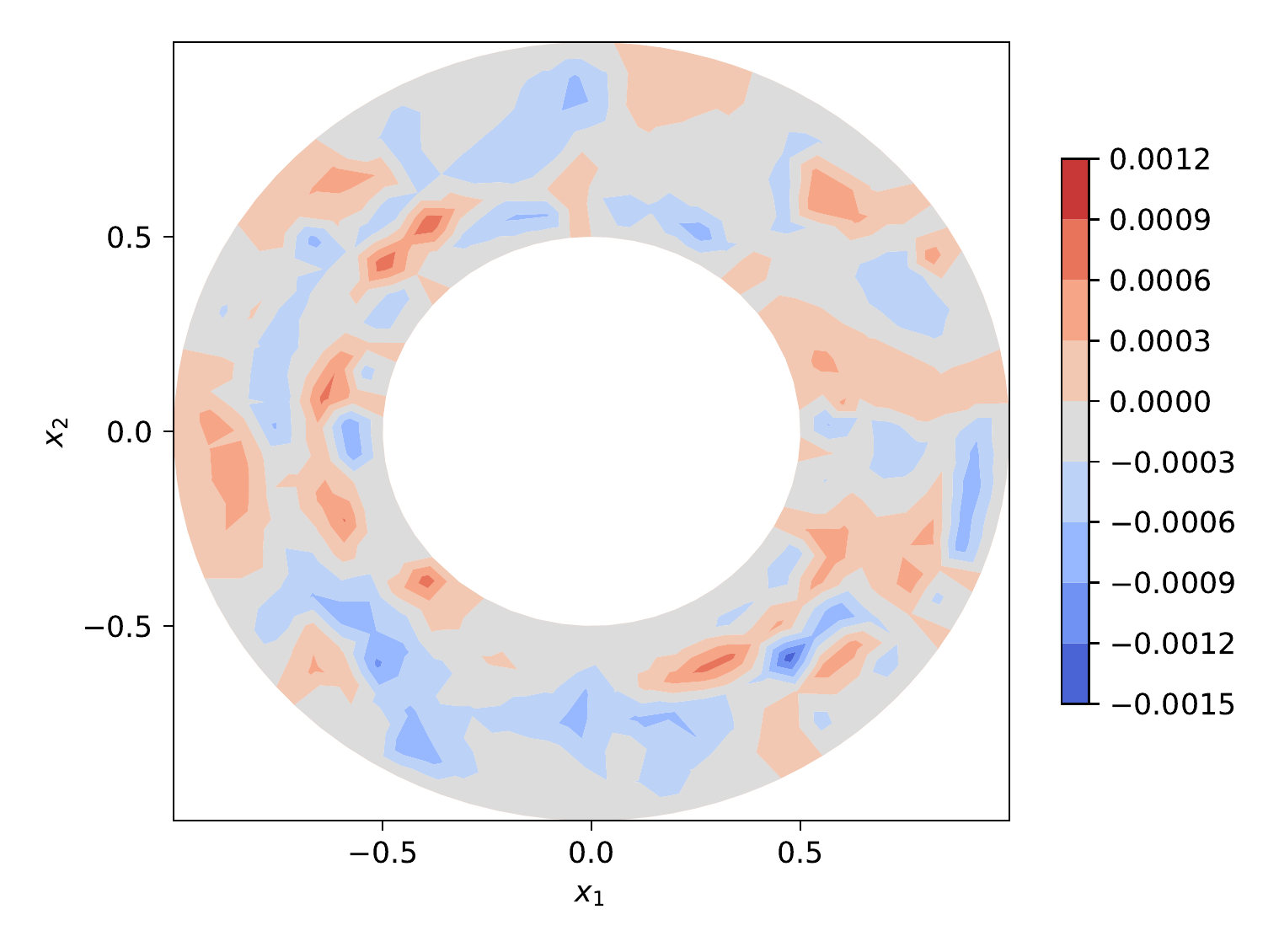}\\ 
 \includegraphics[width=\textwidth]{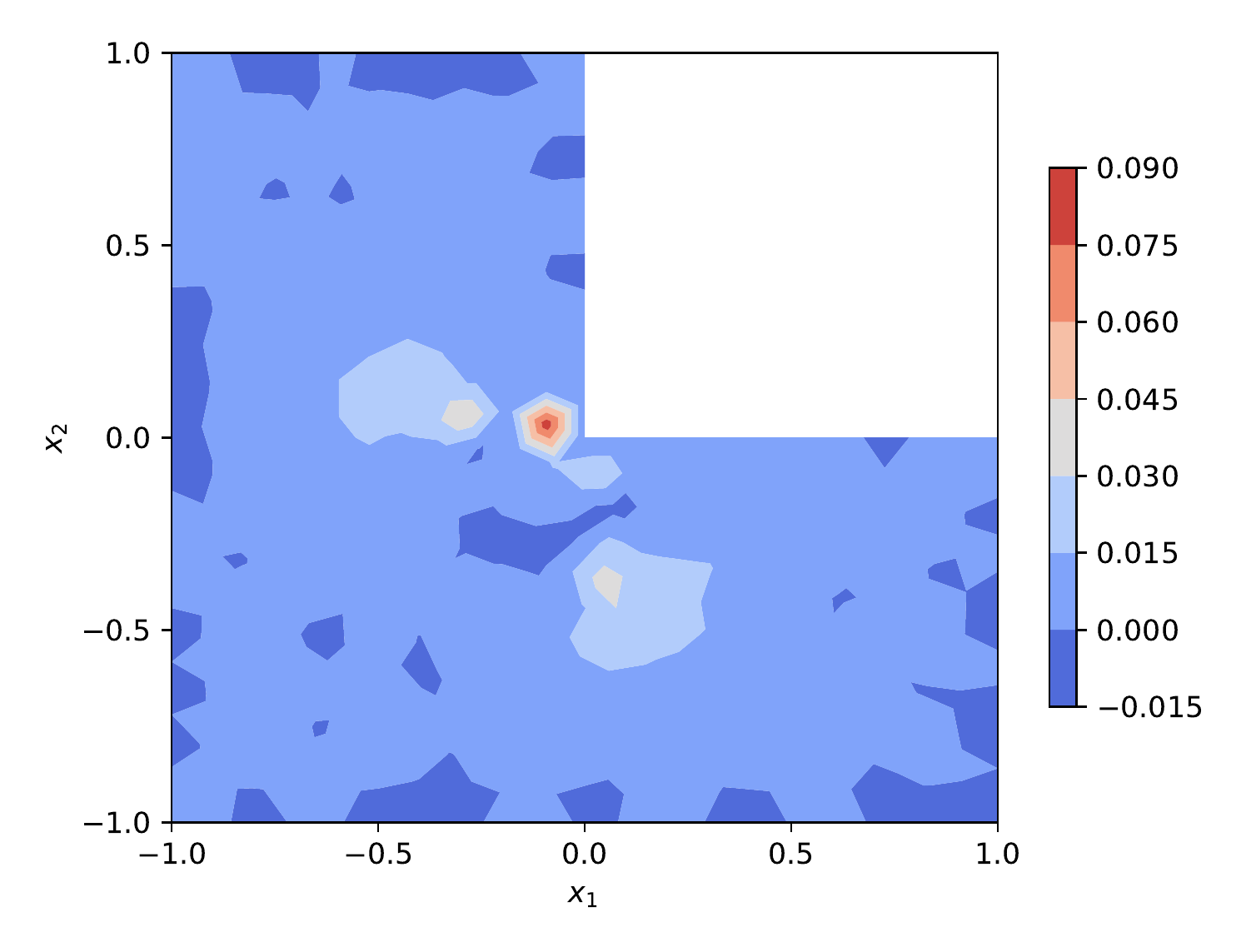}
  \end{minipage}
}}
  \caption{
  Numerical results for the solution of the reaction-diffusion equation \eqref{readiff} in $\Omega_1$ (top row), $\Omega_2$ (middle row) and $\Omega_3$ (bottom row) obtained by using the trained GF-Nets. Left: the exact solutions; middle: the predicted solutions; right: the numerical errors.}
  \label{fig:DiffusionReaction}
 \vspace{-0.4cm}
\end{figure}

We next test the following reaction-diffusion  operator: 
\begin{equation}\label{readiff}
\cL(u) = -\nabla\cdot((1+2x_2^2)\nabla u)+ (1+x_1^2) u 
\end{equation}
defined in the same three typical domains as before. The exact solution is chosen as $u(x_1, x_2)=e^{-(x_1^2+2x_2^2+1)}$ and the boundary conditions and the source term are determined accordingly. 
We use the same parameters as listed in Table \ref{tab:dataset}. 
The predicted Green's functions $G(\vx, \vxi)$ are shown in Figure \ref{fig:DiffusionReaction_GF} for the problem in $\Omega_1$ with the source point $\vxi=(-0.8, -0.8)$, $\Omega_2$ with $\vxi=(0, 0.8)$, and $\Omega_3$ with $\vxi=(-0.2,-0.2)$. Numerical results for the predicted solutions to  the reaction-diffusion equation \eqref{readiff} are reported in Table \ref{tab:DiffusionReaction} and plotted in Figure \ref{fig:DiffusionReaction}. It is observed that the proposed GF-Net method again achieves similar numerical performance as to the Poisson's equation. 

\begin{table}[!ht]\small
\renewcommand{\arraystretch}{}
  \centering
  \begin{tabular}{|cc|cc|cc|}\hline
    \multicolumn{2}{|c|}{$\Omega_1$} & \multicolumn{2}{|c|}{$\Omega_2$} & \multicolumn{2}{|c|}{$\Omega_3$} \\\hline
    $\#\mathcal V_q$ & $Error$ & $\#\mathcal V_q$ & $Error$  & $\#\mathcal V_q$ & $Error$ \\\hline
     145 & 4.82e-3 & 143 & 7.89e-3 & 113 & 4.34e-2\\ 
     289 & 4.52e-3 & 224 & 5.08e-3 & 225 & 5.10e-2\\
     545 & 5.02e-3 & 493 & 2.23e-3 & 411 & 4.50e-2\\
    \hline
  \end{tabular}
  \caption{Numerical errors of the predicted solutions to the reaction-diffusion equation \eqref{readiff} in $\Omega_1, \Omega_2$ and $\Omega_3$, obtained by using the trained GF-Nets.}
  \label{tab:DiffusionReaction}\vspace{-0.5cm}
\end{table}

\section{Conclusion} 
In this paper, we proposed  the neural network model ``GF-Net" to learn the Green's functions of the classic linear reaction-diffusion equations in the unsupervised fashion. Our method 
overcomes the challenges faced by classic and machine learning approaches in determining the Green's functions to differential operators in arbitrary domains. 
A series of procedures were taken to embed underlying properties of the Green's functions into the GF-Net model. In particular, the symmetry 
feature is preserved by adding a penalization term to the loss function, and a domain decomposition approach is used for accelerating training and achieving better accuracy. 
The GF-Nets then can  be used for fast numerical solutions of the target PDE subject to various sources and Dirichlet boundary conditions without the need of network retraining.
Numerical experiments were also performed that show our GF-Nets can well handle the reaction-diffusion equations in arbitrary domains. Some interesting future works include the use of hard constraints for better match of the boundary values, the improvement of the sampling strategies for training GF-Nets for higher dimensional problems, and the extension of the proposed method to time-dependent  and  nonlinear PDEs.

\acks{X. Zhang’s work is partially supported by National Key
Research and Development Program of China (2021YFD1900805-02). 
Z. Wang's work is partially supported by U.S. National Science Foundation grant DMS-2012469. L. Ju's work is partially supported by U.S. Department of Energy grant number DE-SC0022254. }

\bibliography{Ref_DL.bib}

\newpage
\appendix

\section{More Ablation Studies}

\subsection{Based on the Green’s function with a fixed point source}\label{apdx:UnitDisk}
\begin{figure}[!ht]
  \centering
  \begin{minipage}{.47\linewidth}
  \includegraphics[width=1.0\textwidth]{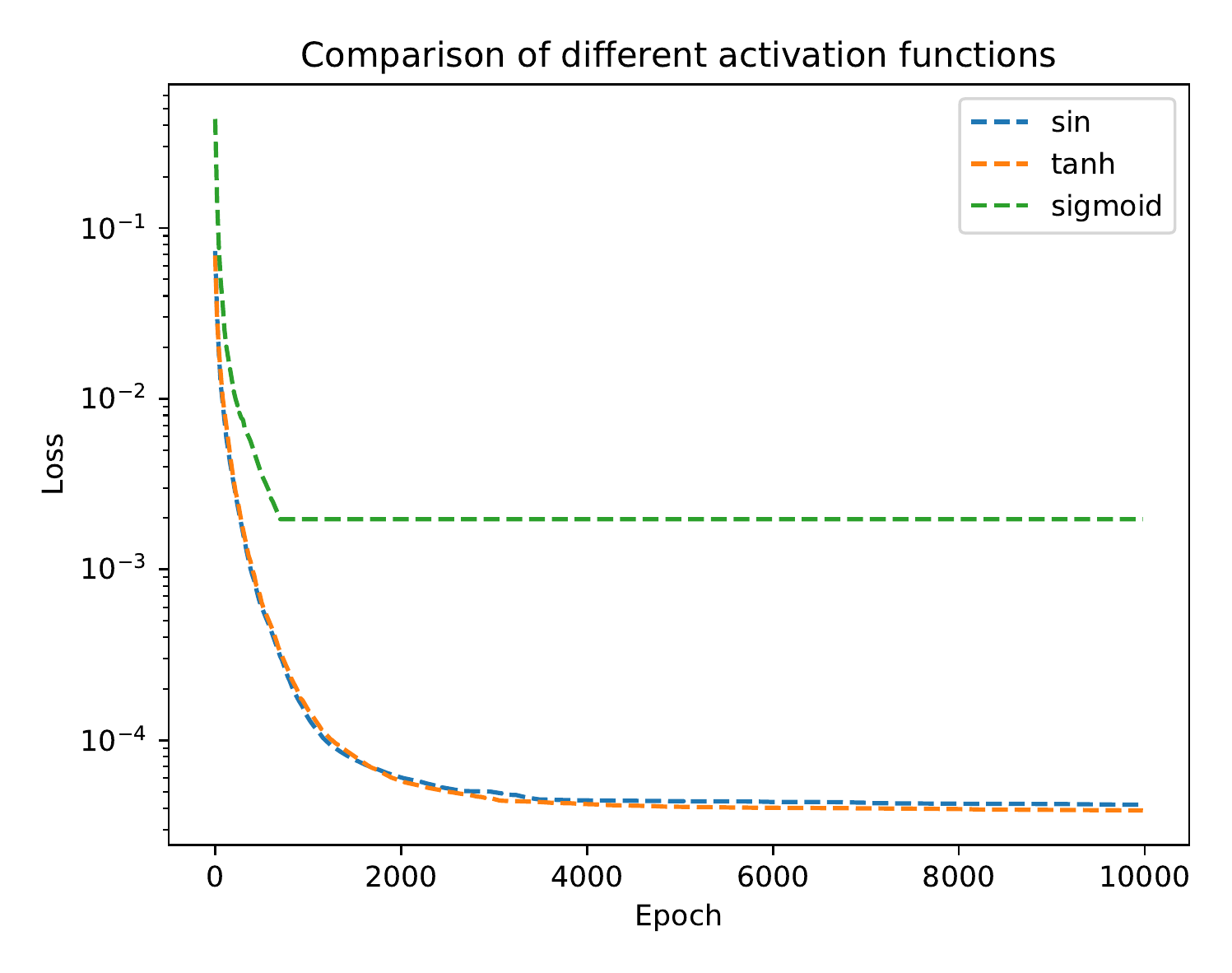}
  \end{minipage}\hfill
  \begin{minipage}{.47\linewidth}
  \includegraphics[width=1.0\textwidth]{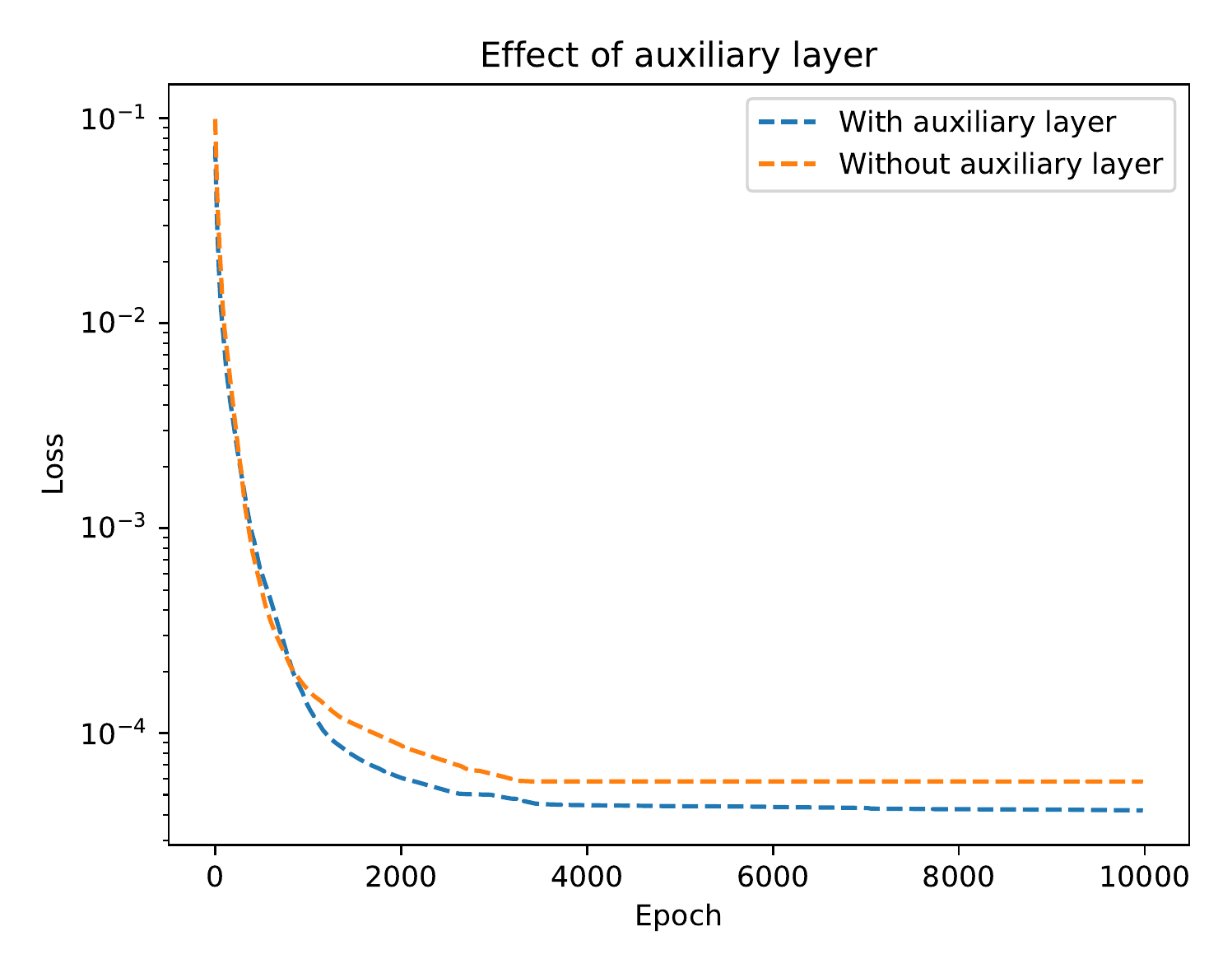}
  \end{minipage}
  \caption{Comparison of different activation functions (left) and the effect of the auxiliary layer (right) for training the GF-Net in the unit disk with the point source at the origin. Note that only the LBFGS steps of the training process are shown here.}
  \label{fig:activation}\vspace{-0.5cm}
\end{figure}

\paragraph{Effect of the activation function}
Choosing a proper activation function sometimes could be a key part to the success of neural network models, thus we compare the effect of three commonly used activation functions ($\sin$, $\tanh$ and ${\rm sigmod}$) for training GF-Net and  their performance  in predicting the Green's function. The training loss curves are displayed in Figure \ref{fig:activation} (left), which shows that both $\tanh$ and $\sin$ work much better than ${\rm sigmoid}$ in terms of the decaying of training loss. Table \ref{tab:ActiCircSingle} reports numerical errors of the predicted Green's function. It is again observed that $\tanh$ and $\sin$ easily outperform ${\rm sigmoid}$, and $\sin$ performs the best among them. Hence, $\sin$ is selected as the activation function for GF-Net in all the experiments in this work.

\paragraph{Effect of the auxiliary layer}
We also compare the decaying of training loss  with and without the auxiliary layer, see Figure \ref{fig:activation} (right). It is easy to see that the training loss decays faster when the auxiliary layer is used.

\begin{table}[!ht]\renewcommand{\arraystretch}{}
  \centering
\begin{tabular}{|c|ccc|}\hline
Activation function & $\sin$ & $\tanh$ & $\text{sigmoid}$\\\hline
Error & 1.29e-3 & 1.38e-3 & 1.72e-2  \\\hline
\end{tabular}
  \caption{Numerical errors of the predicted Green's function in the unit disk with the point source at the
origin when different activation functions are used for  GF-Net.}
  \label{tab:ActiCircSingle}\vspace{-0.5cm}
\end{table}

\subsection{Training times with respect to the domain partitioning strategy}\label{apdx:partitiontime}
\begin{figure}[!ht]
\centerline{
\begin{minipage}{.33\linewidth}
\includegraphics[width=1.95in]{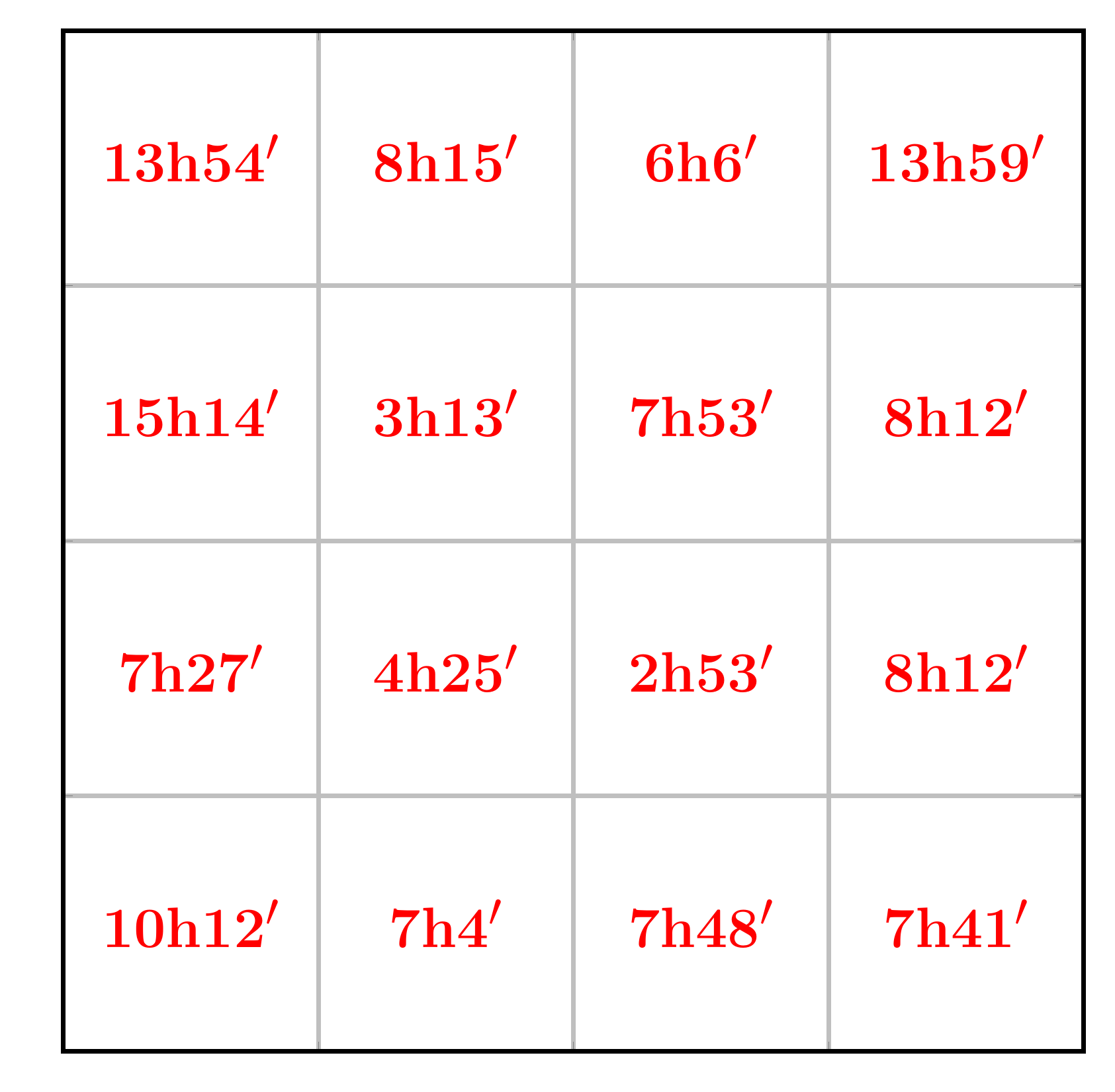}
\end{minipage}\hfill
\begin{minipage}{.33\linewidth}
\includegraphics[width=1.95in]{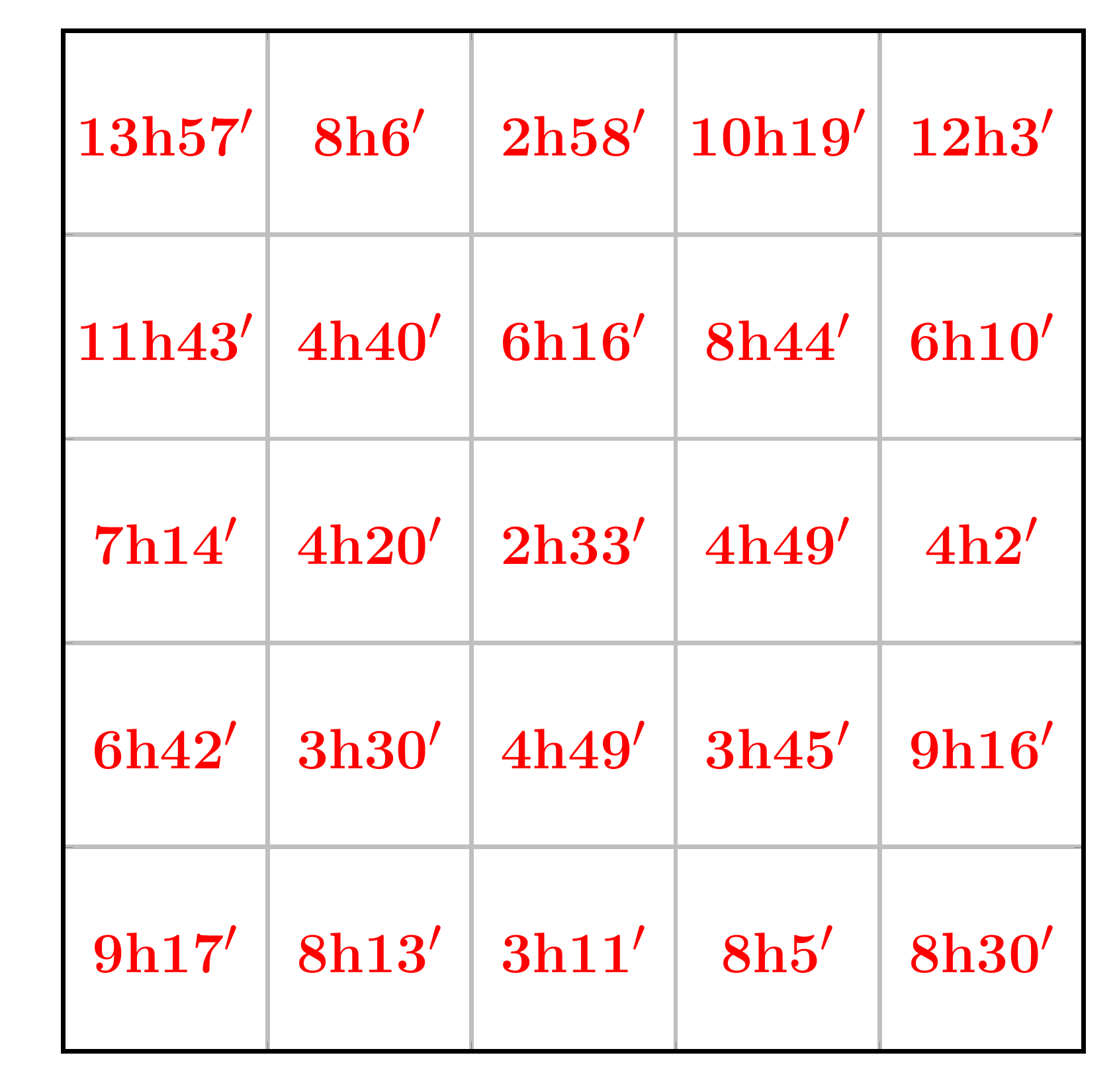}
\end{minipage}\hfill
\begin{minipage}{.33\linewidth}
\includegraphics[width=1.95in]{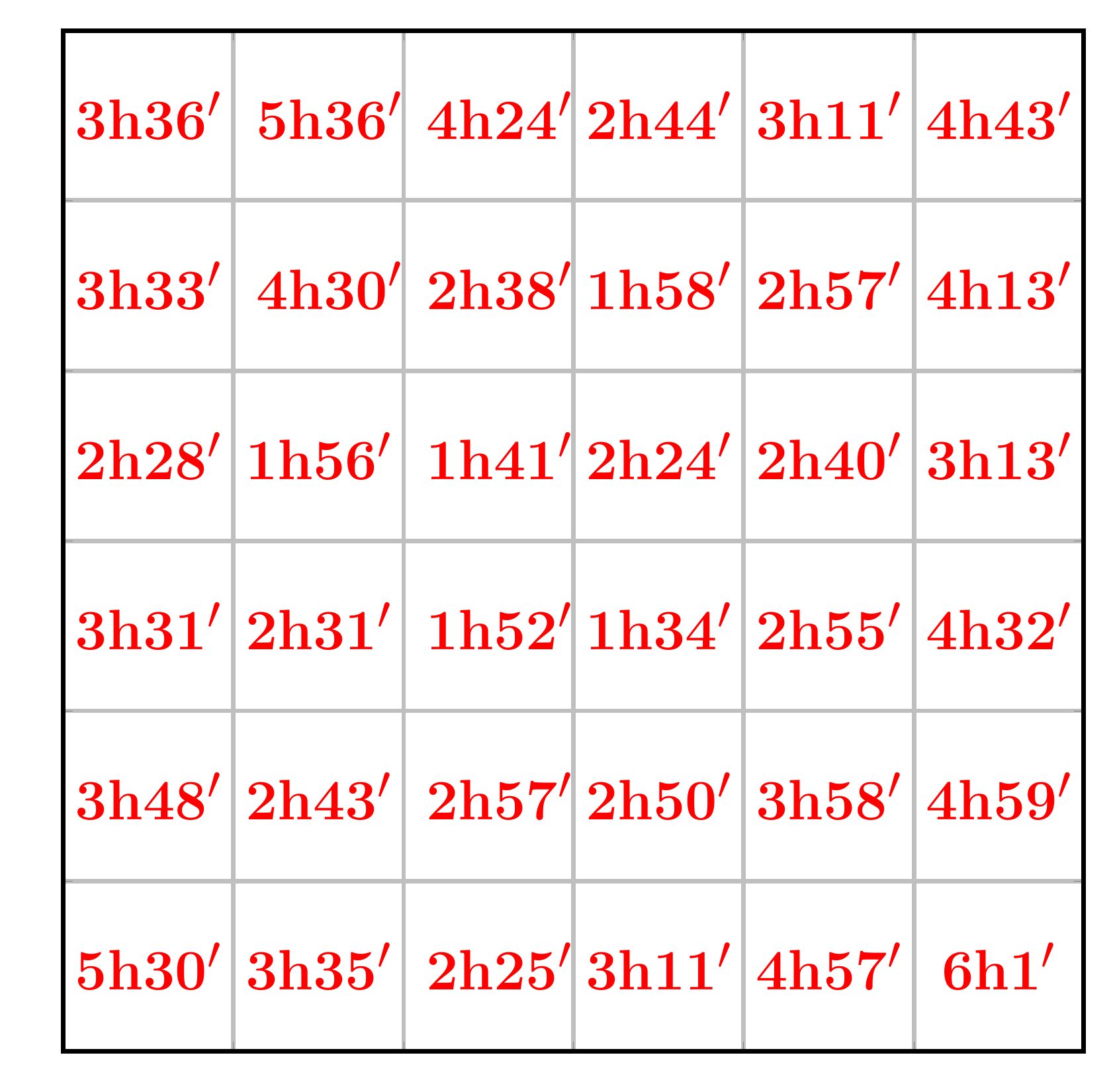}
\end{minipage}}
\caption{Training time of the GF-Net on each block for $4\times 4$ (left), $5\times 5$ (middle) and $6\times 6$ domain partitions (right).}
\label{fig:TrainTime}\vspace{-0.5cm}
\end{figure}

The computation costs of the training process under different domain partition settings are reported in Figure \ref{fig:TrainTime}.
It is observed that: 1) the training on blocks away from corners and boundaries of the domain are generally faster. In fact, it is found through experiments that the training processes for all interior blocks always terminates within several thousand LBFGS steps;  
2) the training time per block decreases as the number of blocks increases, which implies this strategy is very suitable for parallel training when many GPU cards are available.

\section{More Experiments for Poisson's Equation}\label{apdx:ExtraResults}

\begin{figure}[!ht]
 \centerline{\hspace{-0.3cm}
\subfigure[Exact solutions]{\hspace{-0.5cm}
\begin{minipage}[t]{0.31\linewidth}
\centering 
\includegraphics[width=\textwidth]{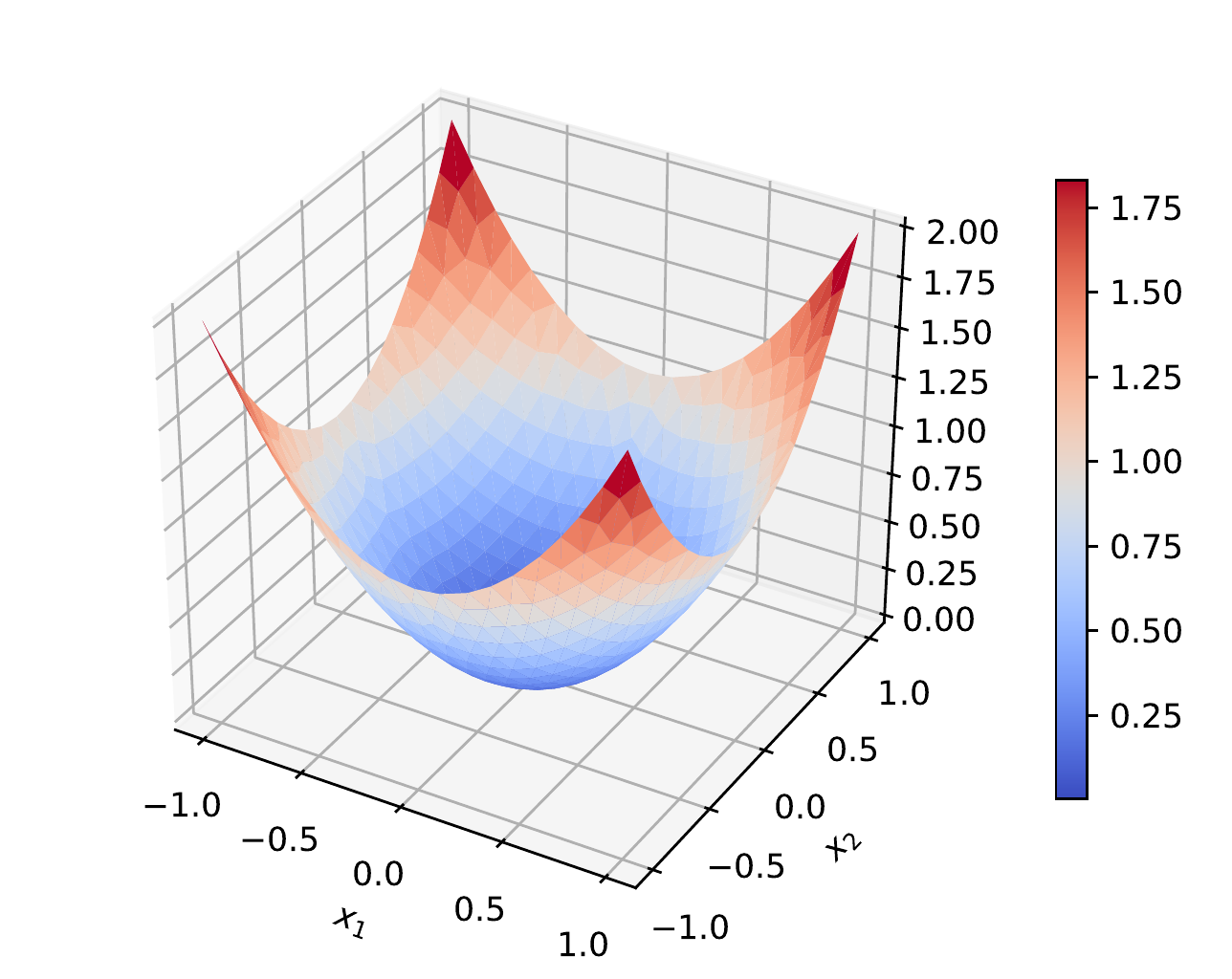}\\
 \includegraphics[width=\textwidth]{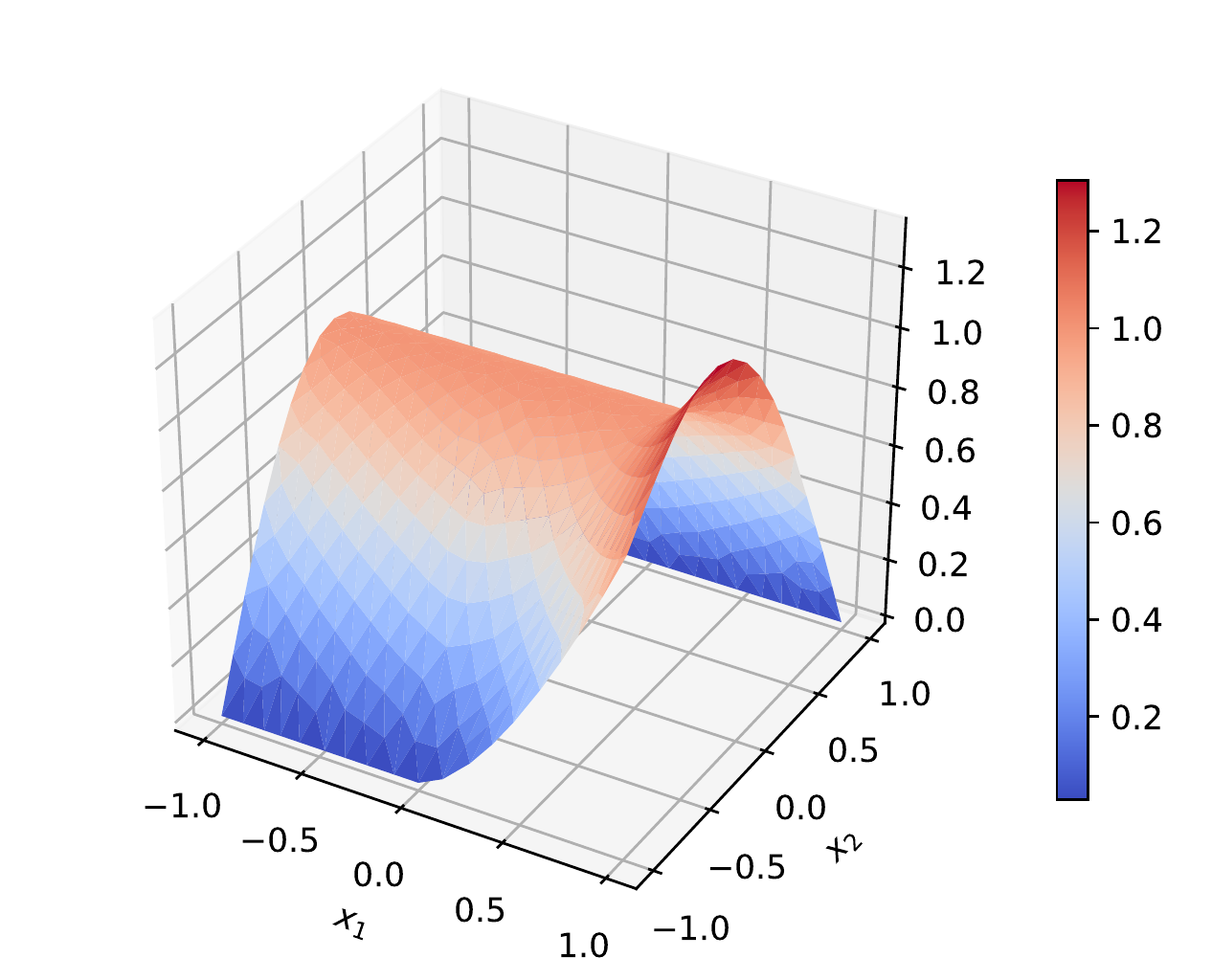}\\
 \includegraphics[width=\textwidth]{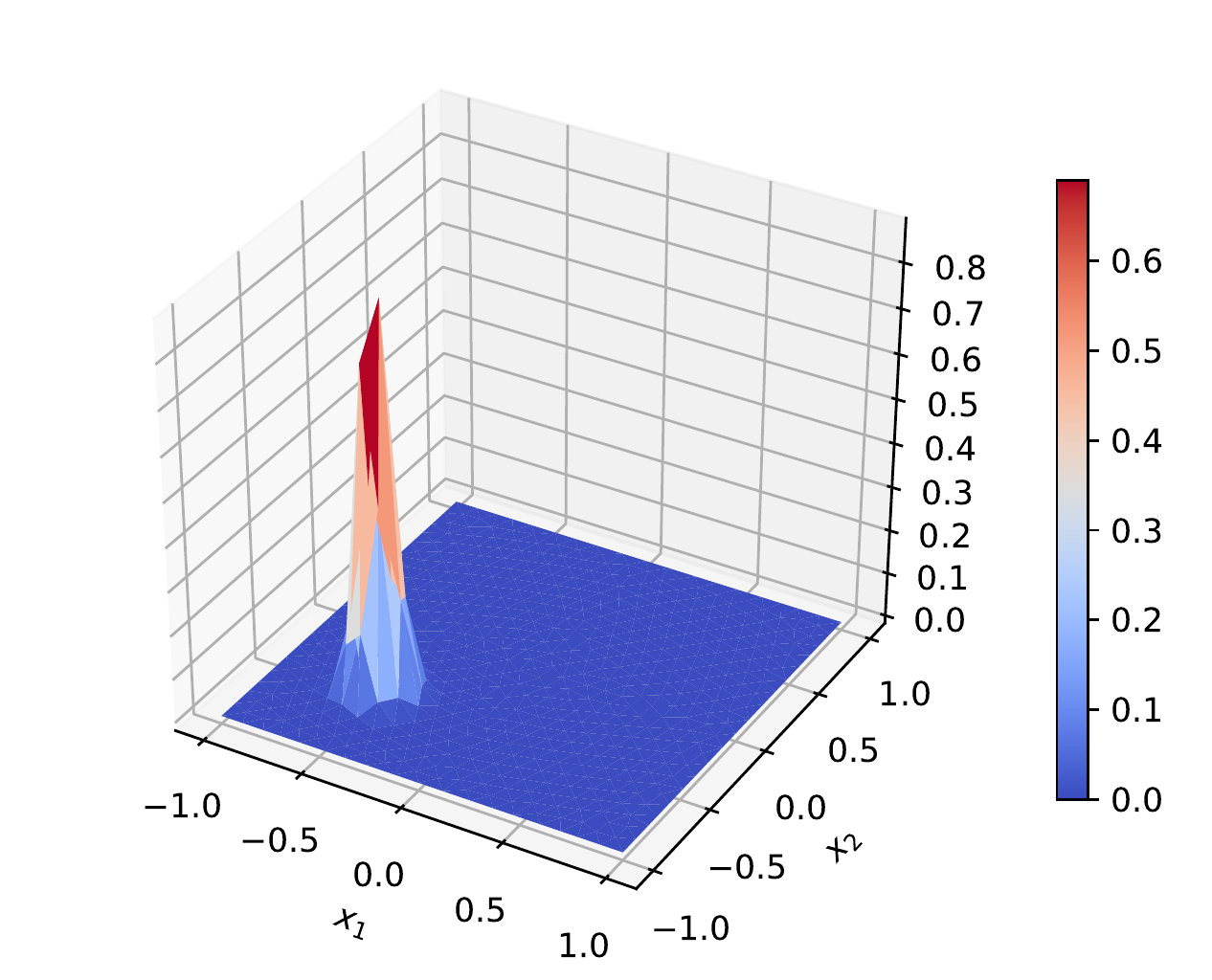}
 \end{minipage}
}
\subfigure[Predicted solutions]{
\begin{minipage}[t]{0.31\linewidth}
\centering 
\includegraphics[width=\textwidth]{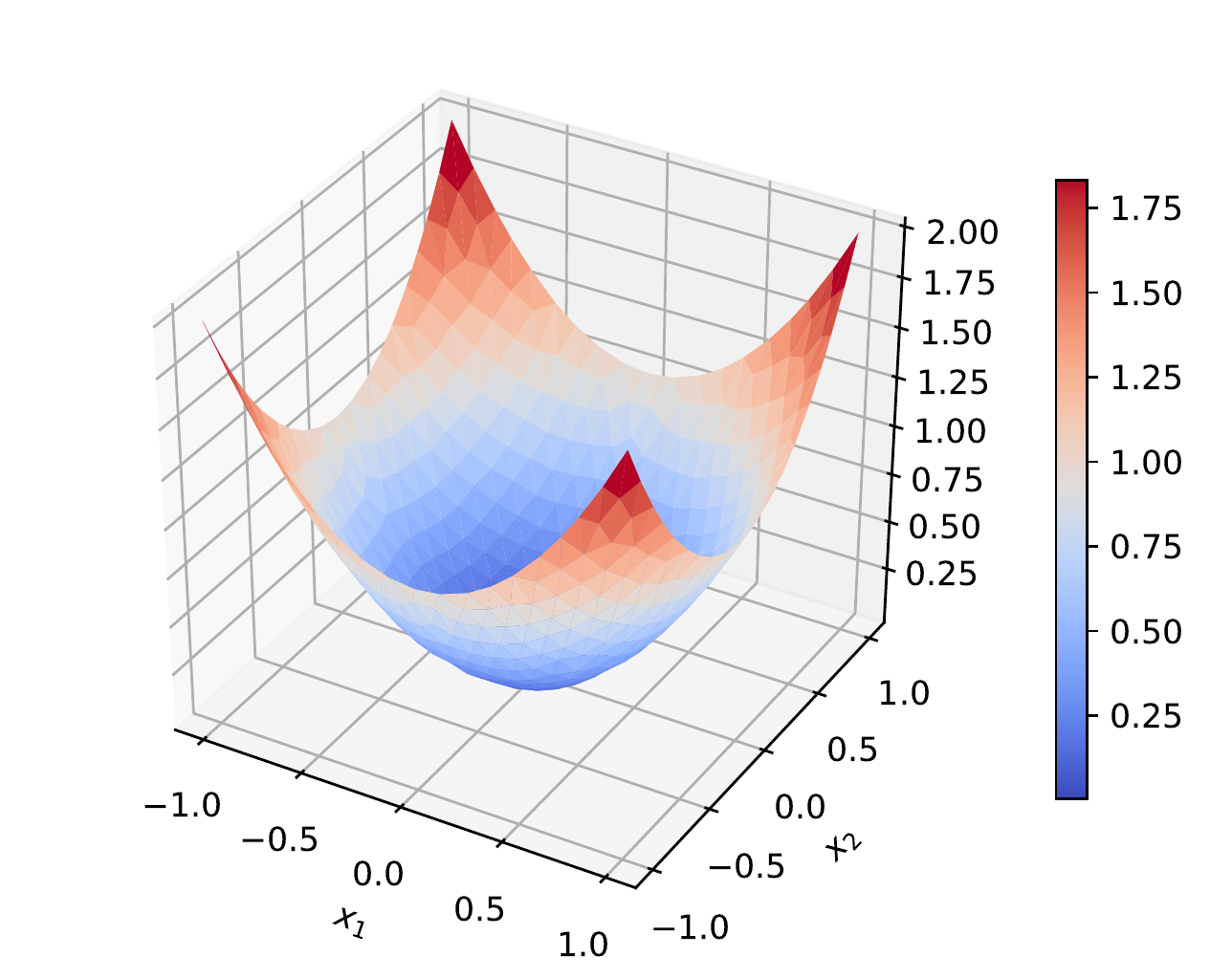}\\
 \includegraphics[width=\textwidth]{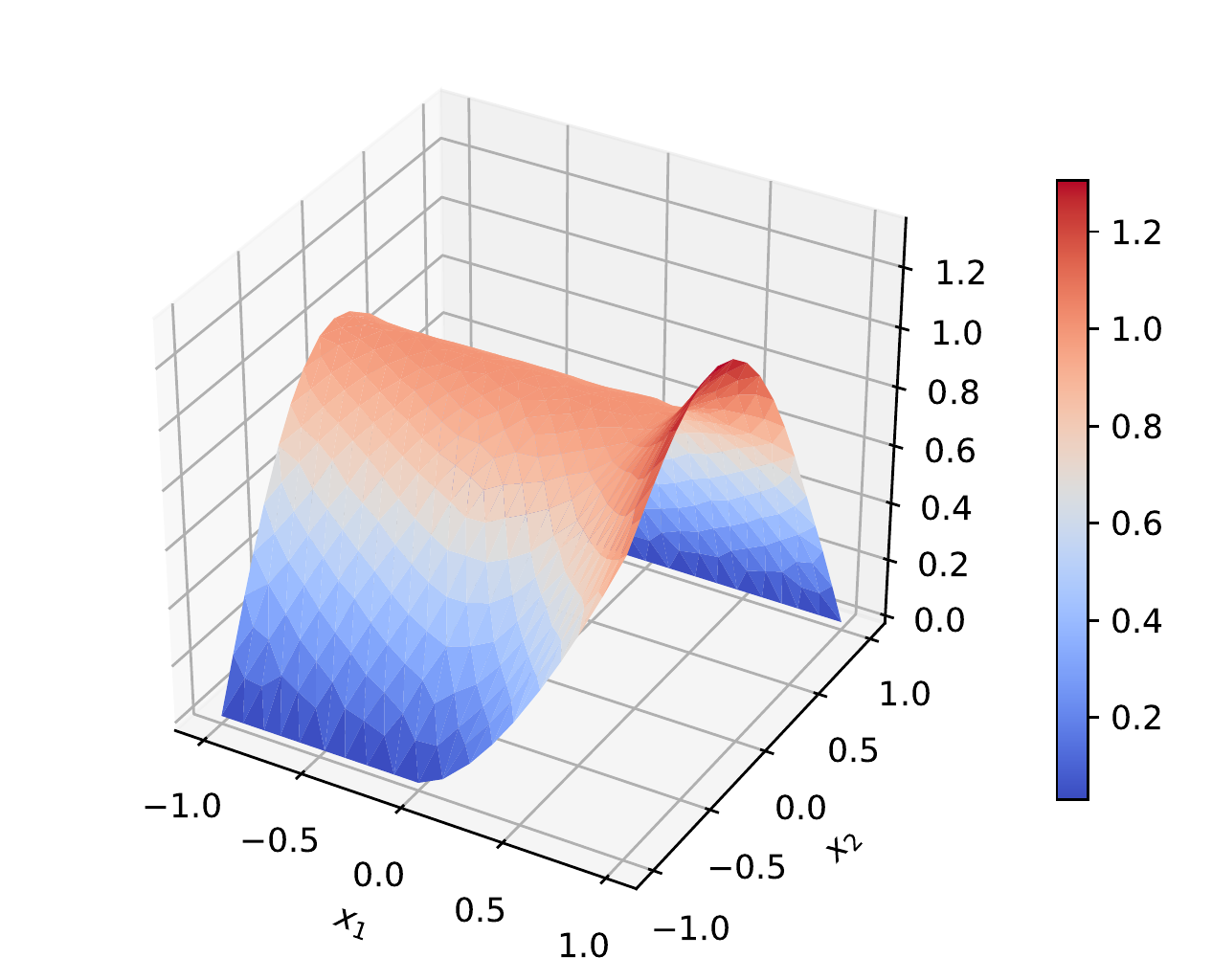}\\
 \includegraphics[width=\textwidth]{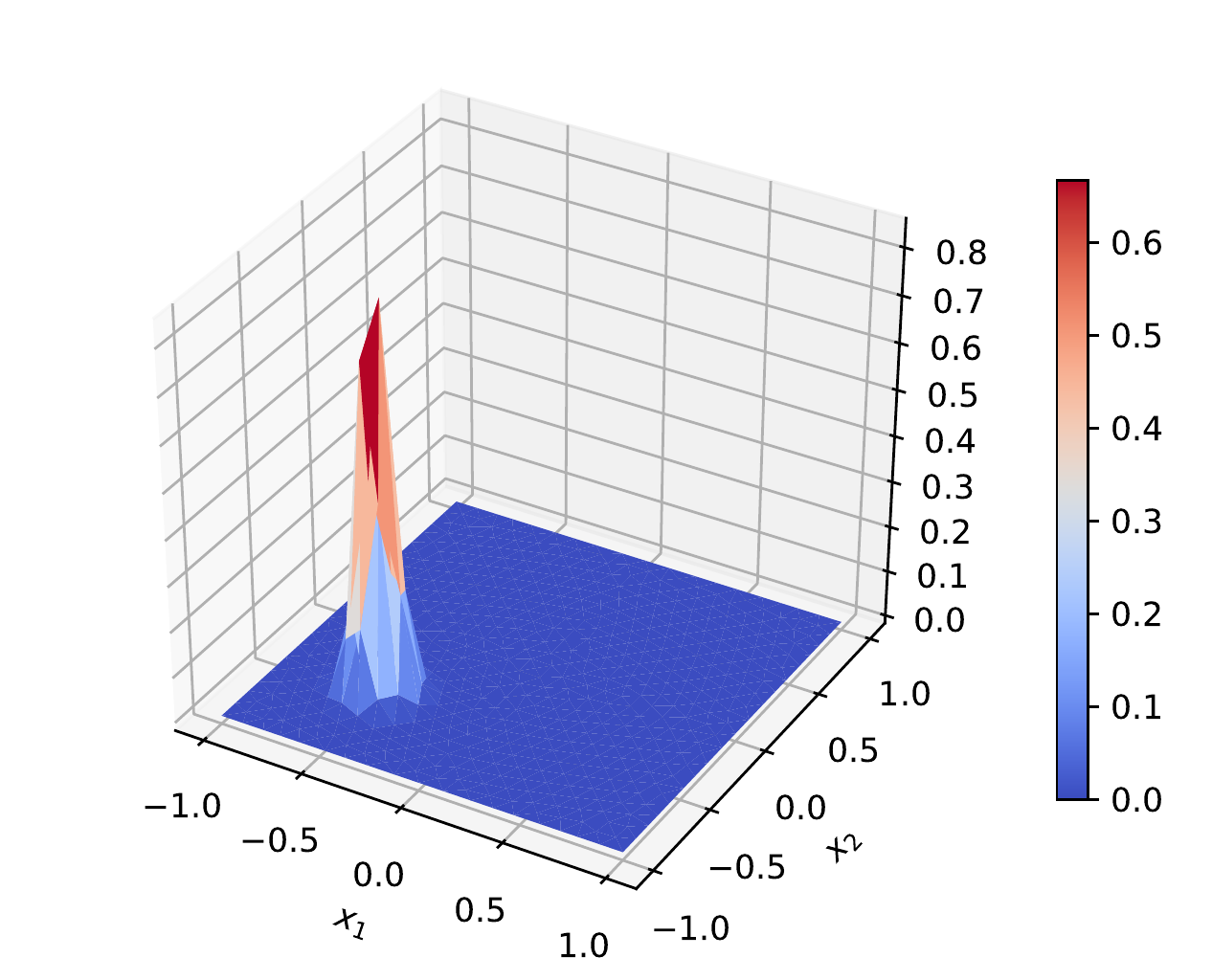}
  \end{minipage}
}
\subfigure[Errors]{
\begin{minipage}[t]{0.31\linewidth}
\centering  
\includegraphics[width=\textwidth]{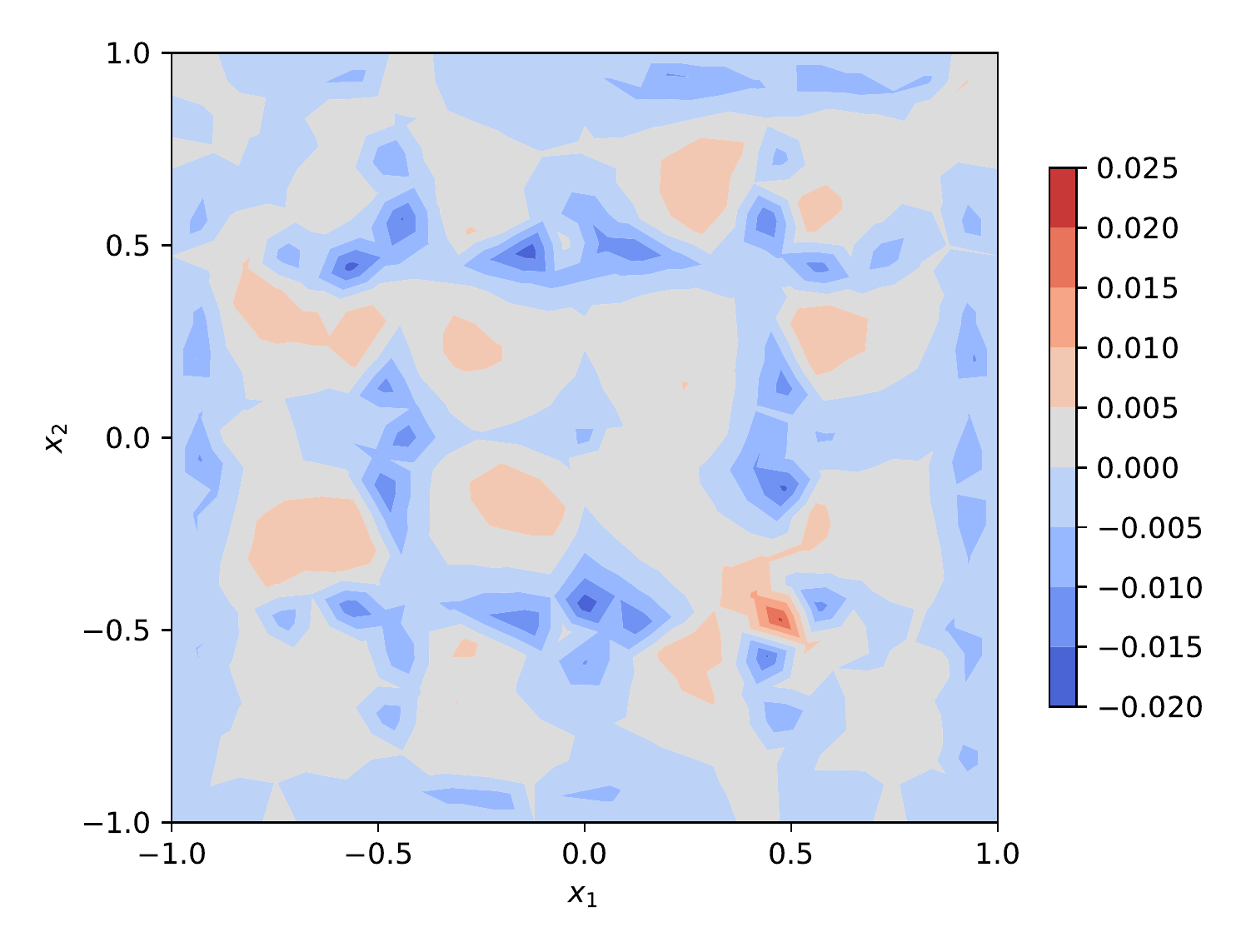}\\ 
 \includegraphics[width=\textwidth]{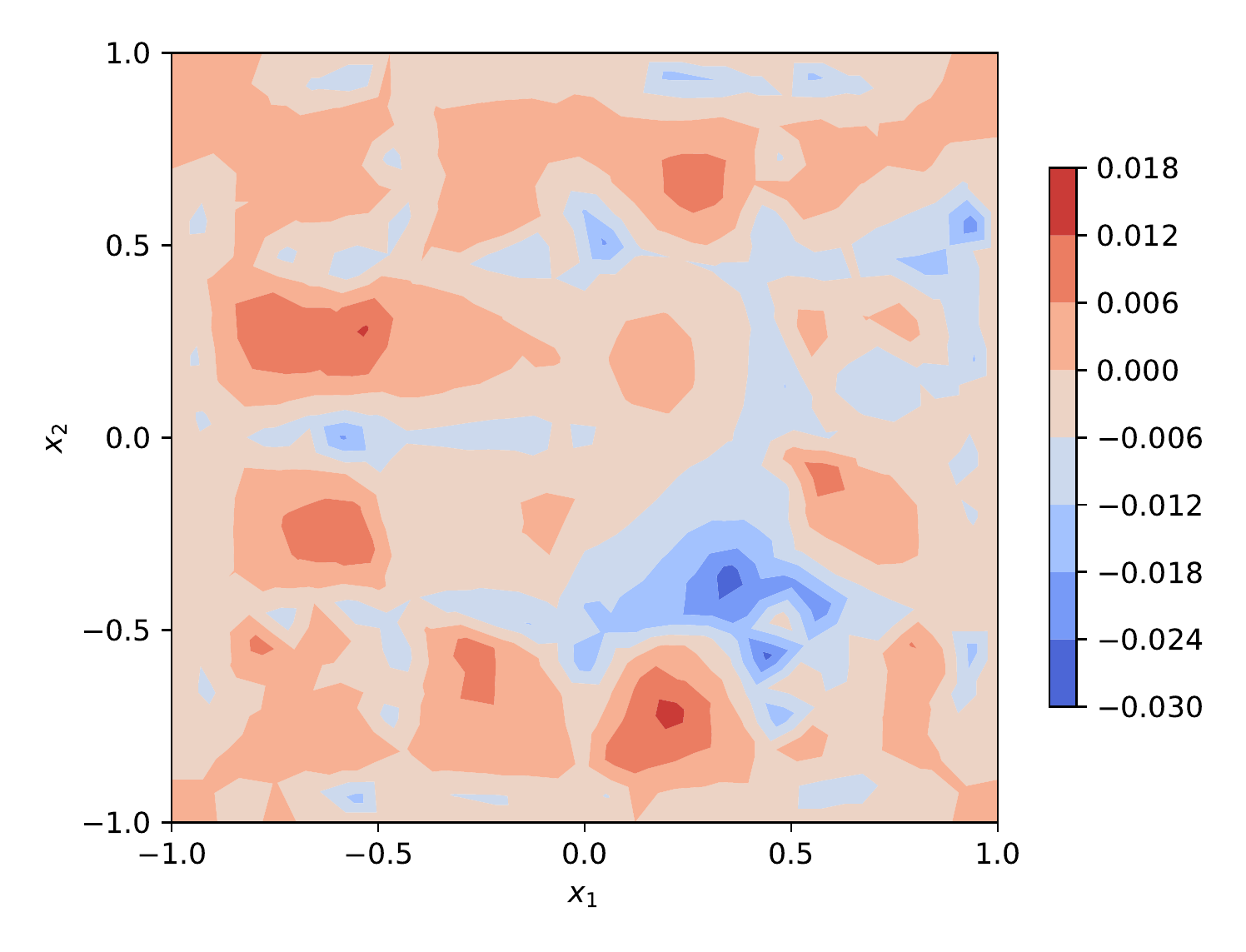}\\ 
 \includegraphics[width=\textwidth]{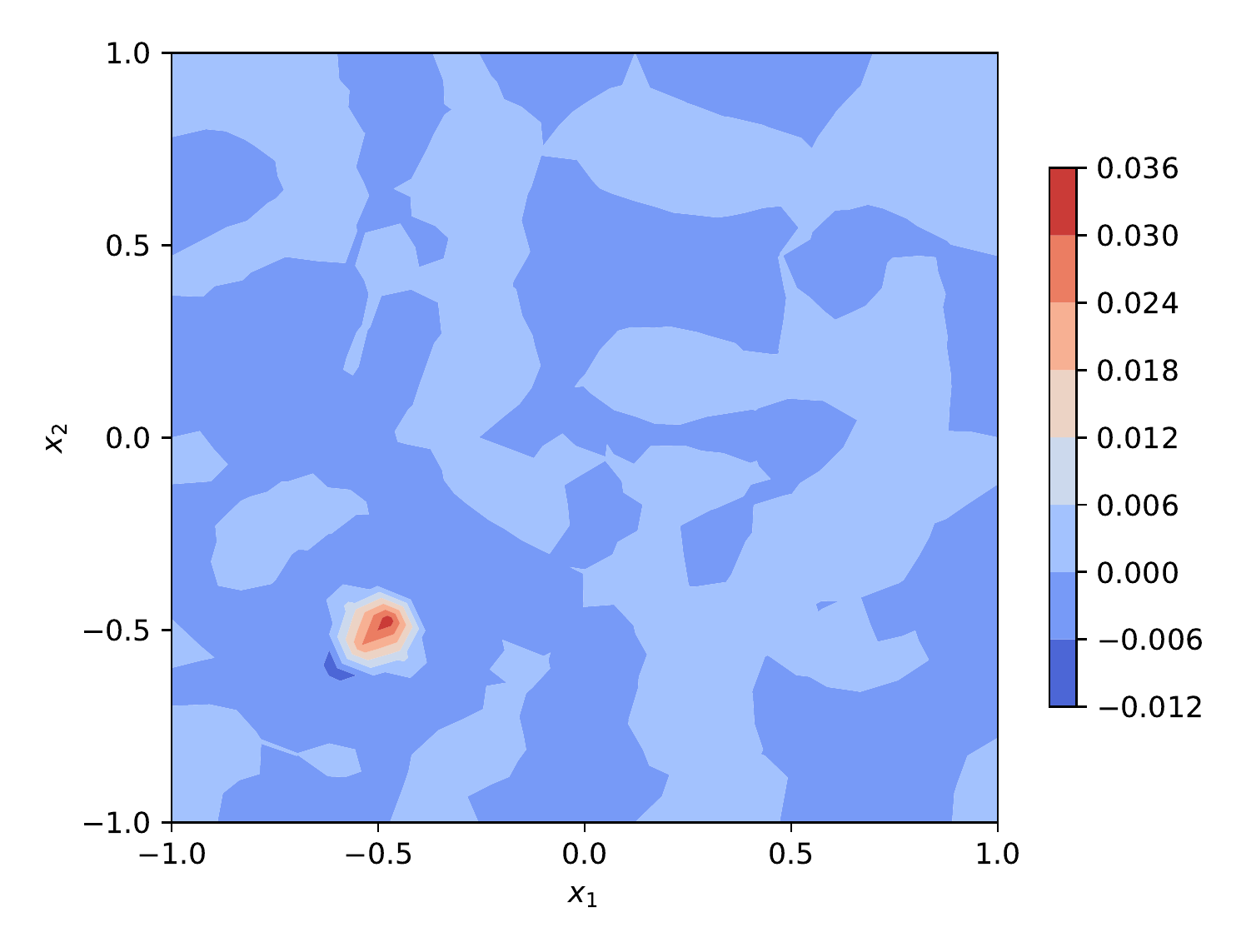}
  \end{minipage}
}}
  \caption{ 
  Numerical result for the solution of Poisson's equation in $\Omega_1$ under different sets of  source terms and boundary conditions, where the exact solutions (left), the predicted solutions by GF-Nets (middle), and the numerical errors (right) are presented. First row: Case \eqref{B1}; second row: Case \eqref{B2}; and last row: Case \eqref{B3}. 
  }
  \label{fig:apdxB1}
 \vspace{-0.3cm}
\end{figure}

\begin{figure}[!ht]
\centerline{\hspace{-0.3cm}
\subfigure[Exact solution]{\hspace{-0.5cm}
\begin{minipage}[t]{0.31\linewidth}
\centering 
\includegraphics[width=\textwidth]{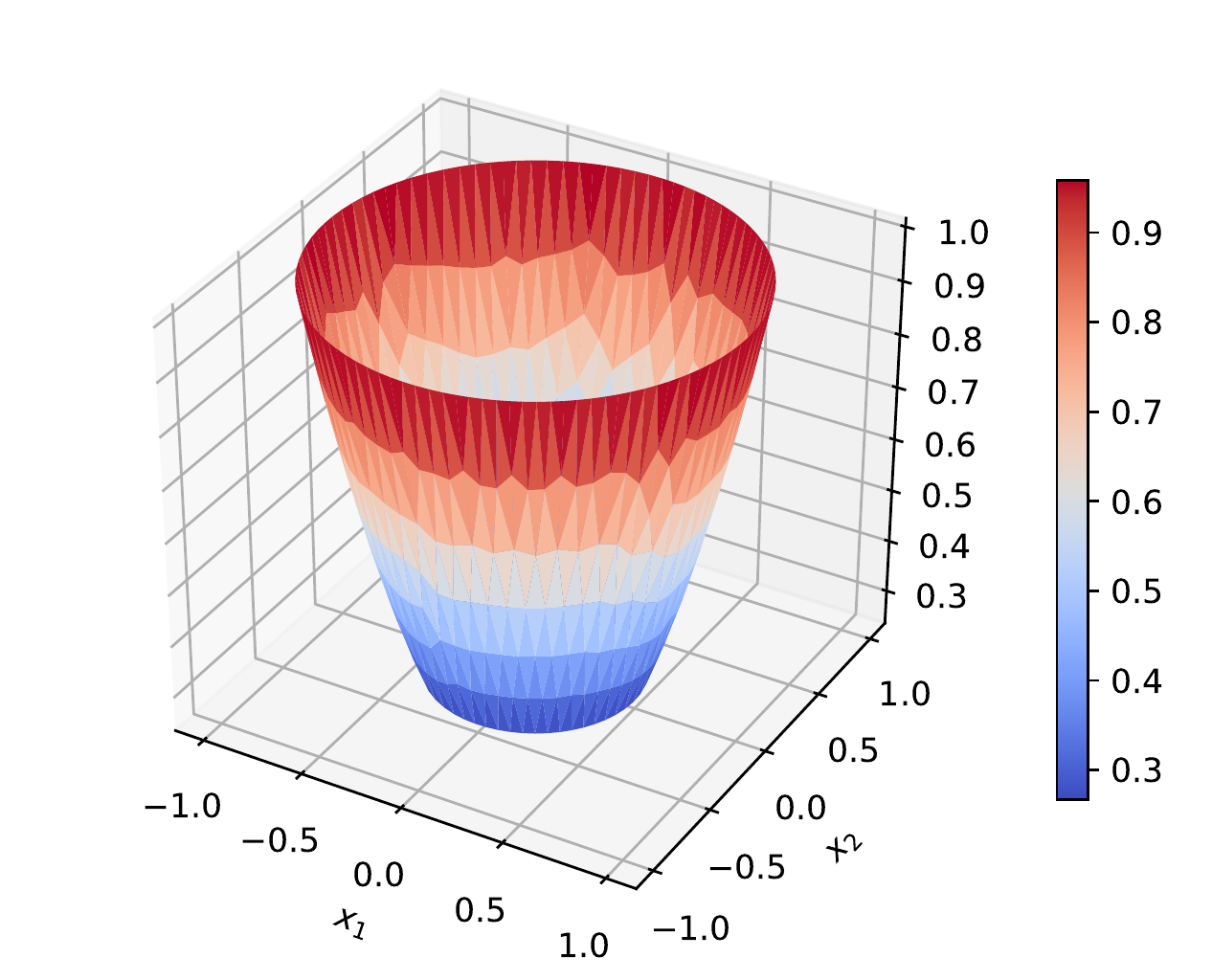}\\
 \includegraphics[width=\textwidth]{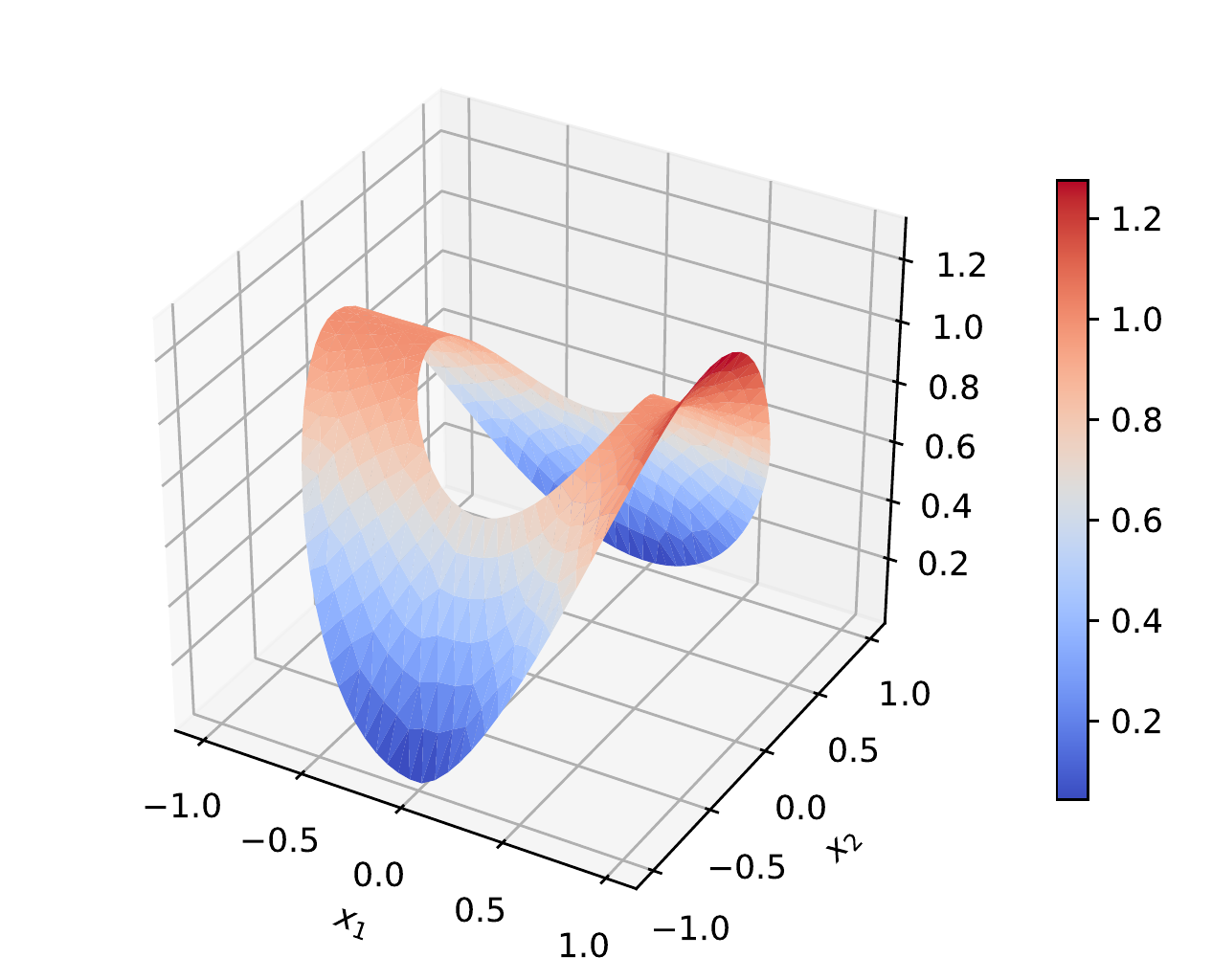}\\
 \includegraphics[width=\textwidth]{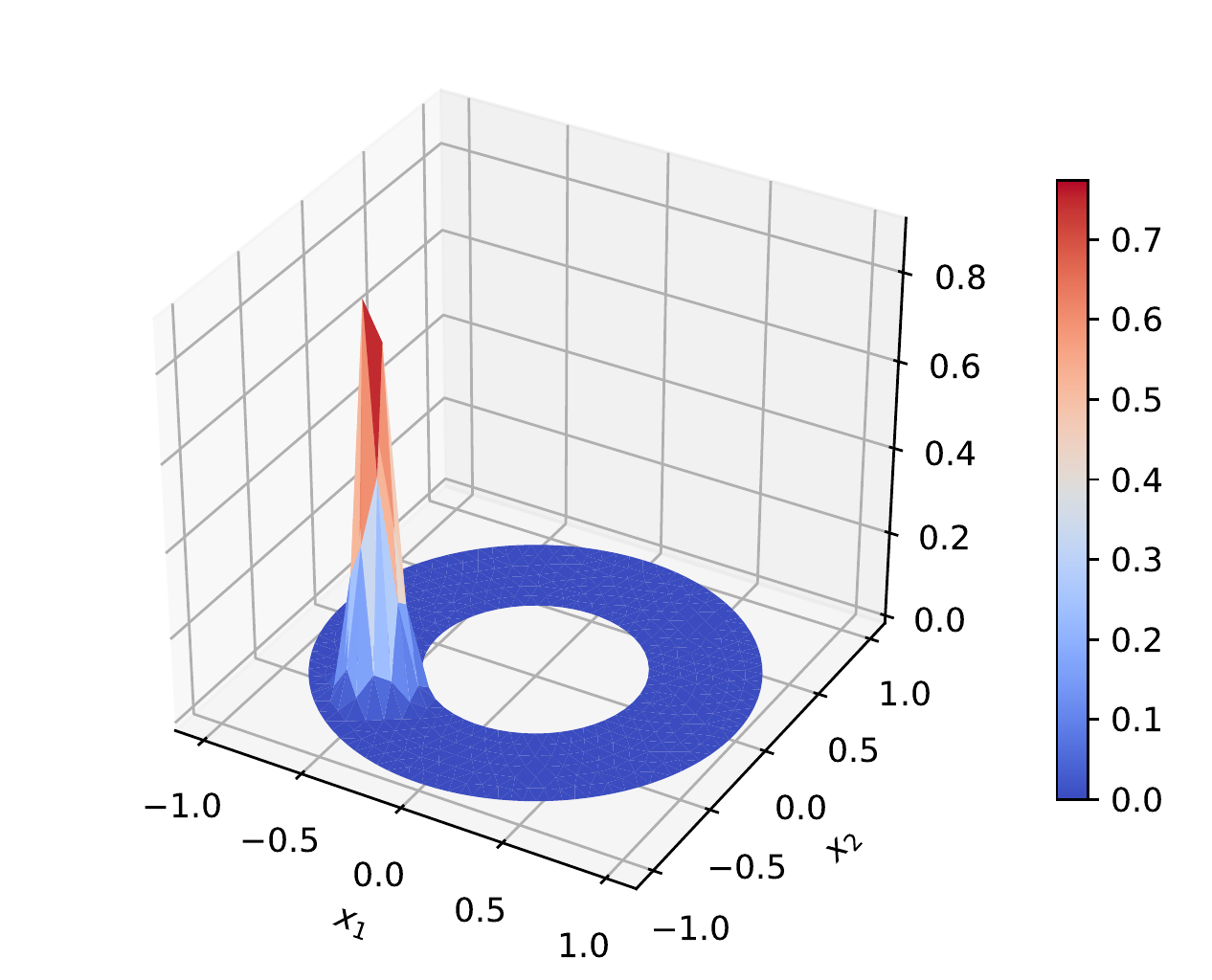}
 \end{minipage}
}
\subfigure[Predicted solution]{
\begin{minipage}[t]{0.31\linewidth}
\centering 
\includegraphics[width=\textwidth]{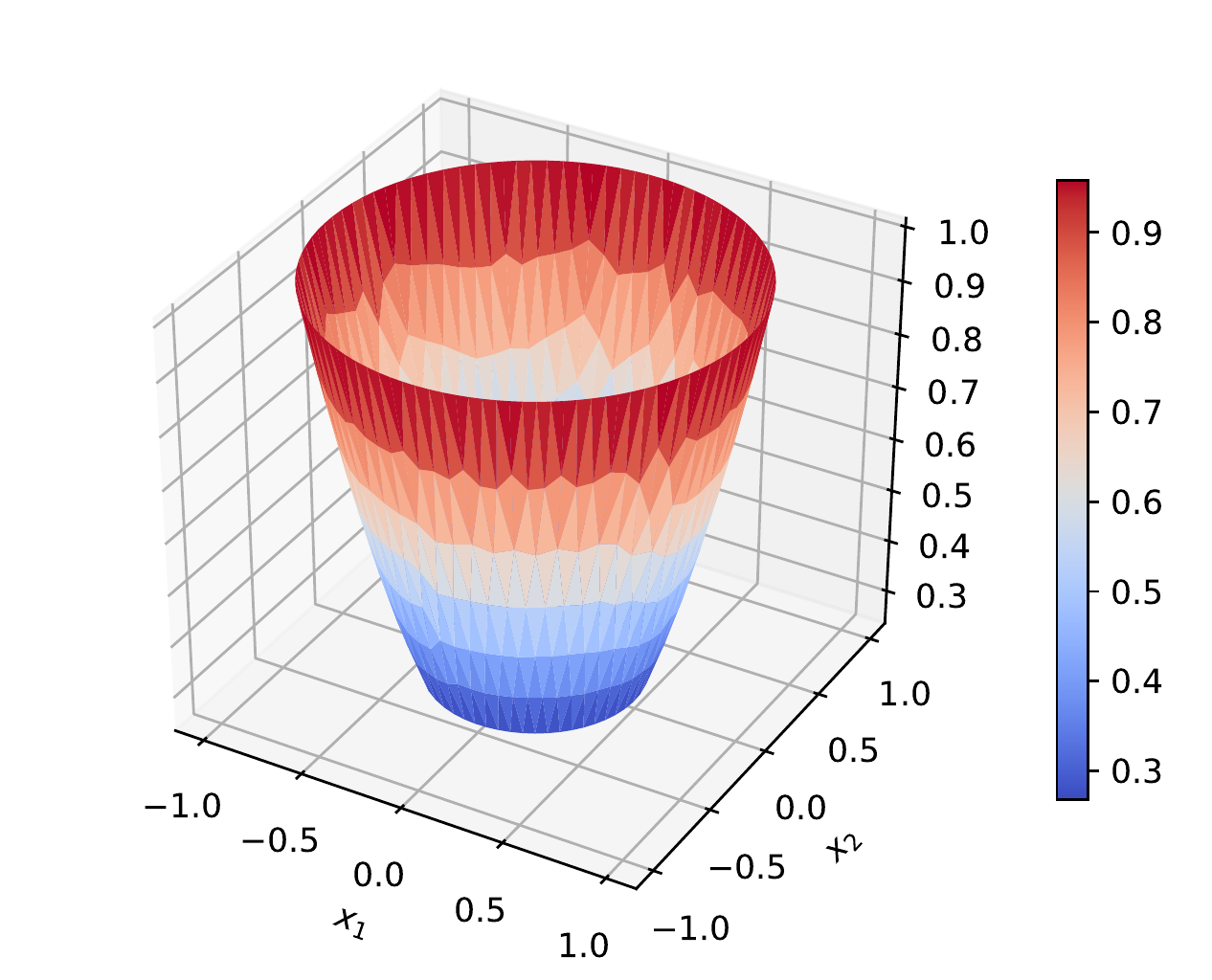}\\
 \includegraphics[width=\textwidth]{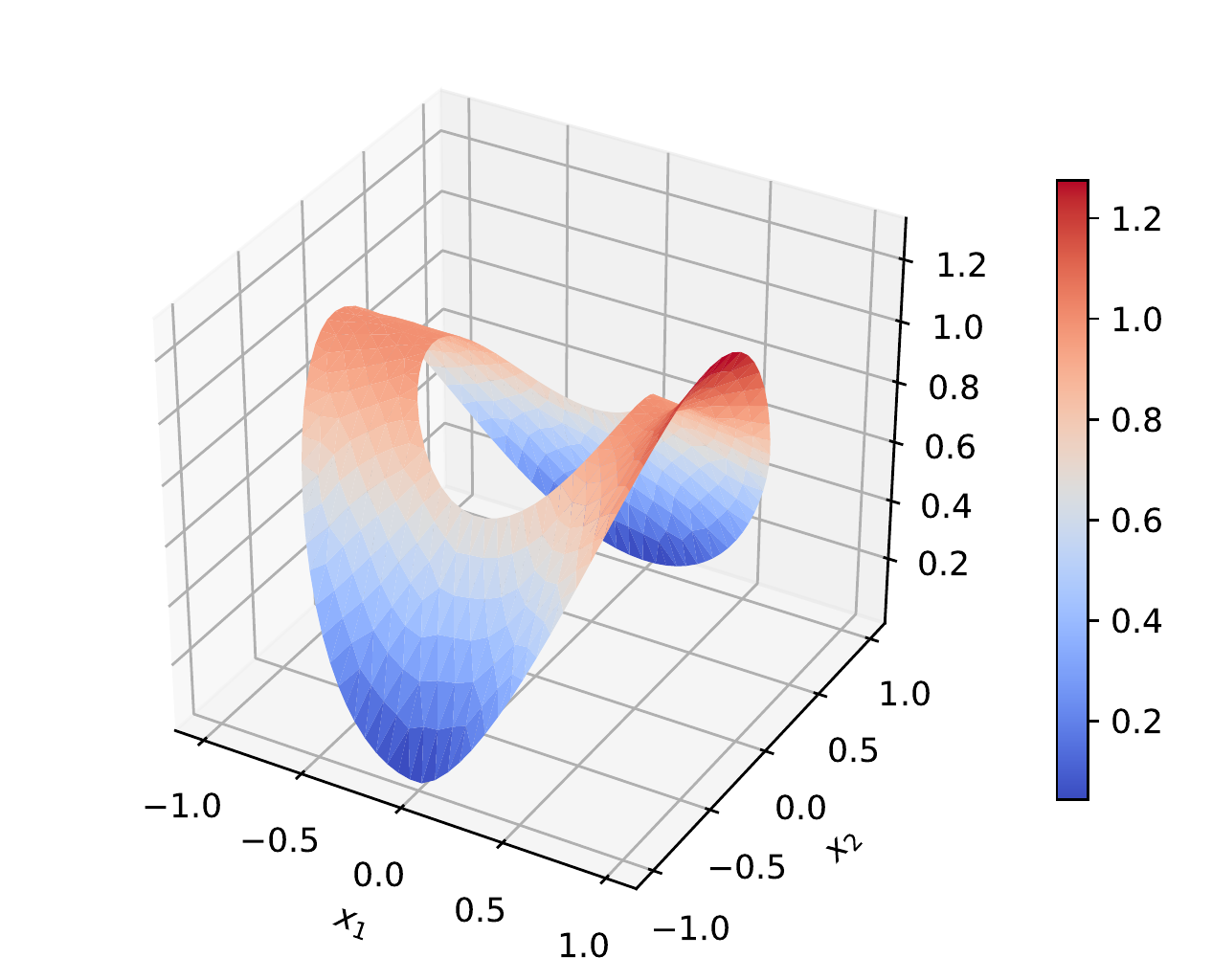}\\
 \includegraphics[width=\textwidth]{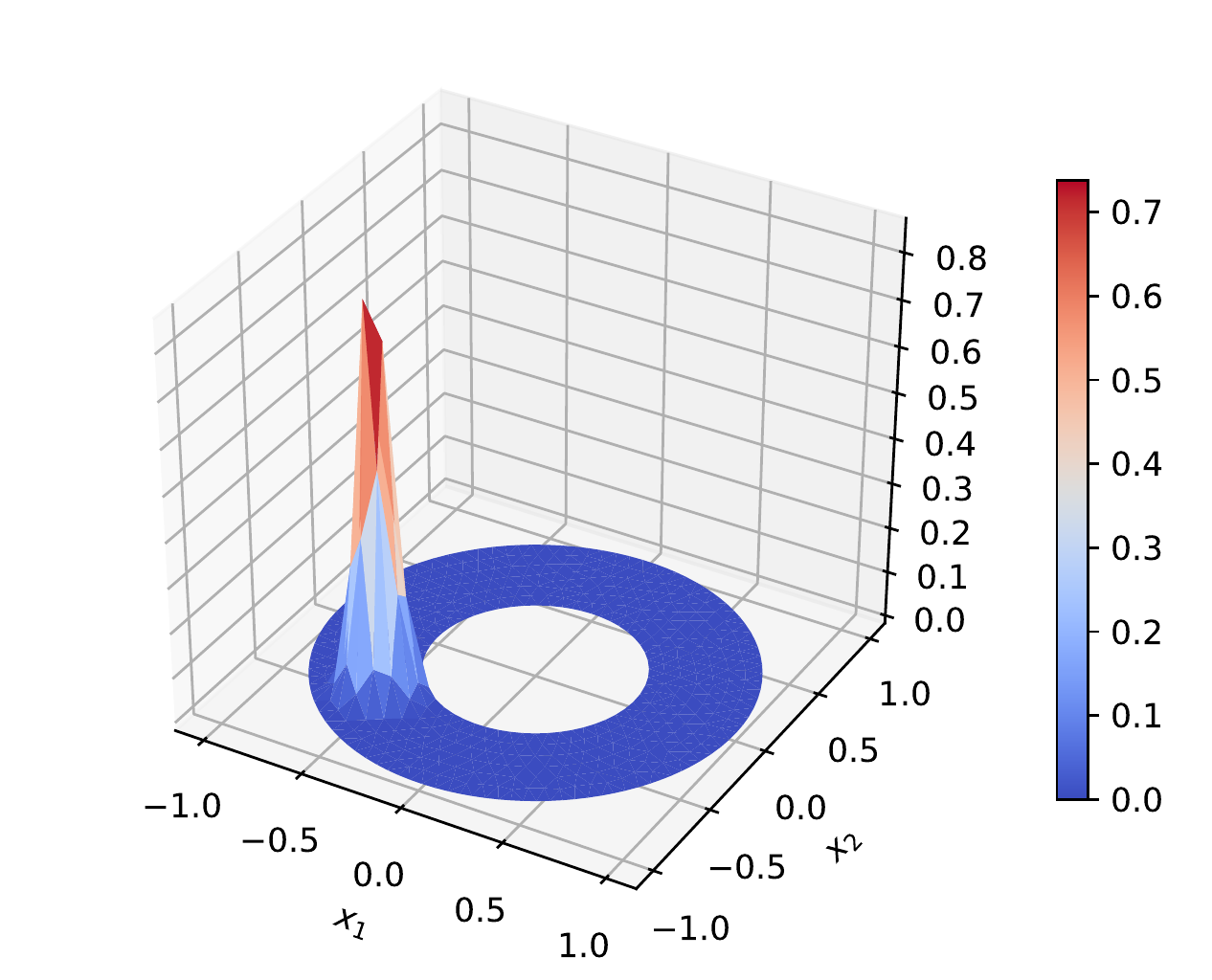}
  \end{minipage}
}
\subfigure[Error]{
\begin{minipage}[t]{0.32\linewidth}
\centering  
\includegraphics[width=\textwidth]{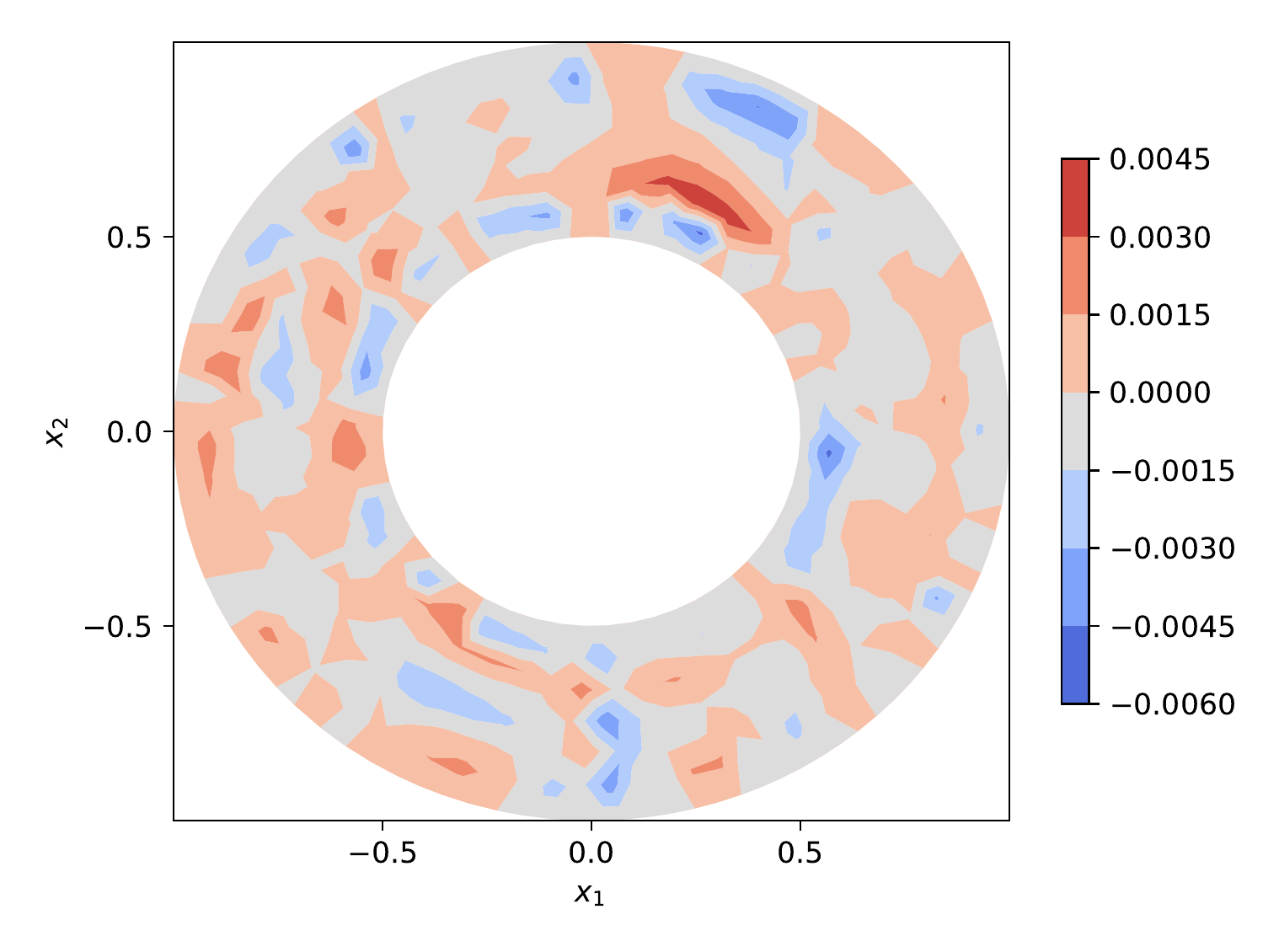}\\ 
 \includegraphics[width=\textwidth]{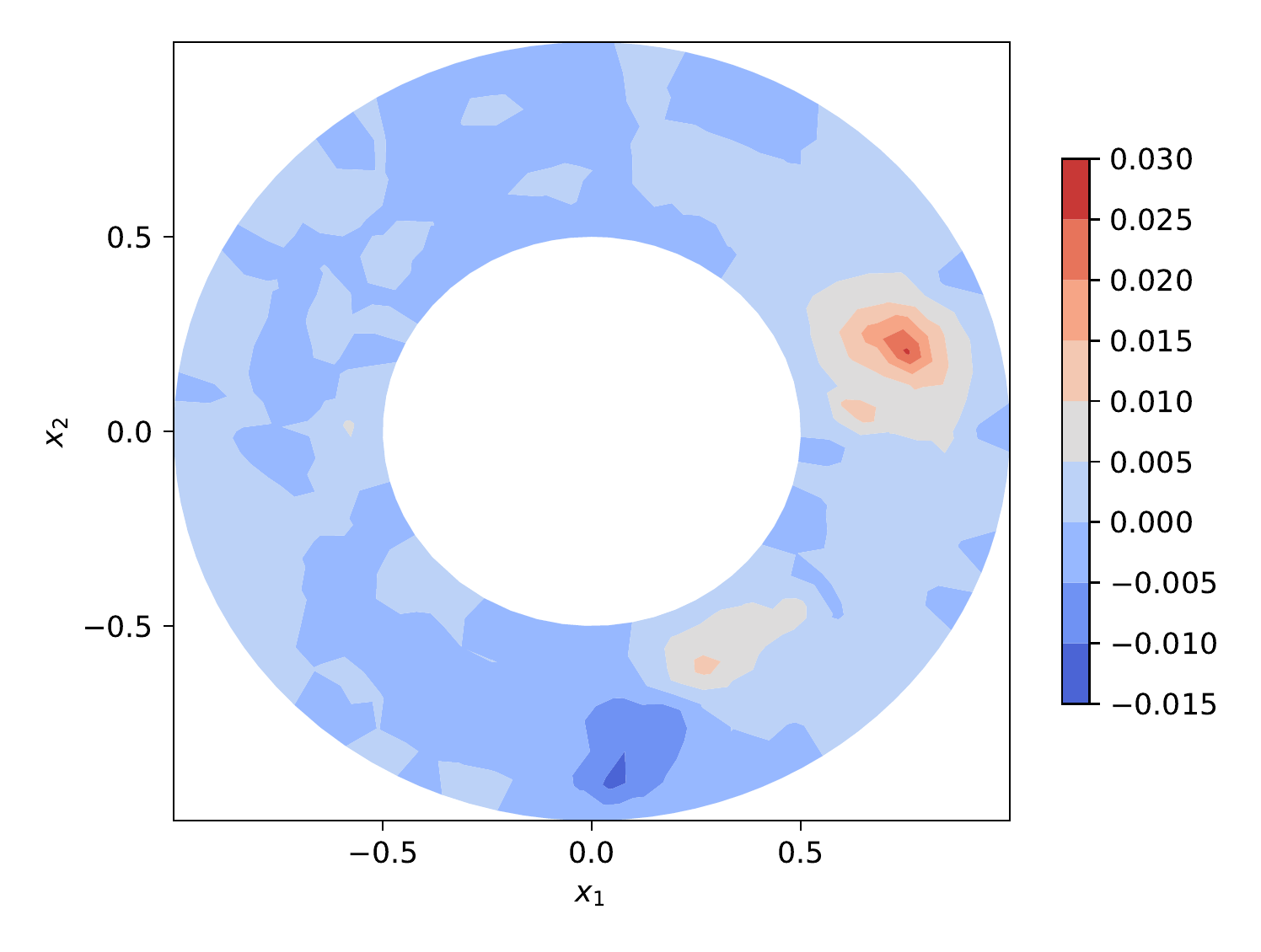}\\ 
 \includegraphics[width=\textwidth]{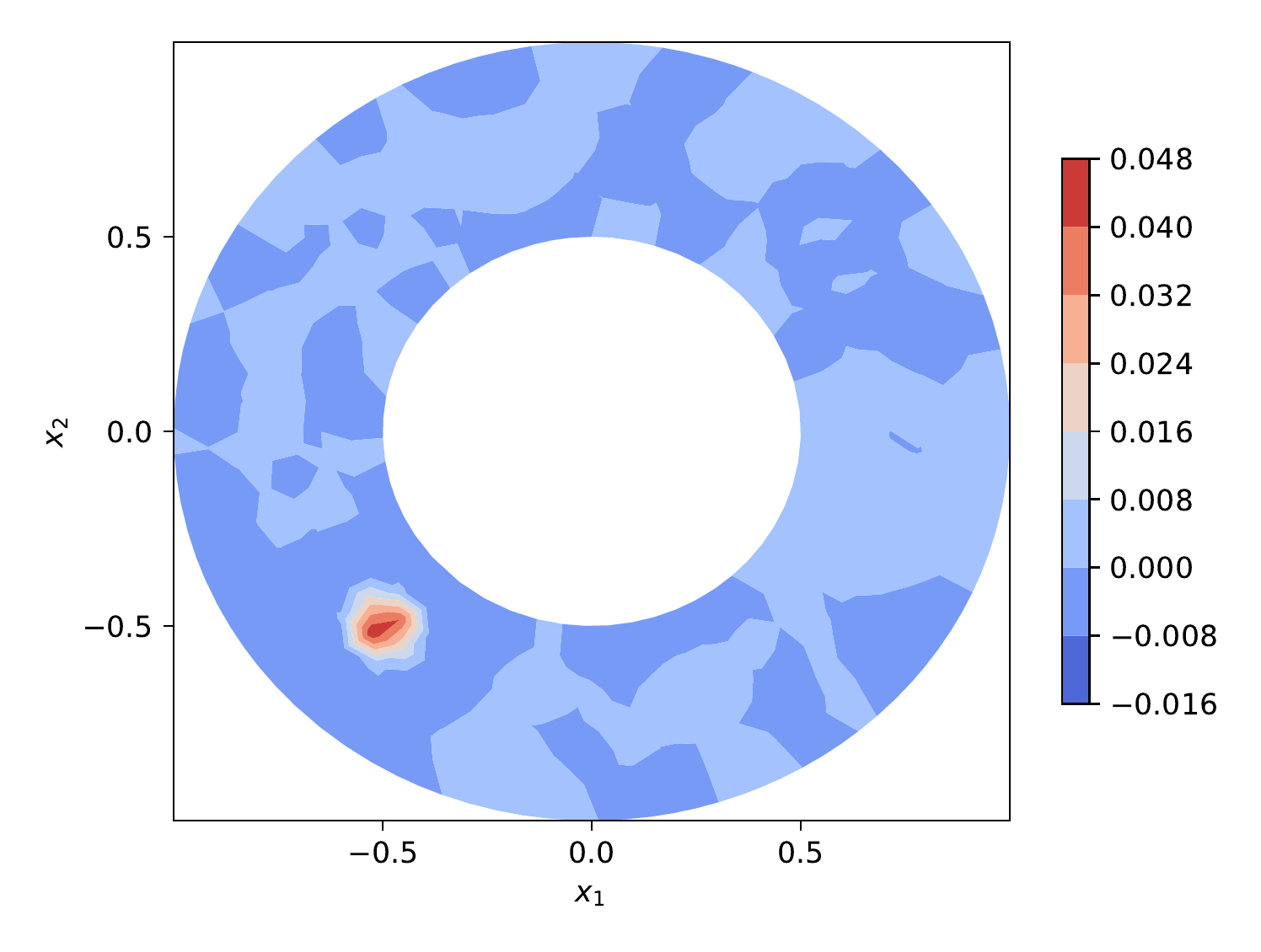}
  \end{minipage}
}}
  \caption{ 
  Numerical tests on Poisson's equation in $\Omega_2$ under different sets of  source terms and boundary conditions, where the exact solutions (left), the predicted solutions by GF-Nets (middle), and the numerical errors (right) are presented. First row: Case \eqref{B1}; second row: Case \eqref{B2}; and last row: Case \eqref{B3}.} 
  \label{fig:apdxB2}
 \vspace{-0.3cm}
\end{figure}

\begin{figure}[!ht]
 \centerline{\hspace{-0.3cm}
\subfigure[Exact solution]{\hspace{-0.5cm}
\begin{minipage}[t]{0.31\linewidth}
\centering 
\includegraphics[width=\textwidth]{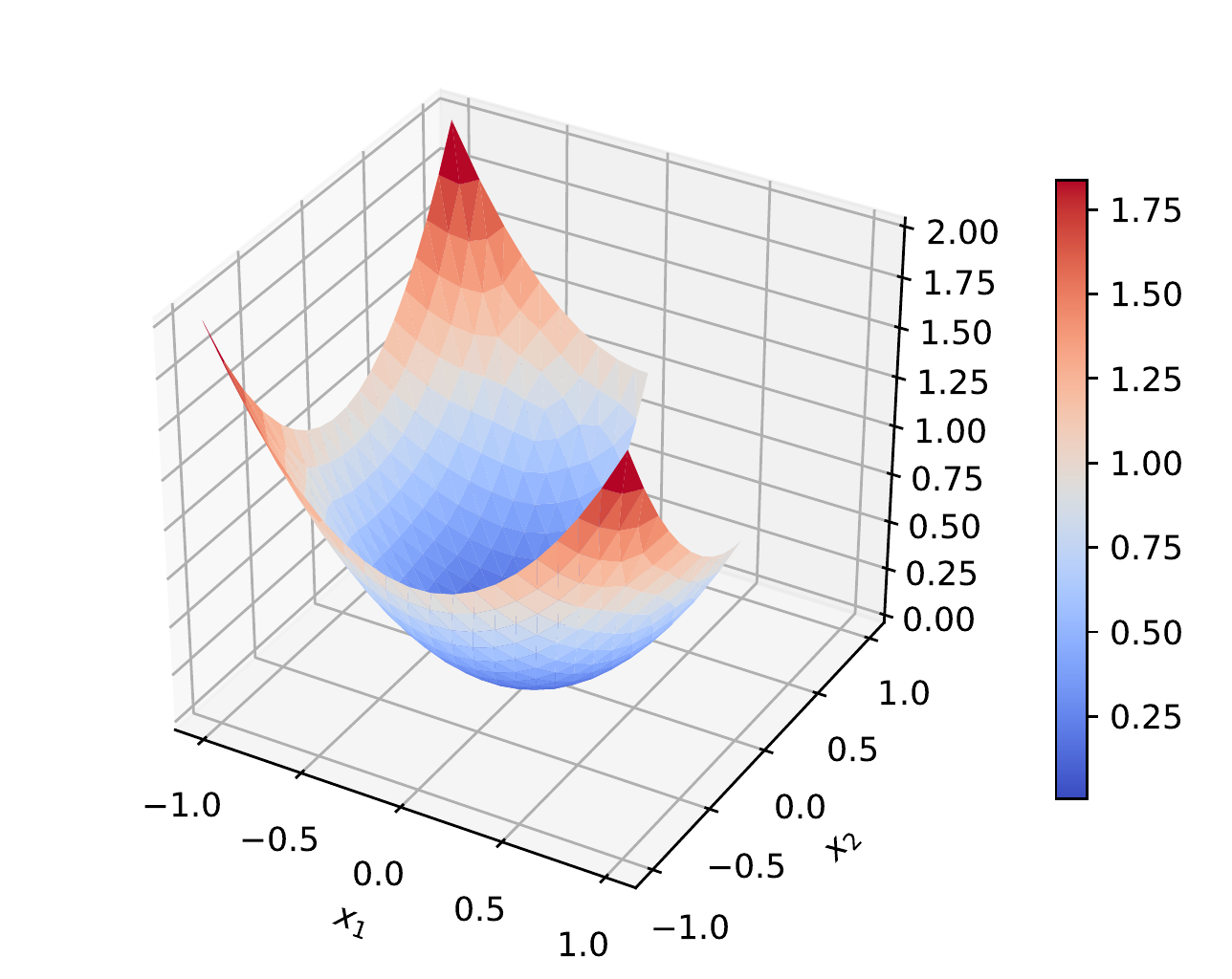}\\
 \includegraphics[width=\textwidth]{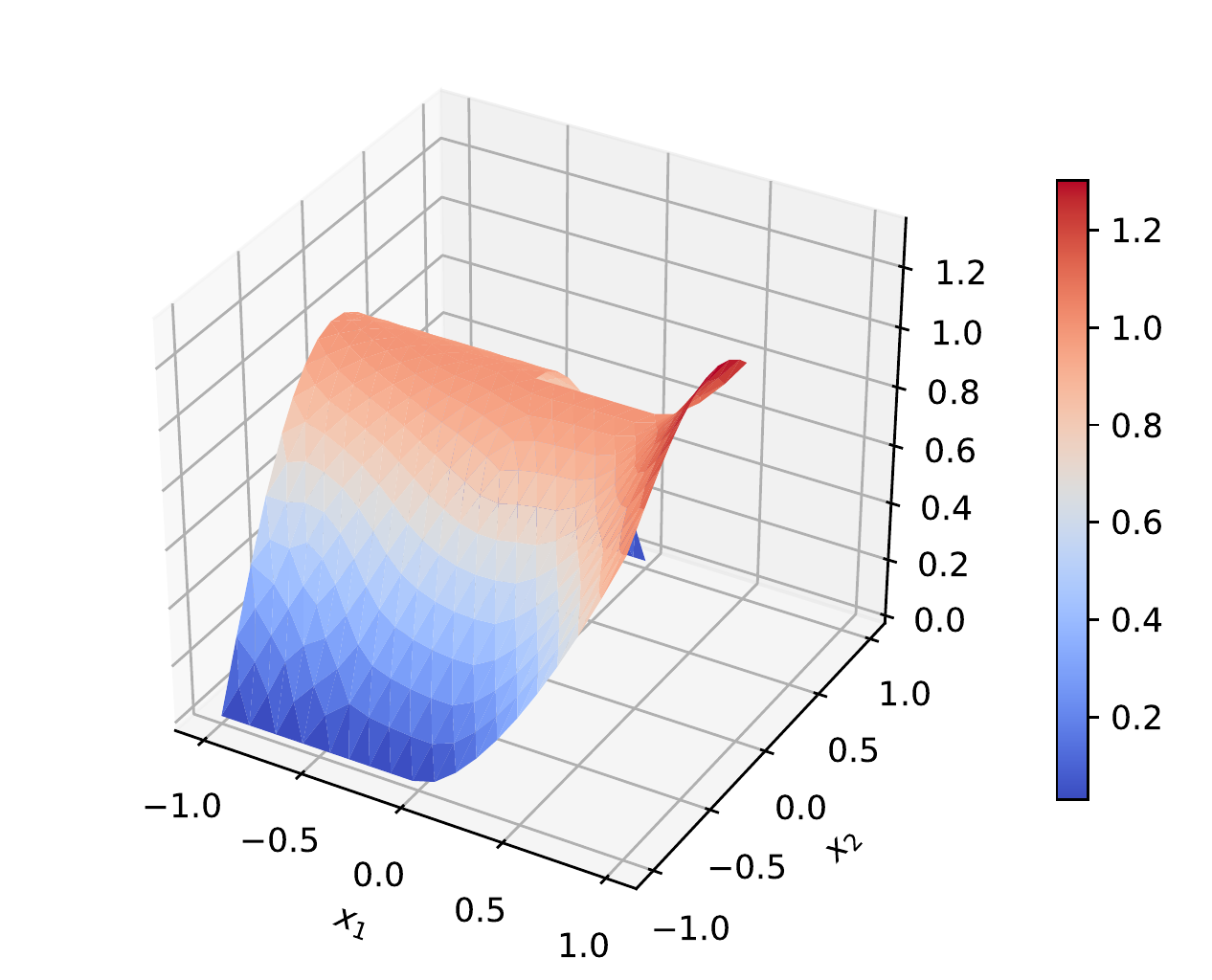}\\
 \includegraphics[width=\textwidth]{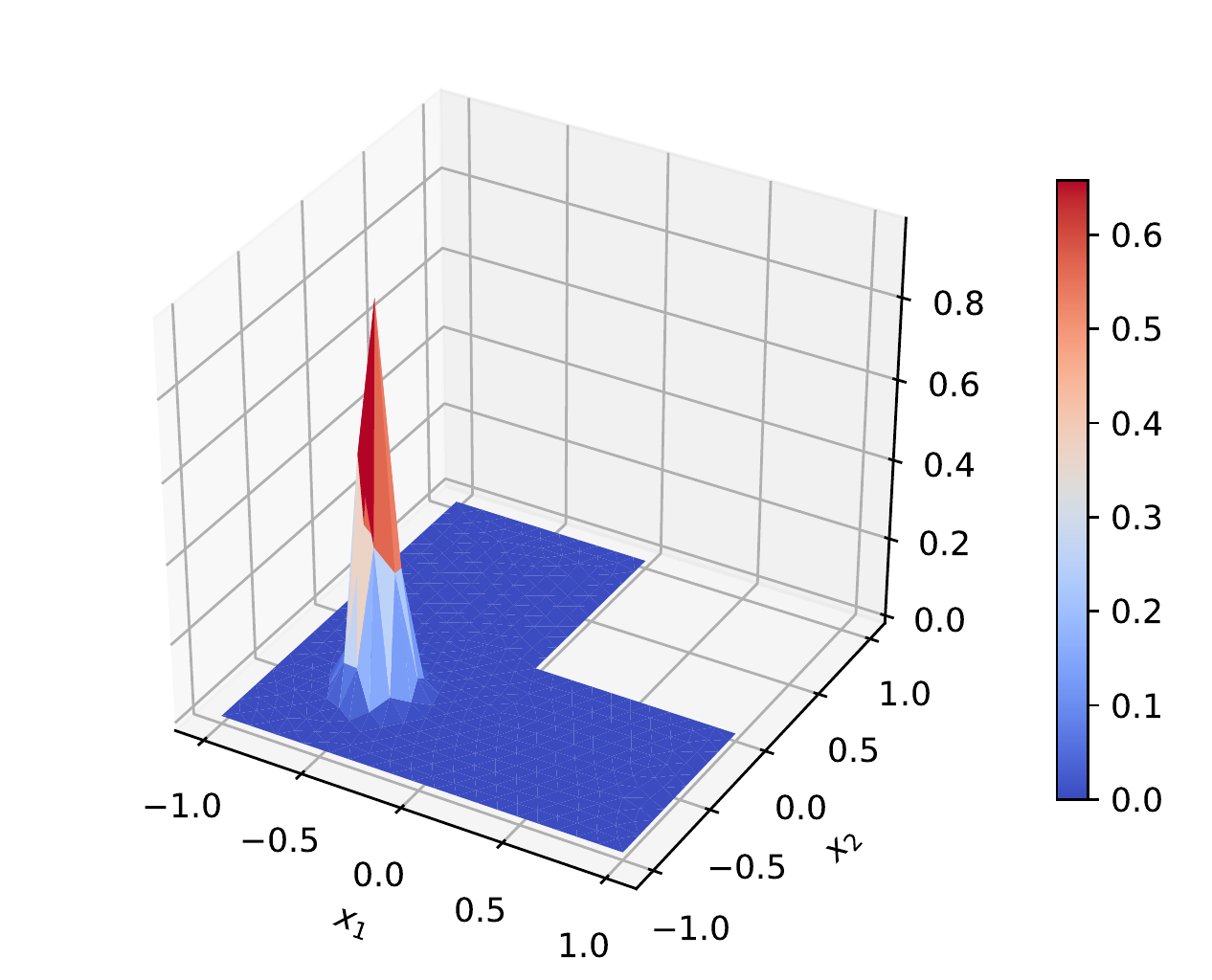}
 \end{minipage}
}
\subfigure[Predicted solution]{
\begin{minipage}[t]{0.31\linewidth}
\centering 
\includegraphics[width=\textwidth]{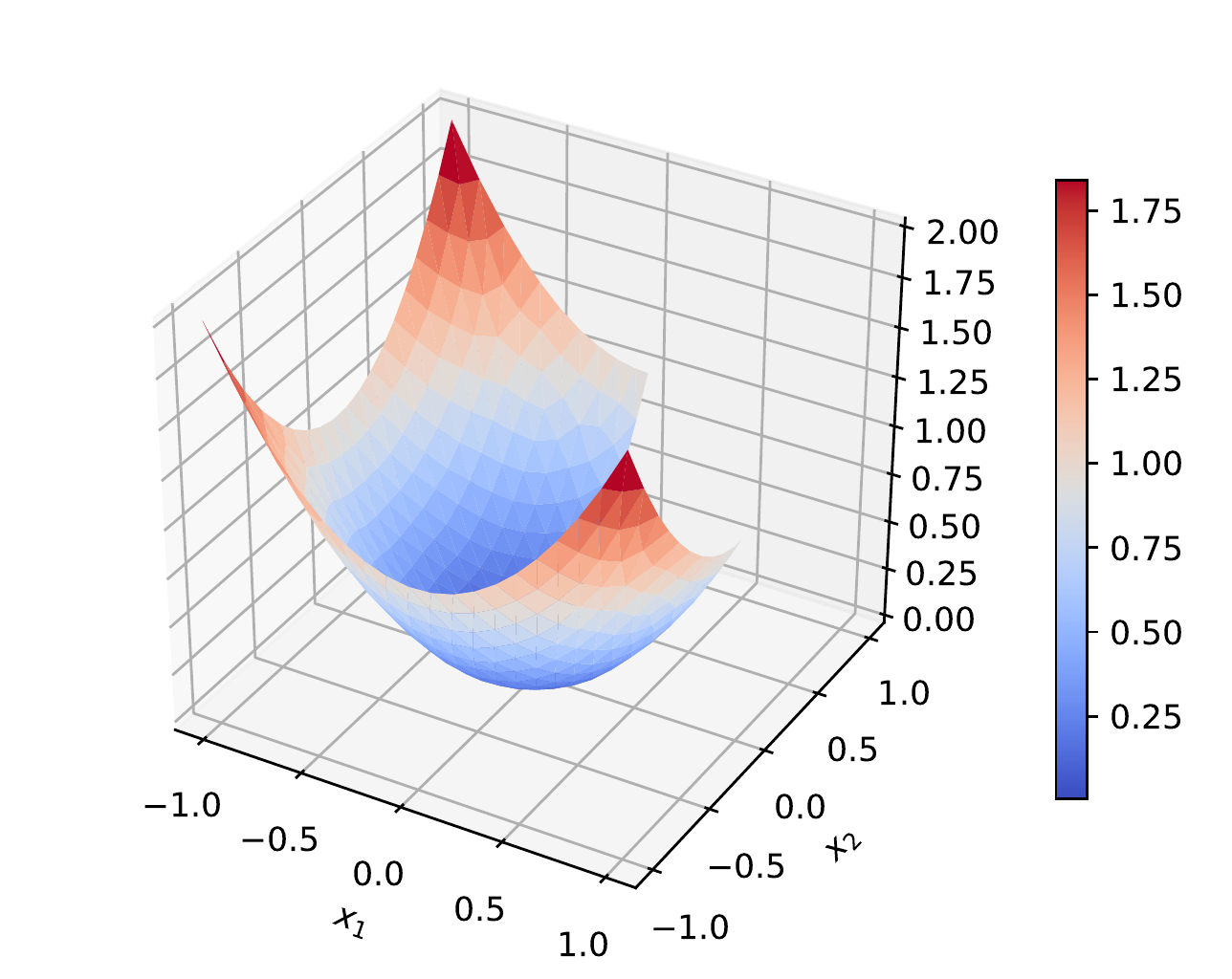}\\
 \includegraphics[width=\textwidth]{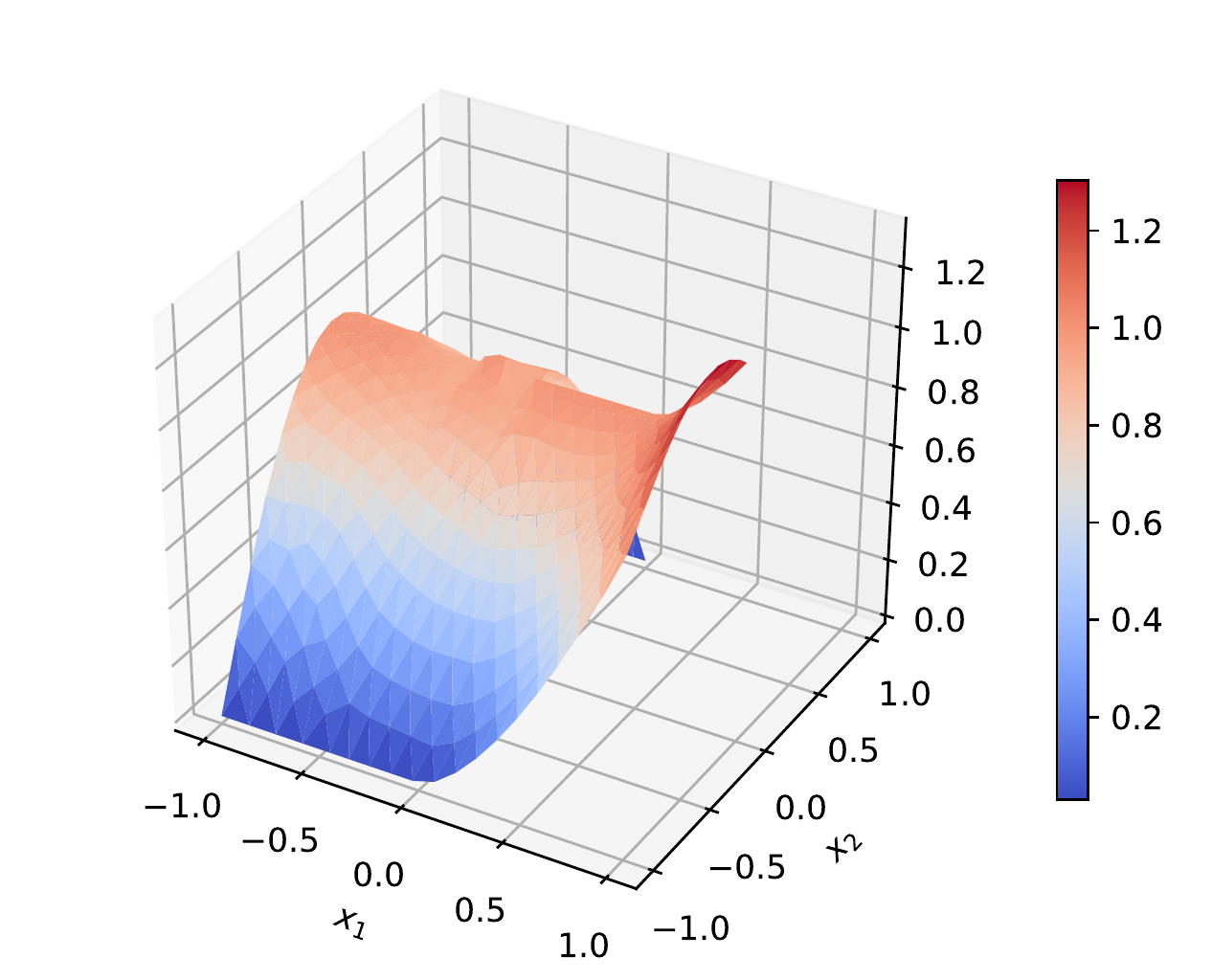}\\
 \includegraphics[width=\textwidth]{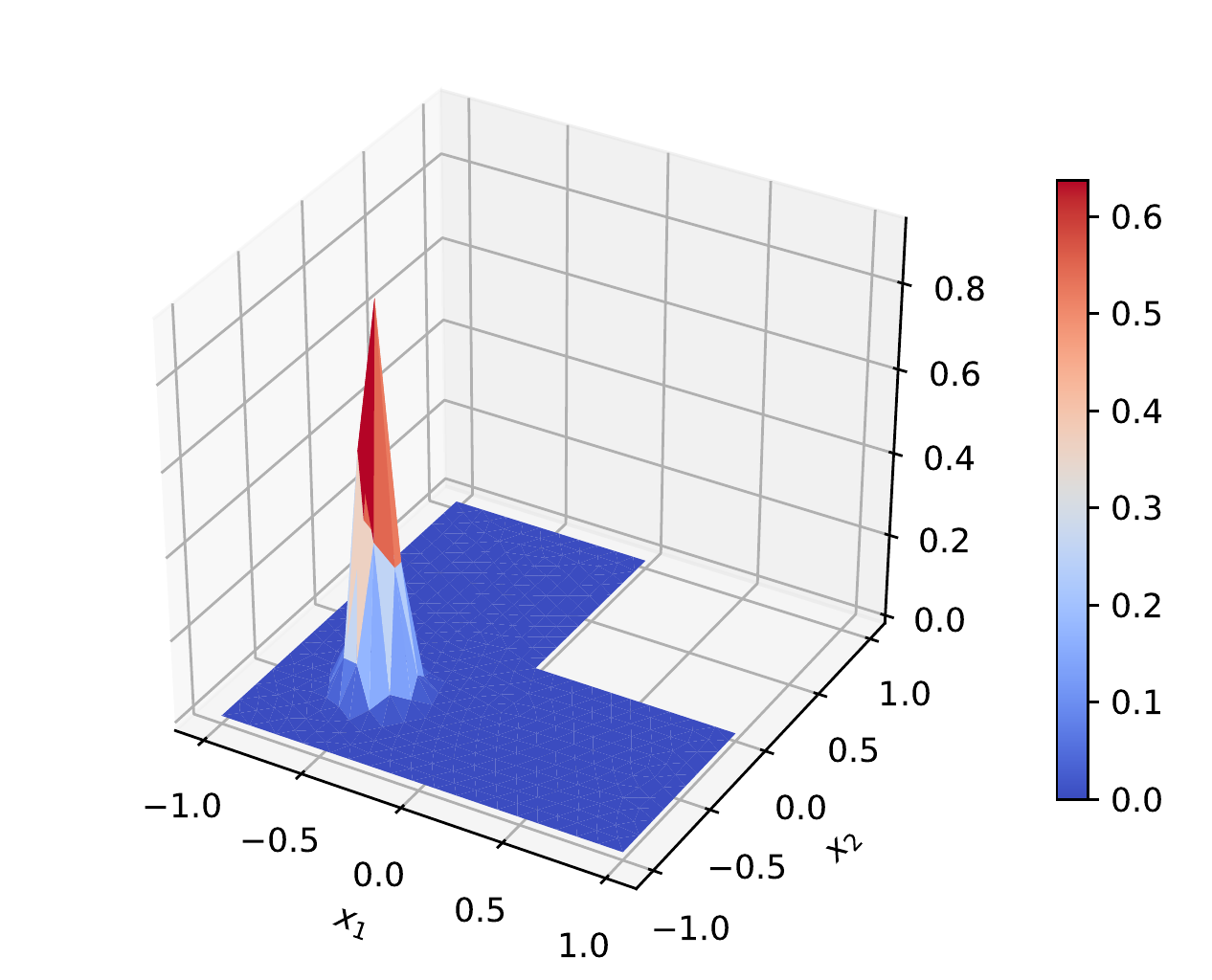}
  \end{minipage}
}
\subfigure[Error]{
\begin{minipage}[t]{0.31\linewidth}
\centering 
\includegraphics[width=\textwidth]{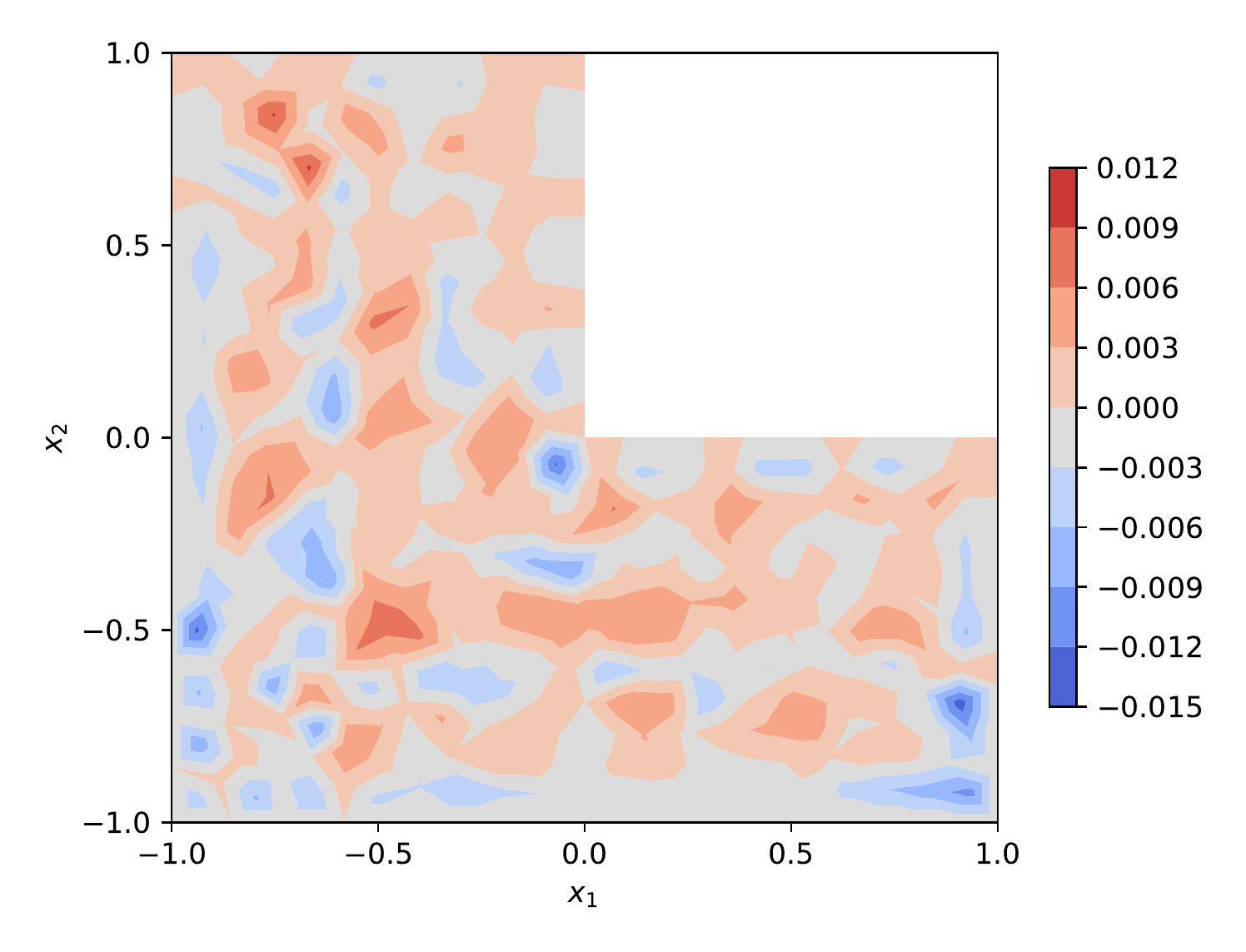}\\ 
 \includegraphics[width=\textwidth]{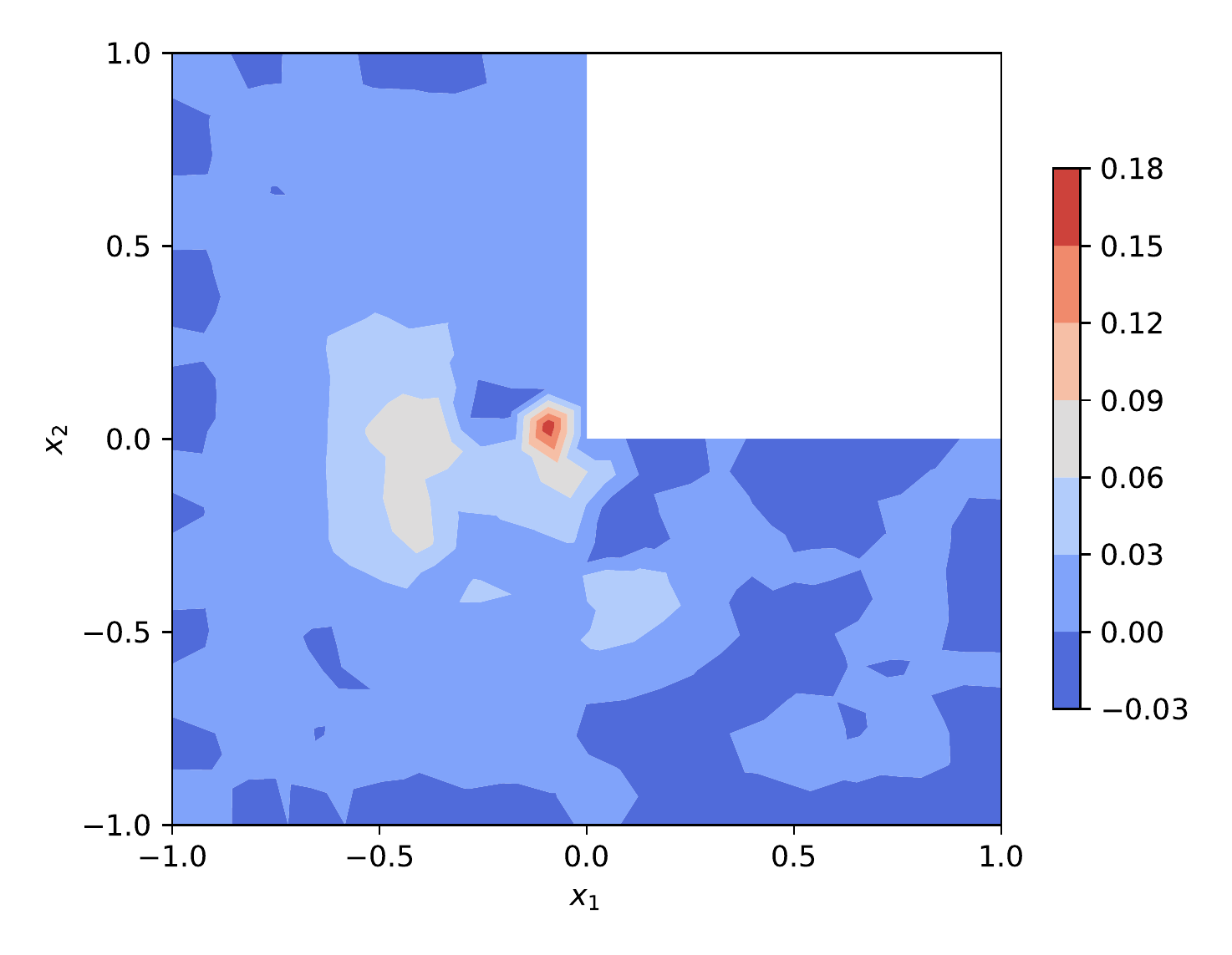}\\ 
 \includegraphics[width=\textwidth]{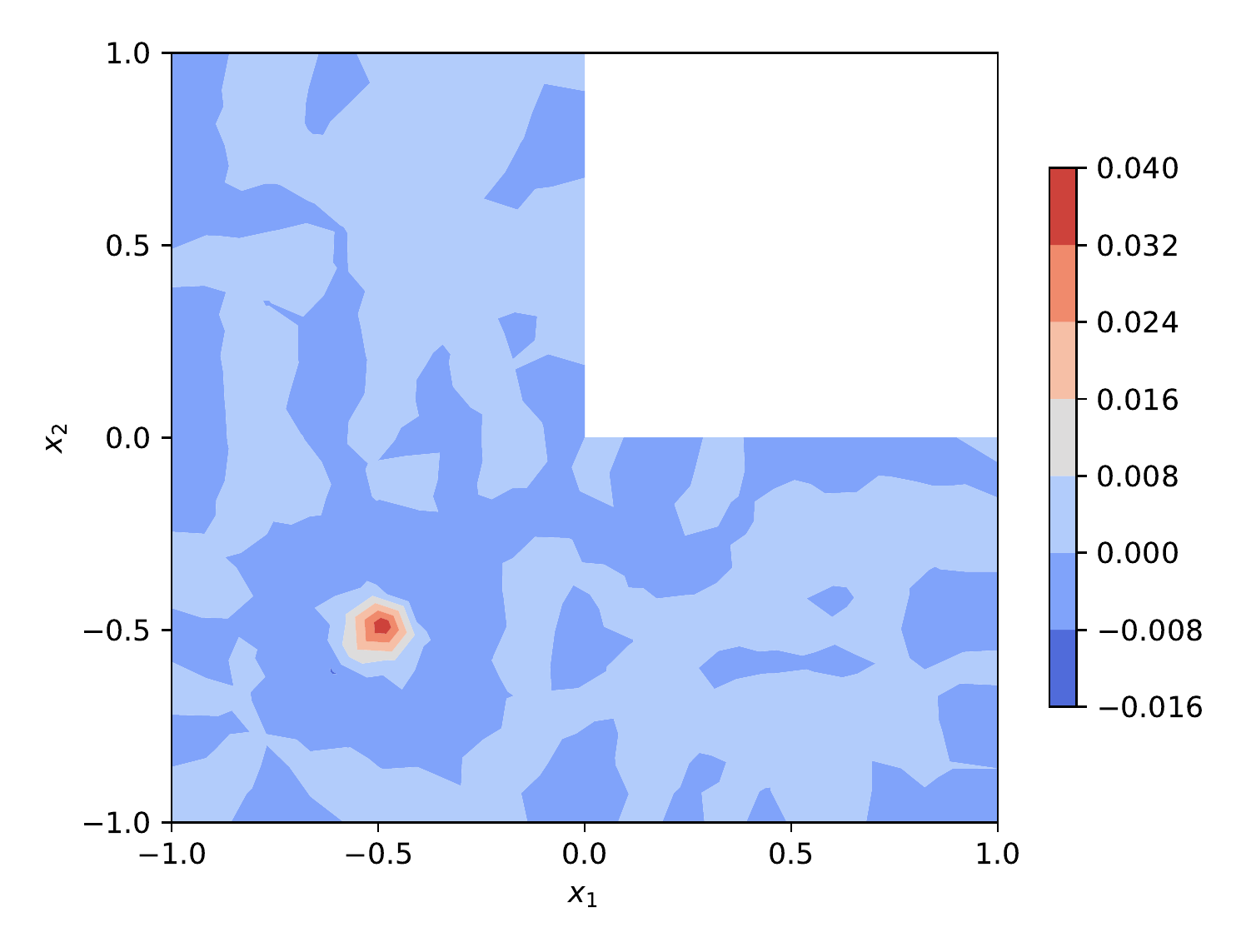}
  \end{minipage}
}}
    \caption{ 
  Numerical tests on Poisson's equation in $\Omega_3$ under different sets of  source terms and boundary conditions, where the exact solutions (left), the predicted solutions by GF-Nets (middle), and the numerical errors (right) are presented. First row: Case \eqref{B1}; second row: Case \eqref{B2}; and last row: Case \eqref{B3}. }
  \label{fig:apdxB3}
 \vspace{-0.3cm}
\end{figure}

We investigate more on the application of the GF-Net to solve Poisson's equation with Dirichlet boundary conditions. The three selected exact solutions are listed below:
\begin{align}
  u(x_1, x_2) &= x_1^2+x_2^2\tag{B.1}\label{B1}\\ 
  u(x_1, x_2) &= \begin{cases} \cos(\pi x_2/2) & x_1 \le 0.6(x_2+1) \\ \cos(\pi x_2/2) + (x_1-0.6 (x_2+1))^{\frac{3}{2}} & x_1>0.6(x_2+1)\end{cases}
\tag{B.2}\label{B2}\\ 
  u(x_1, x_2) &= e^{-100((x_1+0.5)^2+(x_2+0.5)^2)}
\tag{B.3}\label{B3}
\end{align} 
which then accordingly determine the source term $f$ and the Dirichlet boundary condition $g$ for any given domain.
The numerical results are shown in Figures \ref{fig:apdxB1}, \ref{fig:apdxB2}, \ref{fig:apdxB3}, which are produced by the trained GF-Nets with the parameter settings given in Table \ref{tab:dataset}. We observe that the approximation errors and simulation times  remain at the similar magnitudes as those reported in Table \ref{tab:PoissonWasher}.

\end{document}